%% file: main.tex
\pdfoutput=1
\PassOptionsToPackage{table}{xcolor}
\documentclass[journal,twoside,web]{ieeecolor}
\usepackage{generic}
\usepackage{cite}
\usepackage{amsmath,amssymb,amsfonts}
\usepackage{algorithmic}
\usepackage{graphicx}
\usepackage{algorithm,algorithmic}
\usepackage[hyphens]{url}       %
\usepackage[hidelinks]{hyperref} 

\hypersetup{hidelinks}
\usepackage{textcomp}

\usepackage{float}
\usepackage{multirow}
\usepackage{graphicx}
\usepackage{subcaption}
\usepackage{caption}
\usepackage[table]{xcolor}

\def\BibTeX{{\rm B\kern-.05em{\sc i\kern-.025em b}\kern-.08em
    T\kern-.1667em\lower.7ex\hbox{E}\kern-.125emX}}
\begin{document}
\title{VolE: A Point-cloud Framework for Food 3D Reconstruction and Volume Estimation}
\author{Umair Haroon, Ahmad AlMughrabi, Thanasis Zoumpekas, Ricardo Marques, and Petia Radeva.
\thanks{This work was partially funded by the EU project MUSAE (No. 01070421), 2021-SGR-01094 (AGAUR), Icrea Academia’2022 (Generalitat de Catalunya), Robo STEAM (2022-1-BG01-KA220-VET000089434, Erasmus+ EU), DeepSense (ACE053/22/000029, ACCIÓ), CERCA Programme/Generalitat de Catalunya, and Grants PID2022141566NB-I00 (IDEATE), PDC2022-133642-I00 (DeepFoodVol), and CNS2022-135480 (A-BMC) funded by MICIU/AEI/10.13039/501100011033, by FEDER (UE), and by European Union NextGenerationEU/PRTR. Ahmad AlMughrabi acknowledges the support of FPI Becas, MICINN, Spain.}
\thanks{Umair Haroon is with Universitat de Barcelona, Spain (e-mail: umairharoon@ub.edu). Ahmad AlMughrabi is also with Universitat de Barcelona, Spain (e-mail: ahmad.almughrabi@ub.edu). Thanasis Zoumpekas is with Universitat de Barcelona, Spain (e-mail: thanasis.zoumpekas@ub.edu).}
\thanks{Ricardo Marques is with Universitat Pompeu Fabra, Grup de Tecnologies Interactives (GTI), Spain (e-mail: ricardo.marques@upf.edu).}
\thanks{Petia Radeva is with Universitat de Barcelona, Spain, Institut de Neurosciències, Barcelona (e-mail: petia.ivanova@ub.edu).}}

\maketitle

\input{0_abstract}


\input{1_intro}

\input{2_related_work}
\input{3_methodology}

\input{4_results}
\input{5_conclusion}
\input{4_ack}



\section*{References}

{
    \small
    \bibliographystyle{IEEEtran}
    \bibliography{main}
}

\end{document}

%% file: 0_abstract.tex
\begin{abstract} 
Accurate food volume estimation is crucial for medical nutrition management and health monitoring applications, but current food volume estimation methods are often limited by mononuclear data, leveraging single-purpose hardware such as 3D scanners, gathering sensor-oriented information such as depth information, or relying on camera calibration using a reference object. In this paper, we present VolE, a novel framework that leverages mobile device-driven 3D reconstruction to estimate food volume. VolE captures images and camera locations in free motion to generate precise 3D models, thanks to AR-capable mobile devices. To achieve real-world measurement, VolE is a reference- and depth-free framework that leverages food video segmentation for food mask generation. We also introduce a new food dataset encompassing the challenging scenarios absent in the previous benchmarks. Our experiments demonstrate that VolE outperforms the existing volume estimation techniques across multiple datasets by achieving 2.22\% MAPE, highlighting its superior performance in food volume estimation. The source code is available at \footnote{The URL is hidden due to the blind review process.}

\end{abstract}

%% file: 1_intro.tex
\begin{figure}[t]
\setlength{\tabcolsep}{1pt}
\scriptsize
\captionsetup[subfigure]{justification=centering}
\centering
\begin{center}
\begin{tabular}{cc}
    \begin{subfigure}[b]{.33\linewidth}
        \scriptsize
         \includegraphics[width=\textwidth]{ 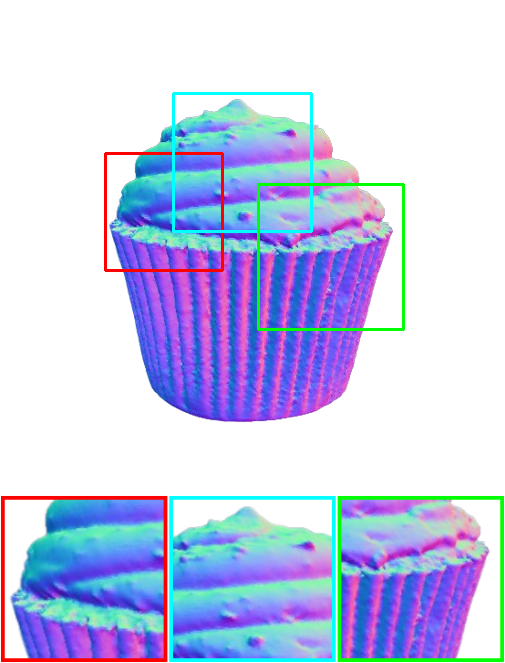}
         \raggedleft\caption{GT \\ Volume=173.13 }
     \end{subfigure} & 
     \begin{subfigure}[b]{.33\linewidth}
     \scriptsize
         \includegraphics[width=\textwidth]{ 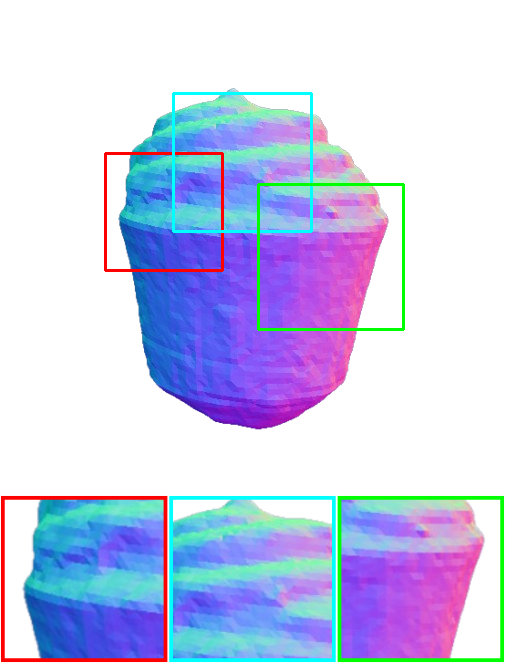}
         \setcounter{subfigure}{1}%
         \raggedleft\caption{VolETA~\cite{almughrabi2024voleta} 
         \\ EP=7.79, CD=0.0030}
     \end{subfigure} 
    \begin{subfigure}[b]{.33\linewidth}
     \scriptsize
         \includegraphics[width=\textwidth]{ 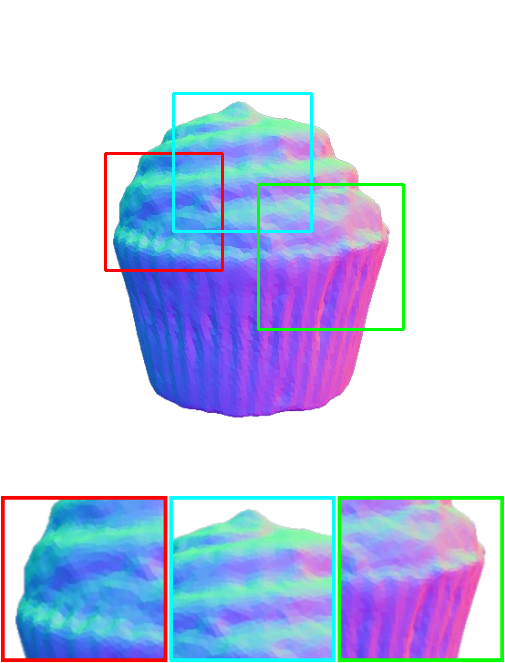}
         \setcounter{subfigure}{2}%
         \raggedleft\caption{Ours
         \\ EP=1.91, CD=0.0012}
     \end{subfigure} 
\end{tabular} 
\end{center}
   \caption{Comparison of 3D reconstruction and volume estimation for the `Cake' scene (MTF dataset): (a) Ground truth (GT) 3D model with a volume of 173.13 cm³. (b) The VolETA~\cite{almughrabi2024voleta} reconstruction shows a volume of 186.62 cm³, with a volume error percentage (EP) of 7.79\% and a Chamfer distance (CD) of 0.0030. (c) In comparison, our VolE framework achieves a volume of 176.43 cm³, with a lower EP of 1.91\% and a CD of 0.0012, demonstrating enhanced accuracy and geometric fidelity.}
\label{fig:Result_Comparison}
\end{figure}

\section{Introduction}
\label{sec:intro}

\IEEEPARstart{D}{ietary} monitoring and nutritional analysis are essential to understanding health requirements and trends at the individual and population levels~\cite{unicef2024guidance}. However, these processes encounter significant obstacles in obtaining data that is accurate, scalable, and efficiently collected~\cite{mahal2024systematic}. Traditional diet assessment methods~\cite{verbeke2024experience}, while widely used, are often time-consuming, costly, and error-prone due to their reliance on human expertise rather than objective measurements. Recent advancements in machine learning and computer vision offer promising solutions for these challenges, particularly by utilizing the informational potential of captured images. Hence, image-based intelligent dietary assessment provides a promising solution by automating tasks such as food segmentation, recognition, and volume estimation, utilizing computational image analysis~\cite{rahman2024food}. Although progress has been made in food segmentation~\cite{almughrabi2024foodmem,aguilar2022bayesian} and recognition~\cite{rodriguez2024lofi,nagarajan2023deep}, the automatic estimation of food volume remains a complex issue and is usually prone to high measurement errors~\cite{amugongo2022mobile}. The automatic estimation of food volume is essential to determine the nutritional or macronutrient content directly from captured images~\cite{he2021end, shao2021integrated}, yet achieving accurate results remains a significant challenge.

Furthermore, volume estimation is fundamentally dependent on 3D information, as understanding the dimensions and shape of an object in three-dimensional space is crucial for determining its volume~\cite{he2024metafood}. This is particularly relevant in dietary assessment, where precise measurement of food volume is essential for estimating nutritional intake~\cite{verbeke2024experience}. However, accurately estimating food volume remains an open problem due to the need for 3D data and physical references in images, which are necessary for inferring real-world sizes~\cite{bobokhidze2024standardised}. Additionally, extracting volume and density information from two-dimensional images is inherently challenging, highlighting the importance of advanced 3D reconstruction methods.

Recent advancements in 3D reconstruction, such as Neural Radiance Fields (NeRF)~\cite{mildenhall2021nerf} and Gaussian splatting~\cite{kerbl20233d}, have shown significant advantages in estimating 3D object shapes from multi-view images~\cite{konstantakopoulos20213d}. NeRFs leverage neural networks to create detailed volumetric representations of 3D scenes, while Gaussian splatting models 3D shapes use Gaussian primitives for smooth and efficient reconstructions~\cite{lyu20243dgsr}. Both techniques rely on accurate 3D camera poses, often obtained through Structure from Motion (SfM) tools (e.g., COLMAP)\cite{almughrabi2023pre}. However, these methods are inherently \textit{metric-less}, as SfM alone cannot recover the real-world scale of a scene due to the inherent ambiguity of 3D reconstruction from images, as multiple scene scales can produce identical image projections \cite{schoenberger2016sfm}. As a result, prior information about scene scale, such as object dimensions or camera distances, is essential to derive precise metric measurements~\cite{he2024metafood}. Thus, these limitations present substantial challenges in food volume estimation, where precise and reliable measurements are critical. 

On the other hand, current food volume estimation techniques ~\cite{dehais2016two, xu2013image}  face several hurdles that limit their effectiveness in real-world applications. For instance, they rely on single-purpose hardware and sensors, such as 3D scanners, depth sensors, or multiple cameras, which can be costly, inaccessible, and difficult to deploy in everyday settings~\cite{dehais2016two}. Others depend on predefined models or templates, which struggle to account for the wide variability in food shapes and textures, often resulting in inaccurate volume estimations~\cite{xu2013image}. 
While supervised learning methods \cite{ferdinand2017diabetes60, meyers2015im2calories} can provide high accuracy, they often require large amounts of annotated data and significant computational resources, limiting their practicality for widespread use. Others \cite{thames2021nutrition5k} require fixed environments, making them less adaptable to free-motion and real-world scenarios. 

Motivated by the aforementioned challenges, we propose a novel framework, entitled \textbf{VolE}, that addresses the limitations of existing methods in estimating the volume of objects, especially in scenarios involving free-motion scenarios, which are known to pose difficulties for accurate 3D reconstruction~\cite{hou2024low}. 
\textbf{VolE} leverages images and camera locations captured by AR-capable mobile devices that leverages ARCore~\cite{arcore2024} or ARKit~\cite{arkit2024} frameworks. 
Obtaining images and camera locations is now readily achievable with most modern smartphones, as these frameworks seamlessly integrate sensor data to provide accurate location and orientation information, simplifying the data acquisition process in real-time \cite{arcore2024,arkit2024}.  

By combining this spatial awareness with the captured images, we reconstruct a scaled, dense 3D point cloud of the scene. Our approach enables the free-motion and reference-free approaches. The main contributions of our proposed method are as follows:



\begin{itemize}
    \item We present a reference- and depth-free framework that leverages images and camera locations to effectively handle food geometries in free-motion scenes. Our framework does not require input masks thanks to a new auto-mask generation module that enhances reliability. Moreover, our framework refines camera locations, enabling accurate, scaled 3D reconstruction in real-world measurement units.
    
    \item We present a new benchmark dataset comprising 21 food objects featuring precise ground truth volumes and masses, specifically designed to advance research in food volume estimation techniques \footnote{The dataset will be public after article acceptance.}.  
    \item We conduct extensive experiments using our proposed VolE dataset and two benchmark datasets: MTF and DTU. Our experiments show superior performance compared with volume estimation and 3D reconstruction baselines for both MTF and DTU datasets.
\end{itemize}

This paper is structured as follows: Sec.~\ref{sec:related_work} provides a review of related work, while Sec.~\ref{fig:methodology} details the proposed methodology. In Sec.~\ref{sec:results} presents the experimental results, and Sec.~\ref{sec:conclusion} summarizes contributions and future work.

%% file: 2_related_work.tex
\section{Related work}
\label{sec:related_work}

Food volume estimation has evolved significantly, addressing the challenges of accurate volume measurement from visual data. SfM techniques exemplified by COLMAP~\cite{schoenberger2016sfm}, laid the groundwork for reconstructing 3D scene geometry and camera locations from multiple images. Furthermore, NeRF-based methods leverages COLMAP~\cite{schoenberger2016sfm} to estimate the camera locations as an input with images. For instance, InstantNGP ~\cite{mueller2022instant} is focused on rapid neural graphics primitive training to enhance reconstruction speed.  Moreover, NeuS/2~\cite{wang2021neus, wang2023neus2} offered improvements in reconstruction quality and efficiency but continued to face challenges related to adapting to the variability of shapes and textures found in food volume estimation. NeuS/2 continues to face difficulties adapting to food items' diverse shapes and textures~\cite{wang2021neus,wang2023neus2}, as shown in Figure \ref{fig:Results_DTU_dataset}. Despite all of their SfM- and NeRF-based methods benefits, they struggle with scale ambiguity that necessitates careful calibration or depth information for accurate volume estimation \cite{schoenberger2016sfm, mueller2022instant, wang2021neus, wang2023neus2}. 

To apply NeRF on food volume estimation, MetaFood CVPR challenge~\cite{he2024metafood} introduced new food volume estimation novels based on 3D reconstruction and NeRF methods, where VolETA ~\cite{almughrabi2024voleta}, the winning approach, leverages SfM \cite{schoenberger2016sfm} and NeuS/2 \cite{wang2023neus2} to create detailed food meshes and calculate precise volumes. Notably, the need for reference objects in some methods, such as VolETA~\cite{almughrabi2024voleta}, can also limit their practicality in specific real-world scenarios. ININ-VIAUN~\cite{he2024metafood} integrated deep learning with traditional computer vision techniques, utilizing Multi-View Stereo for depth information. FoodRiddle~\cite{he2024metafood} employed 3D Gaussian splatting to estimate food volumes, addressing challenges of data scarcity and complex food geometries. 

Additionally, many datasets provided the object masks \cite{jensen2014large,he2024metafood}, making the 3D reconstruction easier. However, mask generation must be leveraged for real-world scenarios. Yet, FoodMem \cite{almughrabi2024foodmem} is a good candidate to be integrated within the 3D reconstruction process as it is a near-real-time food video segmentation model leveraging SeTR \cite{zheng2021rethinking} and XMem++\cite{bekuzarov2023xmem++} to track food objects in a food scene video. Its memory mechanism enhances segmentation consistency in dynamic scenes and occlusions \cite{almughrabi2024foodmem}. In contrast, some approaches require single-purpose hardware~\cite{dehais2016two} or fixed environments \cite{thames2021nutrition5k}, limiting their practical application~\cite{gao2018food}. 

In response to these challenges, we introduce \textbf{VolE}, a novel reference- and depth-free framework that significantly improves 3D reconstruction and volume estimation for food scenes, as shown in Figure~\ref{fig:Result_Comparison}. Unlike VolETA \cite{almughrabi2024voleta}, ININ-VIAUN \cite{he2024metafood}, and FoodRiddle \cite{he2024metafood}, our framework does not rely on reference objects, depth information, or single-purpose hardware to scale reconstructed 3D models. Instead, it leverages standard mobile device capabilities, utilizing ARCore~\cite{arcore2024} for Android and ARKit~\cite{arkit2024} for iOS to capture real-world measurements through device location and IMU data during video recording or image capture. By leveraging advanced segmentation techniques such as FoodMem \cite{almughrabi2024foodmem} and robust 3D reconstruction methods, our framework aims to provide an automated and accurate volume estimation across diverse capture conditions and food types.  Our framework addresses the scale ambiguity issue by leveraging AR-based spatial mapping, ensures adaptability to various food shapes through advanced segmentation, and maintains efficiency without requiring extensive training data or complex calibration procedures. By enabling accurate food volume estimation for free-motion food scenes, our framework enhances the accessibility of food volume estimation for everyday use. 

%% file: 3_methodology.tex
\begin{figure*}[h]
\begin{center}
\includegraphics[width=1\linewidth]{ 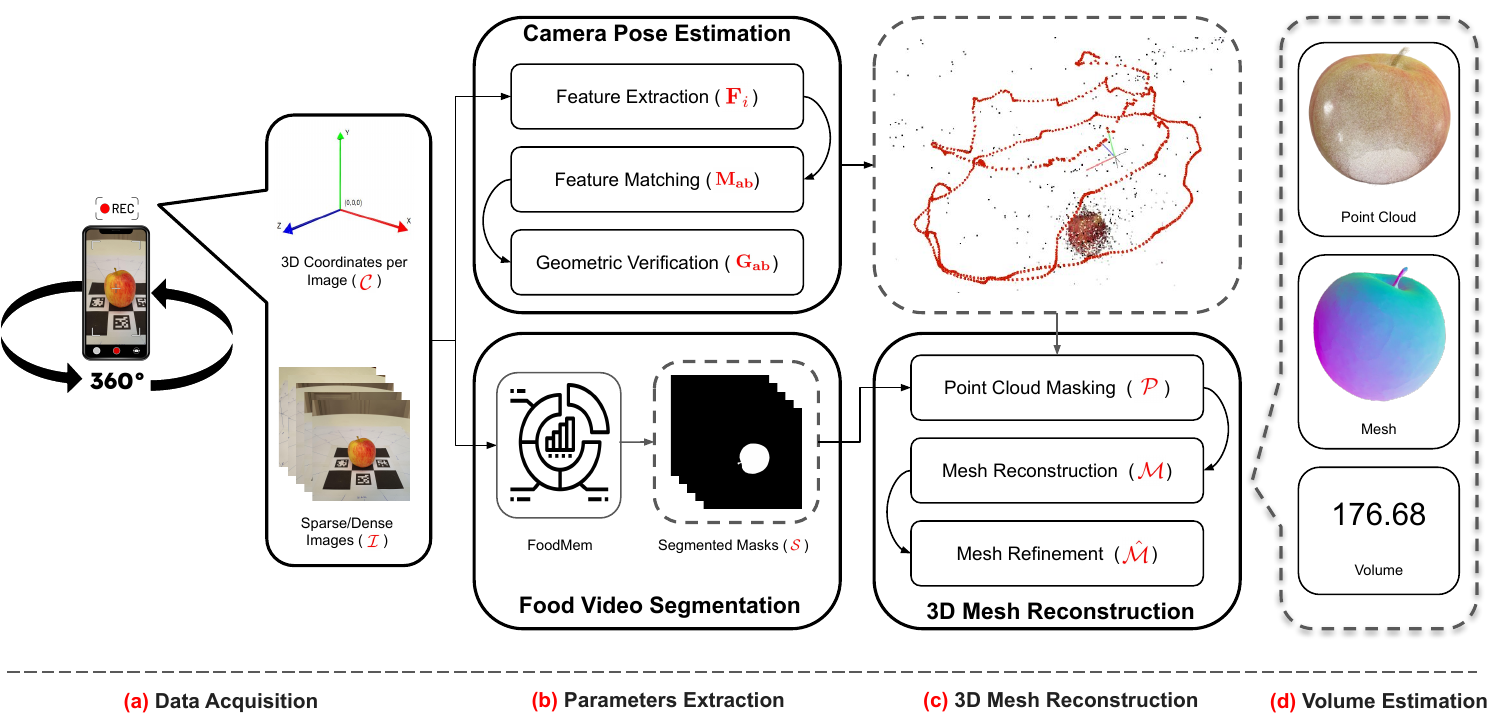}
\end{center}
    \caption{Overview of the VolE framework, illustrating its four key stages: \textbf{(a) Data Acquisition} which utilizes \textbf{ARCore/ARKit} to generate initial images (\textcolor{red}{$\mathcal{I}$}) and 3D coordinates per image (\textcolor{red}{$\mathcal{C}$});  \textbf{(b) Parameters Extraction} processing \textbf{Structure from motion} including Feature Extraction (\textcolor{red}{$\mathbf{F}_{i}$}), Feature Matching (\textcolor{red}{$\mathbf{M_{ab}}$}) and Geometric Verification (\textcolor{red}{$\mathbf{G_{ab}}$}), alongside it also includes \textbf{FoodMem} to segment interested object from RGB images and provide Segmented Masks (\textcolor{red}{$\mathcal{S}$}); \textbf{(c) 3D Mesh Reconstruction} integrating point cloud masking (\textcolor{red}{$\mathcal{P}$}), Mesh Reconstruction (\textcolor{red}{$\mathcal{M}$}) and Mesh Refinement (\textcolor{red}{$\hat{\mathcal{M}}$}) for detailed 3D representation; \textbf{(d) Volume Estimation}, where the object's volume is calculated from the refined mesh.}

\label{fig:methodology}
\end{figure*}

\section{Proposed Methodology}
\label{sec:methodology}

Our framework leverages images and camera locations using AR-capable mobile devices to accurately estimate food volumes. Our methodology section is structured as follows: in Sec.~\ref{sec:overview}, we highlight an overview; Then, we discuss the Data Acquisition in Sec.~\ref{sec:camera_parameterization}, followed by Food Video Segmentation in Sec.~\ref{sec:semantic_segmentation}; after that, we discuss Point cloud masking in Sec ~\ref{sec:point_cloud_masking}; then, Sec.~\ref{sec:mesh_reconstruction} presents the approach taken to perform Mesh reconstruction. Sec.~\ref{sec:mesh_refinement} details the \textbf{Mesh refinement} phase; and finally, Sec.~\ref{sec:volume_estimation} details our approach to actually perform the volume estimation.

\subsection{Overview}
\label{sec:overview}
As illustrated in Figure \ref{fig:methodology}, our framework consists of four main stages: (a) Data Acquisition, (b) Parameters Extraction, (c) 3D Mesh Reconstruction, and (d) Volume Estimation. In the first stage, we use mobile devices equipped with ARCore/ARKit to capture initial images ($\mathcal{I}$) and their 3D coordinates ($\mathcal{C}$). The second stage processes these inputs $(\mathcal{I}, \mathcal{C})$ using SfM for camera pose estimation, including feature extraction ($\mathbf{F}_{i}$), feature matching ($\mathbf{M_{ab}}$), and geometric verification ($\mathbf{G_{ab}}$). Simultaneously, FoodMem isolates the object of interest, generating segmented masks $\mathcal{S}$. In the third stage, we perform point cloud masking on the output of the second stage to refine the point cloud $\mathcal{P}$ and then create a 3D mesh $\mathcal{M}$, which is then refined to produce the detailed mesh $\hat{\mathcal{M}}$. Finally, the fourth stage calculates the volume of the object based on the refined mesh.

\subsection{Data Acquisition}
\label{sec:camera_parameterization}

We begin by using the capabilities of ARCore~\cite{arcore2024} or ARKit~\cite{arkit2024} on mobile devices to capture a sequence of images and camera locations, represented as $\mathcal{I} = \{I _{i}| i = 1 \ldots N_{I}\}$. Importantly, ARCore~\cite{arcore2024} and ARKit~\cite{arkit2024} provide real-time estimates of the camera pose for each captured image, allowing us to associate each image with a specific 3D location in space. We denote these 3D locations as $\mathcal{C} = \{C _{i}| i = 1 \ldots N_{I}\}$, where $C _{i}$ refers to the estimated 3D coordinates of the camera's optical center at the time the image $I _{i}$ is captured.

\subsection{Food Video Segmentation}
\label{sec:semantic_segmentation}

The food masks are essential for accurate food 3D reconstruction and volume estimation, where they help to highlight the food locations in the food scene. Therefore, the Food Video Segmentation process in our framework can be described as: given an input image $I$ containing the food item, our segmentation model assigns a semantic label to each pixel $p \in I$, classifying it into one of the predefined food categories. This can be expressed as: $S(p) = \sum_{i=1}^N i \cdot \mathbf{1}_{{p \in \mathcal{C}_i}}$, where $S(p)$ is the segmentation label assigned to pixel $p$, $\mathcal{C}$ represents the set of predefined food categories, and $N$ is the total number of categories. For a video sequence, we obtain a set of segmented masks, each corresponding to a specific food item: $\mathcal{S} = {S_1, S_2, \ldots, S_T}$, where $T$ is the total number of frames in the sequence. These segmentation masks are crucial for the subsequent stages of our framework, particularly for 3D mesh reconstruction (Sec.~\ref{sec:mesh_reconstruction}). By providing accurate and consistent food item isolation, the semantic segmentation component enables our framework to focus its reconstruction process on the food items of interest, thereby improving the overall accuracy and efficiency of the volume estimation process.

\subsection{Camera Pose Estimation}
\label{sec:camera_pose_est}

In parallel to the Food Video Segmentation, accurate camera pose estimation is crucial for precise volume estimation. The process relies on initial camera parameterization obtained from mobile devices using ARCore/ARKit. The camera poses estimation stage takes as input a set of images, denoted as $I$, along with their corresponding initial 3D coordinates, represented as $\mathcal{C}$. These inputs provide a starting point for determining the precise position and orientation of the camera for each image in the set. Our framework utilizes SfM using COLMAP~\cite{schoenberger2016sfm} to refine these initial estimates.

\begin{figure}[htb]
    \centering
    \includegraphics[width=1\linewidth]{ 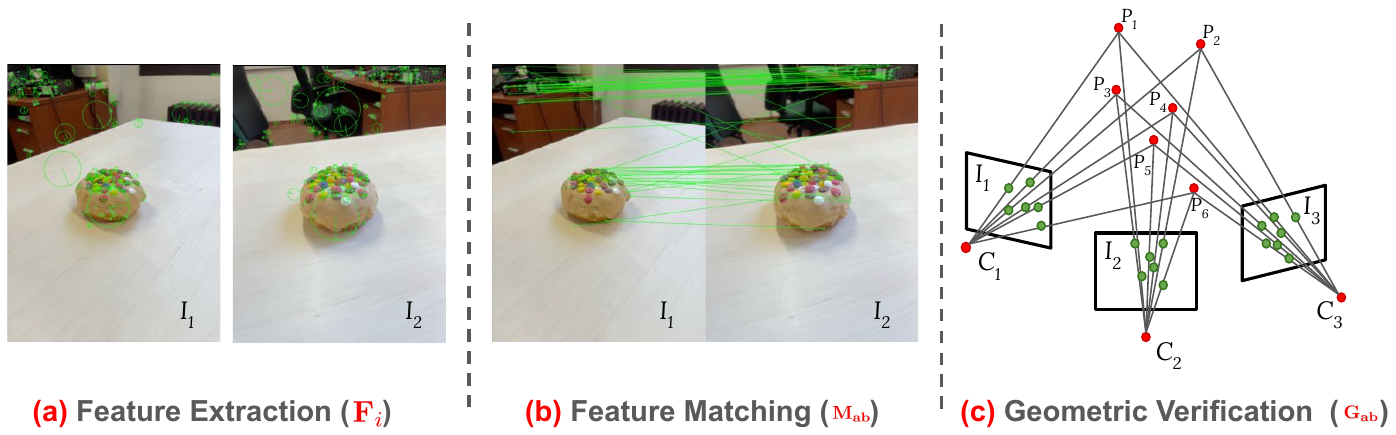}
    \caption{Visual Representation of outputs of each block of SfM (COLMAP), where \textcolor{red}{(a)} represents extracted features per image \textcolor{red}{$\mathbf{F}_{i}$}, \textcolor{red}{(b)} represents the matching features and paired images \textcolor{red}{$\mathbf{M_{ab}}$} and \textcolor{red}{(c)} represented their geometric relationships \textcolor{red}{$\mathbf{G_{ab}}$}.}
    \label{fig:camera_pose_estimation_vis}
\end{figure}

The SfM process begins by feature extraction, where distinctive local features are detected in each image $I_i$. These features are represented as $\mathbf{F}_{i} = \{(x_{j}, f_{j}) | j = 1...N_{F_{i}}\}$, where $x_{j}$ indicates the feature's location and $f_{j}$ is the appearance descriptor, with SIFT~\cite{lowe2004distinctive} and its variants~\cite{tuytelaars2008local} preferred for reliability. Fig.~\ref{fig:camera_pose_estimation_vis}(a) shows the extracted features (in green) on a given food image. After feature extraction, the feature matching phase connects images that represent the same extracted features by comparing descriptors $f_{j}$ to generate overlapping image pairs $\mathbf{C} =\{ \{I_{a}, I_{b}\} | I_{a}, I_{b} \in \mathcal{I}, a < b \}$  along with their associated feature matches $\mathbf{M_{ab}} \in \mathbf{F}_{a} \times \mathbf{F}_{b}$. Fig.~\ref{fig:camera_pose_estimation_vis}(b) illustrates the matching features (in green) on a given food image. Potential image pairs are geometrically verified using projective constraints to ensure spatial consistency of matched features, and robust methods \cite{fischler1981random}, resulting in a set of geometrically verified image pairs $\mathbf{\bar{C}}$ with inlier correspondences $\mathbf{\bar{M}{ab}}$ and their geometric relationships $\mathbf{G{ab}}$. Fig.~\ref{fig:camera_pose_estimation_vis}(c) displays the geometric verification for three food images. Finally, iterative bundle adjustment optimizes camera poses and scene geometry to minimize reprojection errors, progressively enhancing reconstruction accuracy.



\subsection{Point Cloud Masking}
\label{sec:point_cloud_masking}

Point cloud masking~\cite{cernea_openmvs} isolates specific objects within a 3D point cloud of a scene using 3D point clouds and camera poses generated from SfM. To extract the point cloud of the object of interest, we employ segmented masks $\mathcal{S}$ for each image along with the corresponding refined camera poses $C = \{(R_i, t_i) | i = 1, 2, \ldots, N_I\}$, where $R_i$ is the rotation matrix and $t_i$ is the translation vector for the $i$-th camera pose. Let $P$ denote the complete point cloud of the scene: $P = \{(x_i, y_i, z_i) | i = 1, 2, \ldots, N_P\}$, where $(x_i, y_i, z_i)$ are the 3D coordinates of the points, and $N_P$ is the total number of points. To perform point cloud masking, we project the 3D points from the scene point cloud onto the 2D segmented masks $\mathcal{S}$ using the camera intrinsic matrix $K$ and poses $C$. For each point $P_i$, its projection onto image $j$ is computed as $(u, v, w)^\top = K \cdot [R_j | t_j] \cdot P_i$, followed by homogenization to obtain $p_{ij} = \left(\frac{u}{w}, \frac{v}{w}\right)$. We then identify the points that fall within the segmented masks $\mathcal{S}_j$ for each image, defining a set of valid points $M_j = \{ P_i \mid p_{ij} \in \mathcal{S}_j \}$. The final segmented object point cloud $\mathcal{P}$ is the intersection of all these valid points across all images: $\mathcal{P} = \bigcap_{j=1}^{N_I} M_j$. This ensures that only points visible and masked in all views are retained, leading to the final segmented point cloud of the required object, given by $\mathcal{P} = \{ (x_i, y_i, z_i) \in P \mid \forall j \, (m_{ij} = 1) \}$. Points that do not fall within every mask are considered part of the background and are discarded, efficiently isolating the object of interest.

\begin{figure}[htb]
\setlength{\tabcolsep}{1pt}
\captionsetup[subfigure]{justification=centering}
\centering
\begin{center}
\begin{tabular}{cc}
    \begin{subfigure}[b]{.24\linewidth}
        \scriptsize
         \captionsetup{font=scriptsize}
         \includegraphics[width=\textwidth]{ 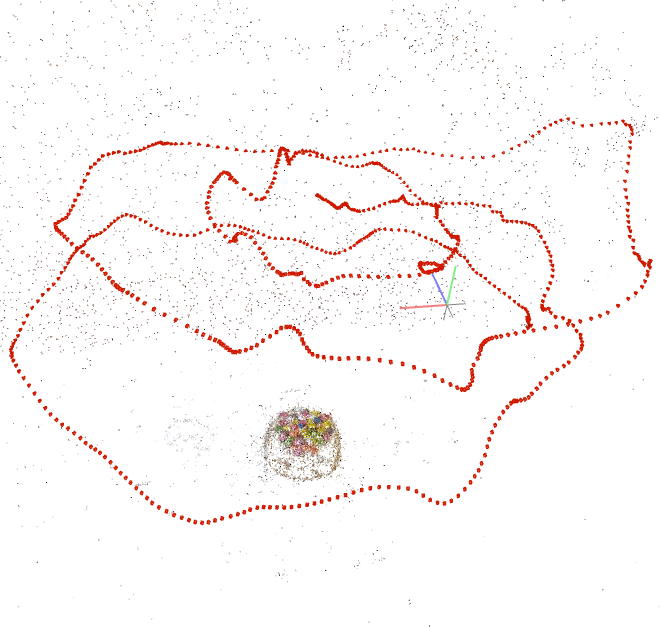}
         \caption{Colmap Scene Pointcloud (\textcolor{red}{$P$}) }
     \end{subfigure} & 
     \begin{subfigure}[b]{.24\linewidth}
     \scriptsize
         \captionsetup{font=scriptsize}
         \includegraphics[width=\textwidth]{ 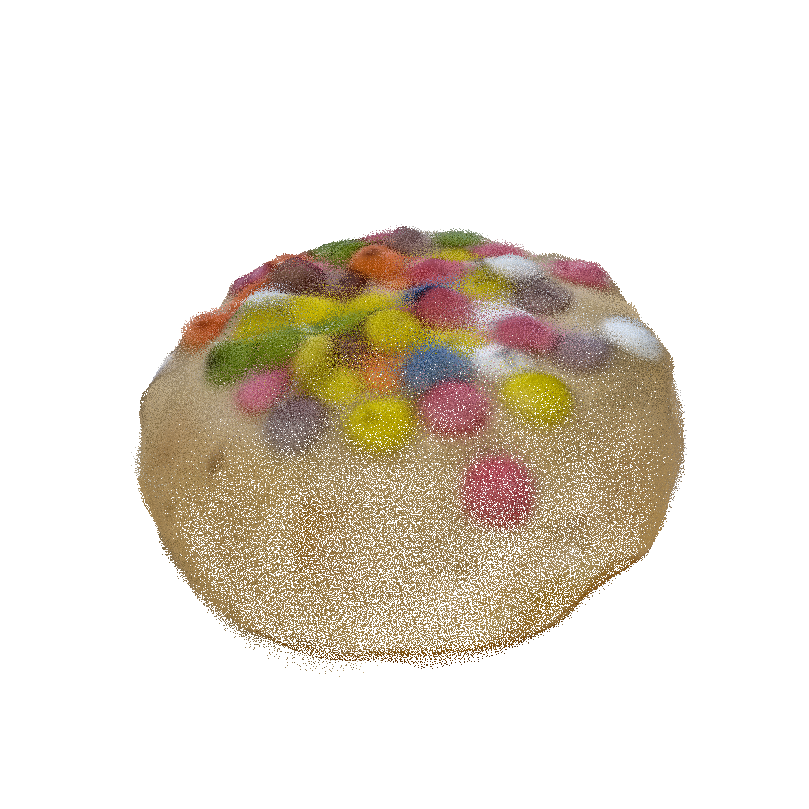}
         \setcounter{subfigure}{1}%
         \caption{ Point Cloud Masking (\textcolor{red}{$\mathcal{P}$})}
     \end{subfigure} 
    \begin{subfigure}[b]{.24\linewidth}
     \scriptsize
         \captionsetup{font=scriptsize}
         \includegraphics[width=\textwidth]{ 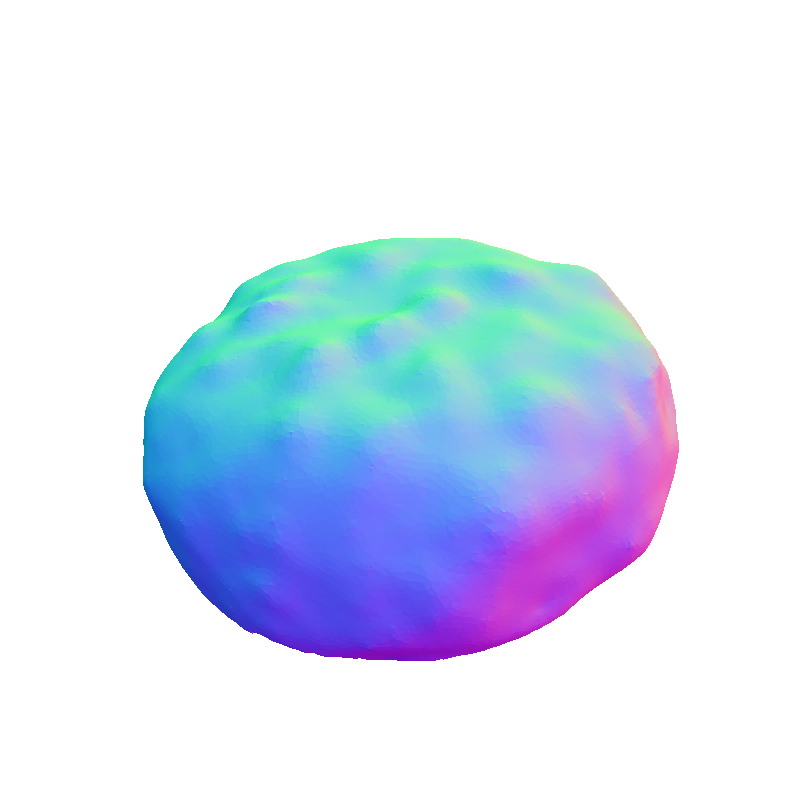}
         \setcounter{subfigure}{2}%
         \caption{Mesh Reconstruction (\textcolor{red}{$\mathcal{M}$})}
     \end{subfigure}
     \begin{subfigure}[b]{.24\linewidth}
     \scriptsize
         \captionsetup{font=scriptsize}
         \includegraphics[width=\textwidth]{ 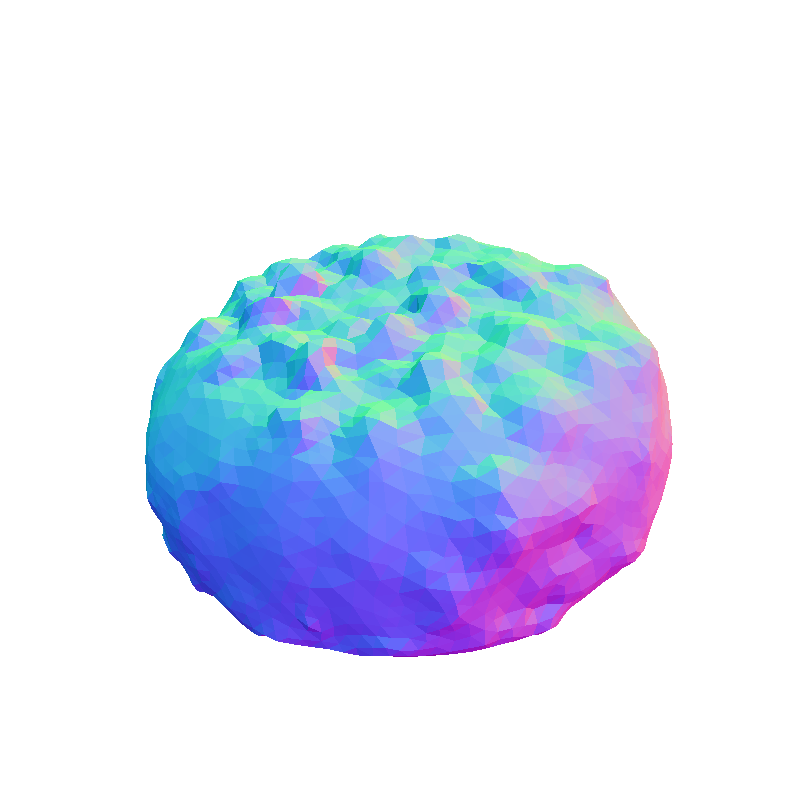}
         \setcounter{subfigure}{3}%
         \caption{Mesh Refinement (\textcolor{red}{$\hat{\mathcal{M}}$})}
     \end{subfigure}
\end{tabular} 
\end{center}
   \caption{The illustration of the VolE framework output as a 3D reconstruction pipeline, covering each step from SfM to mesh refinement. This includes (a) Sparse point cloud reconstruction using COLMAP (\textcolor{red}{$\mathcal{P}$}), (b) Segmented point cloud masking (\textcolor{red}{$\mathcal{P}$}), (c) Initial mesh reconstruction (\textcolor{red}{$\mathcal{M}$}), and (d) Final refined mesh (\textcolor{red}{$\hat{\mathcal{M}}$}), showcasing the transition from sparse to dense, high-quality mesh through successive processing stages.}
\label{fig:VolE_framework_outputs}
\end{figure}

\subsection{Mesh Reconstruction}
\label{sec:mesh_reconstruction}
Mesh reconstruction~\cite{cernea_openmvs} involves creating a mesh from the segmented point cloud $\mathcal{P}$ using a Multi-View Stereo (MVS)~\cite{haroon2024mvsboost} approach. The point cloud $\mathcal{P}$ is first transformed into a tetrahedral mesh $\mathcal{T}$ through the $f_{D}$ Delaunay triangulation function~\cite{cernea_openmvs}: $ \mathcal{T} = f_{D}(\mathcal{P})$. Then, a graph-cut optimization $f_{G}$ determines whether each tetrahedron is inside or outside the object: $\mathcal{L} = f_G(\mathcal{T})$. Finally, the marching cubes algorithm~\cite{cernea_openmvs} $\mathbf{M}$ extracts the mesh surface from the labeled tetrahedra: $\mathcal{M} = \mathbf{M}(\mathcal{T}, \mathcal{L})$. The resulting mesh $\mathcal{M}$ is a seamless and accurate representation of the object's geometry based on the original point cloud data. 

\subsection{Mesh Refinement}
\label{sec:mesh_refinement}

Mesh refinement~\cite{cernea_openmvs} is a crucial process for improving the quality of reconstructed meshes, involving techniques to enhance accuracy, smoothness, and overall representation of the object's surface characteristics. In our framework, mesh refinement is essential for ensuring that reconstructed meshes accurately represent the object's geometry while eliminating artifacts from earlier steps. High-quality meshes are vital for downstream applications like visualization, simulation, or analysis, where inaccuracies can significantly impact results.  While \textbf{mesh simplification}~\cite{cernea_openmvs} $f_S$ might seem counterintuitive in this context, it helps manage computational resources by reducing the number of vertices $\mathcal{M}_{s} = f_{S}(\mathcal{M})$, making subsequent refinement steps more efficient. Following simplification, mesh smoothing $f_{\bar{S}}$ uses Laplacian or bilateral filtering to eliminate noise and outliers $\mathcal{M}_{\bar{s}} = f_{\bar{S}}(\mathcal{M}_{s})$, creating a more uniform surface. \textbf{Mesh denoising} $f_D$~\cite{cernea_openmvs}  further enhances quality by removing remaining noise through standard voting tensor filtering $\mathcal{M}_{d} = f_D (\mathcal{M}_{\bar{s}})$. Finally, \textbf{mesh optimization}~\cite{cernea_openmvs} $\hat{f}$, such as vertex relaxation and edge flipping, refine triangle quality to produce a clean and precise mesh $\hat{\mathcal{M}} = \hat{f}(\mathcal{M}_{d})$. By integrating these simplification, smoothing, denoising, and optimization processes, our framework achieves high-quality meshes $\hat{\mathcal{M}}$ that balance computational efficiency with geometric accuracy.

\subsection{Volume Estimation}
\label{sec:volume_estimation}

The volume estimation of a 3D mesh is based on the divergence theorem \cite{divergence_theorem}, which connects volume integrals to surface integrals. For a closed triangular mesh $\hat{\mathcal{M}}$ consisting of $N$ faces, the volume $V$ is computed by summing the signed volumes of tetrahedra formed by each triangle face and the origin: $V = \frac{1}{6} \sum_{k=1}^{N} (v_1^k \cdot (v_2^k \times v_3^k))$, where $v_1^k, v_2^k, v_3^k \in \mathbb{R}^3$ are the vertices of the $k$-th triangle. Each tetrahedron is formed using the vertices of a triangle, along with the origin as the fourth vertex. The volume of each tetrahedron is calculated using the scalar triple product; the cross product ($\times$) determines the orientation and area of the triangle, while the dot product ($\cdot$) gives the volume of the tetrahedron. The factor of $\frac{1}{6}$ comes from the volume formula for tetrahedra. By summing these individual tetrahedron volumes, the total mesh volume is obtained efficiently in a single pass through all triangles, making this method computationally effective for complex 3D meshes.

%% file: 4_results.tex
\section{Experimental Results}
\label{sec:results}

This section comprehensively evaluates our framework for object volume estimation and 3D reconstruction. We conduct extensive experiments across multiple datasets to validate its accuracy and robustness. We compare our approach with state-of-the-art methods like COLMAP~\cite{schoenberger2016sfm}, NeuS/2~\cite{wang2021neus,neus2}, InstantNGP~\cite{mueller2022instant}, InstantNSR~\cite{zhao2022human},  VolETA~\cite{almughrabi2024voleta}, ININ~\cite{he2024metafood}, and FoodR.~\cite{he2024metafood}, which are recognized for their performance on the MTF~\cite{he2024metafood} and DTU~\cite{jensen2014large} datasets. This comparison includes both quantitative and qualitative results, ensuring a rigorous evaluation. Additionally, we analyze the performance of our proposed framework under varying conditions, including changes in object shape, size, and capturing topology. Our goal is to demonstrate its adaptability and robustness in real-world applications.

\subsection{Implementation settings}

We conducted our experiments using an NVIDIA GPU: GeForce RTX 3090/24 GB. For the Foodkit and MTF datasets, we configured point cloud masking at "max-resolution" 512 to balance detail preservation with computational efficiency, followed by mesh reconstruction with "close-holes" 50 to address occlusions and "smooth" 5 for surface regularization. For the DTU dataset, we required enhanced reconstruction capabilities, so we configured the settings to "max-resolution" 3072 and "sub-resolution-levels" 8 to effectively handle high-resolution scans. Additionally, we implemented a multi-stage refinement process with "iters" 3 for depth estimation and "geometric-iters" 2 to ensure geometric consistency.

\subsection{Evaluation Protocol}

In order to assess the accuracy of our proposed framework in volume estimation, we employ the Mean Average Percentage Error (MAPE) metric. MAPE is calculated as:
\begin{equation}
  \text{MAPE} = \frac{1}{n} \sum_{i=1}^n \left|\frac{V_{est,i} - V_{gt,i}}{V_{gt,i}}\right| \times 100\% \, ,
\end{equation}
where $V_{est,i}$ is the estimated volume, $V_{gt,i}$ is the ground truth volume, and $n$ is the number of objects. For evaluating 3D reconstruction accuracy, we utilize the Chamfer distance metric~\cite{barrow1977parametric}, which measures the average distance between the reconstructed and ground truth point cloud sampled from the surface of the mesh.

\subsection{Datasets}

We evaluate our method on the proposed Foodkit dataset, as well as, MTF~\cite{he2024metafood}, and DTU~\cite{jensen2014large} datasets. Below, we present a detailed explanation of each dataset
\subsubsection{Foodkit: Our proposed dataset}

\begin{figure}[htb]
    \centering
    \captionsetup[subfigure]{labelformat=empty, font=tiny}
    \setlength{\tabcolsep}{1pt}
    \begin{tabular}{ccccccc}
        \subcaptionbox{\centering Apple}{\includegraphics[clip,width=0.13\linewidth]{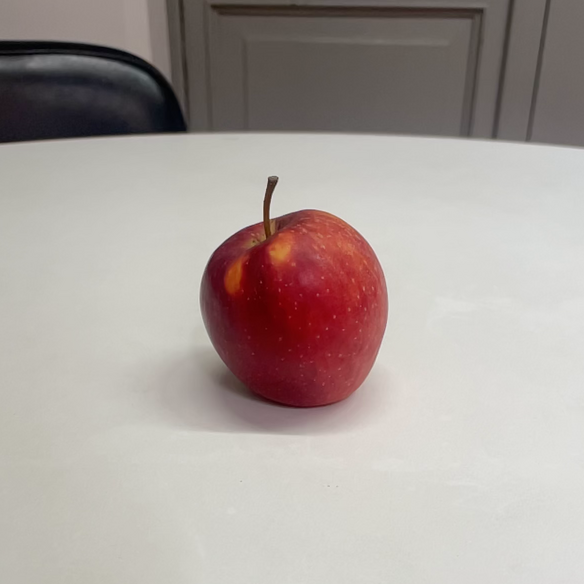}} &
        \subcaptionbox{\centering Orange}{\includegraphics[clip,width=0.13\linewidth]{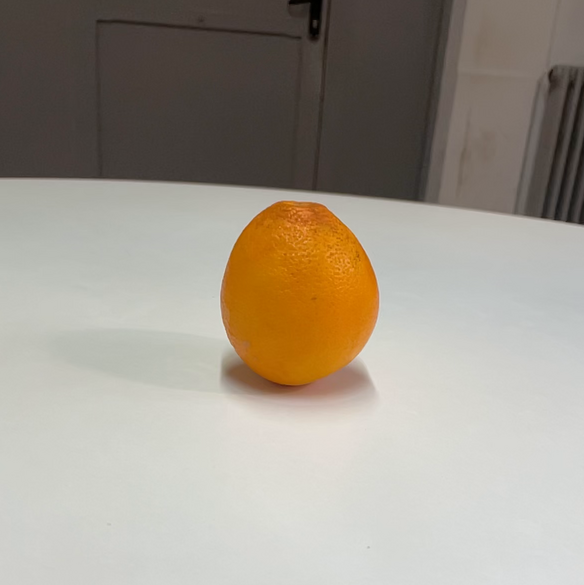}} &
        \subcaptionbox{\centering Aguacate}{\includegraphics[clip,width=0.13\linewidth]{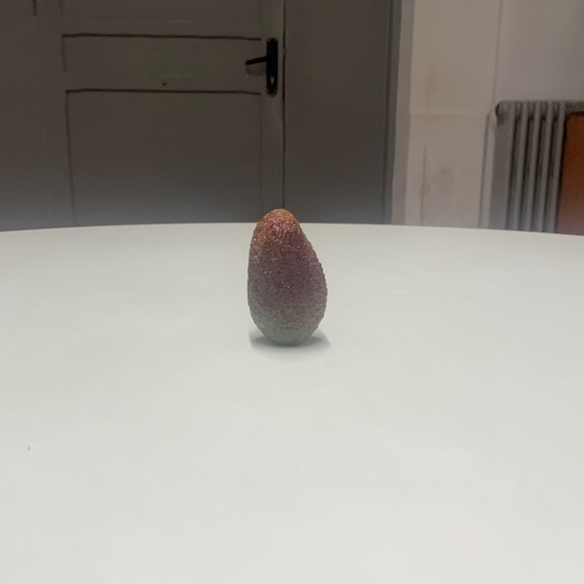}} &
        \subcaptionbox{\centering Lemon}{\includegraphics[clip,width=0.13\linewidth]{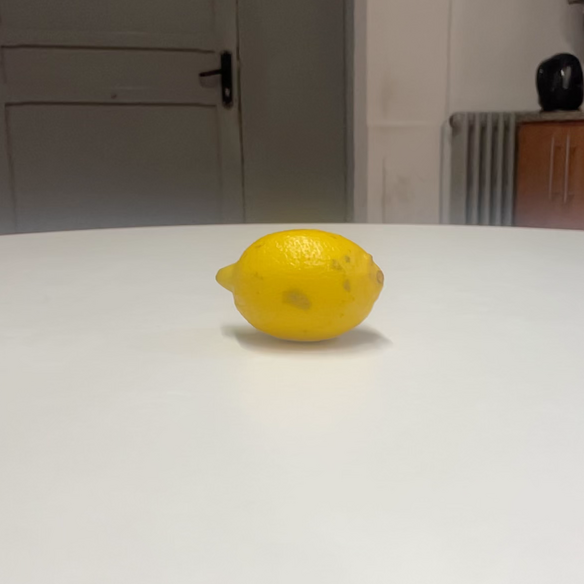}} &
        \subcaptionbox{\centering Donut}{\includegraphics[clip,width=0.13\linewidth]{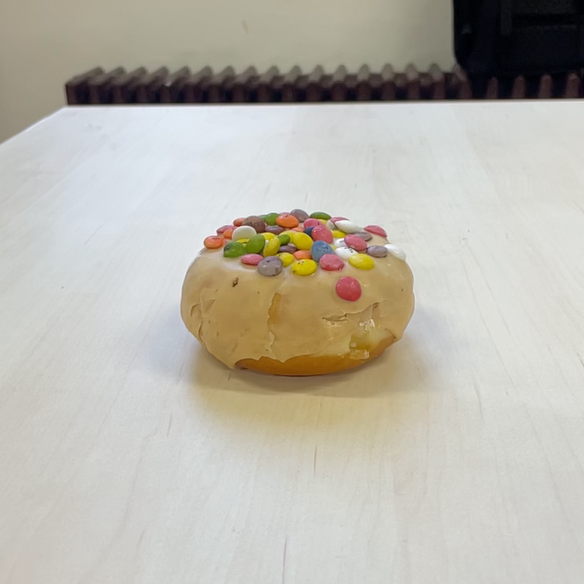}} &
        \subcaptionbox{\centering Durum}{\includegraphics[clip,width=0.13\linewidth]{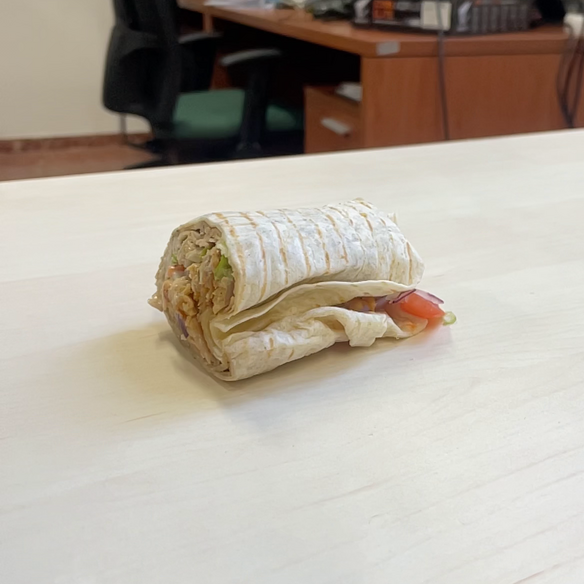}} &
        \subcaptionbox{\centering Pear}{\includegraphics[clip,width=0.13\linewidth]{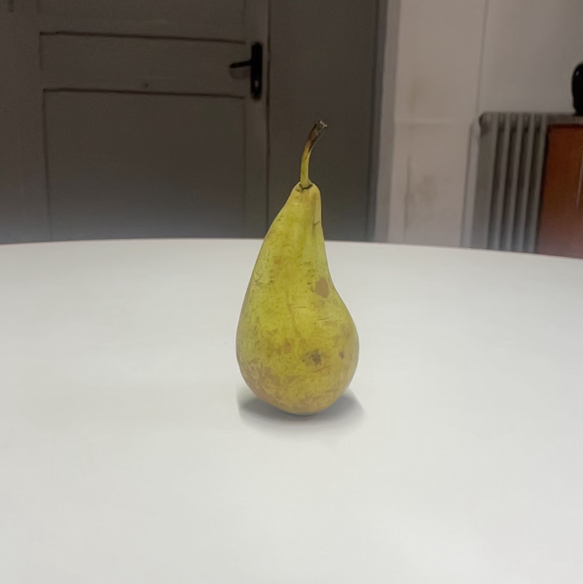}}
        \\
        \subcaptionbox{\centering Samosa}{\includegraphics[clip,width=0.13\linewidth]{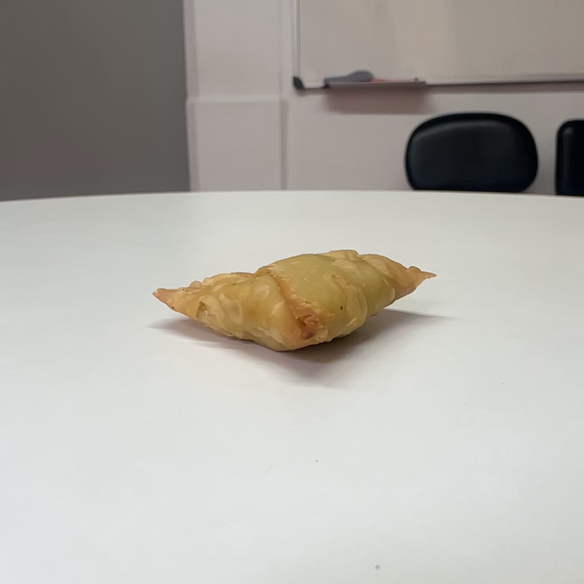}} &
        \subcaptionbox{\centering Apple Pie}{\includegraphics[clip,width=0.13\linewidth]{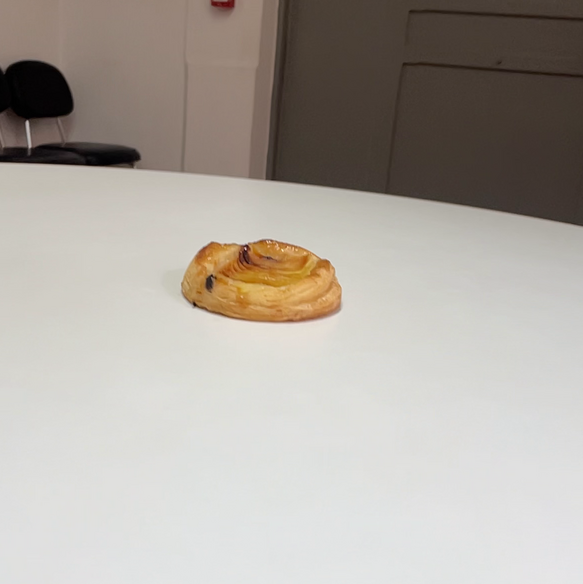}} &
        \subcaptionbox{\centering Empanadilla}{\includegraphics[clip,width=0.13\linewidth]{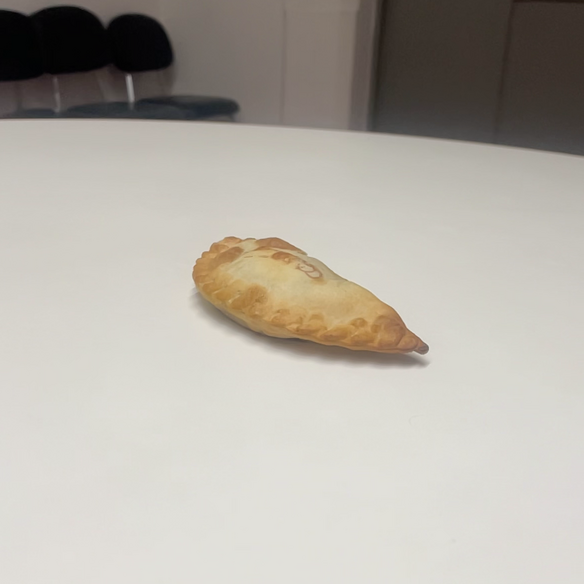}} &
        \subcaptionbox{\centering Falafel}{\includegraphics[clip,width=0.13\linewidth]{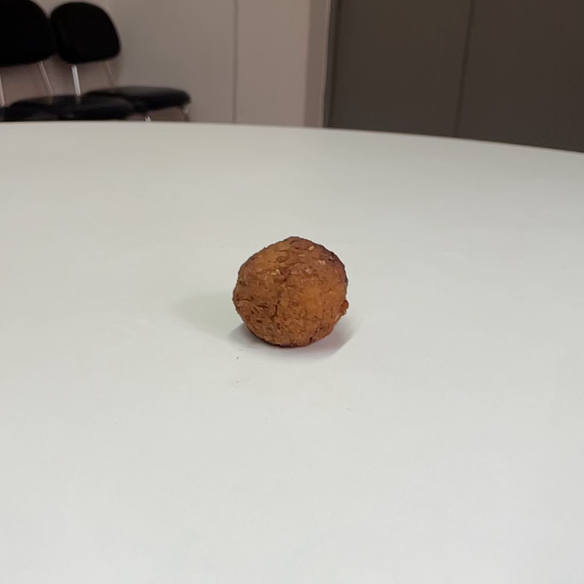}} &
        \subcaptionbox{\centering Napolitanas}{\includegraphics[clip,width=0.13\linewidth]{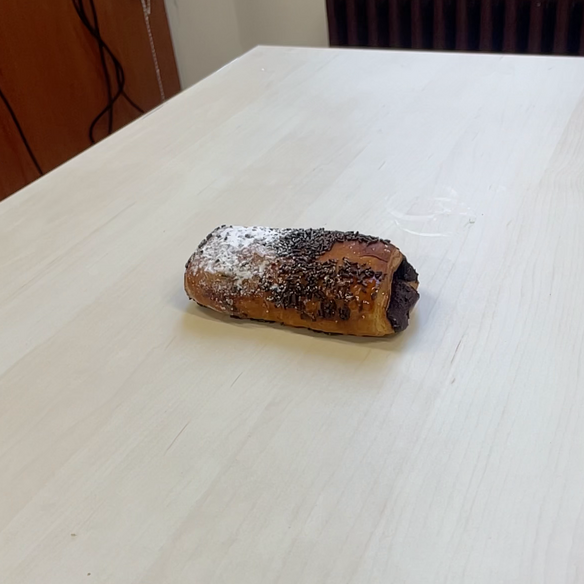}} &
        \subcaptionbox{\centering Capsicum}{\includegraphics[clip,width=0.13\linewidth]{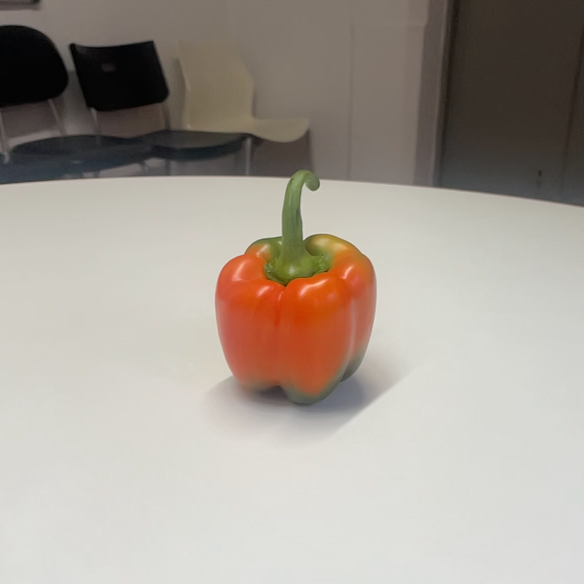}} &
        \subcaptionbox{\centering Banana}{\includegraphics[clip,width=0.13\linewidth]{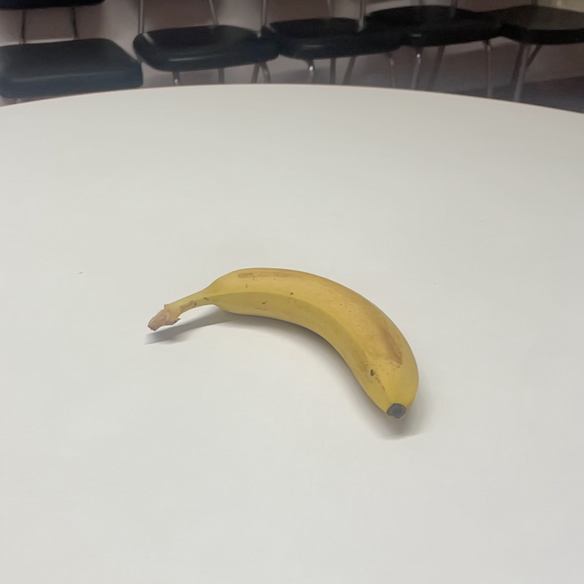}}
        \\
        \subcaptionbox{\centering Paxoco Mini}{\includegraphics[clip,width=0.13\linewidth]{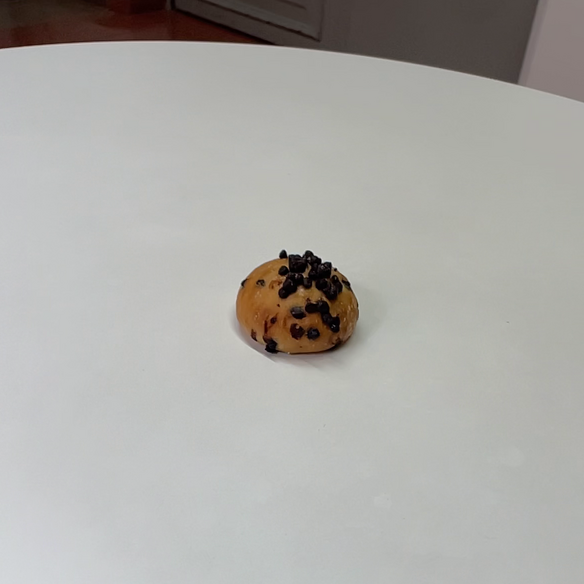}} &
        \subcaptionbox{\centering Chocolate Cake}{\includegraphics[clip,width=0.13\linewidth]{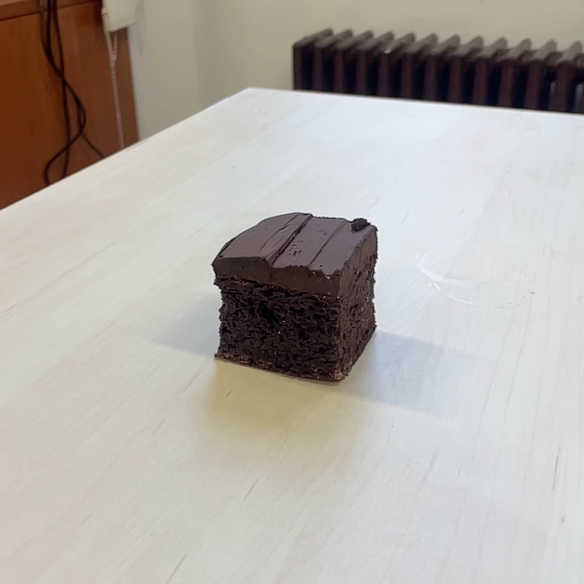}} &
        \subcaptionbox{\centering Chocolate Croissant}{\includegraphics[clip,width=0.13\linewidth]{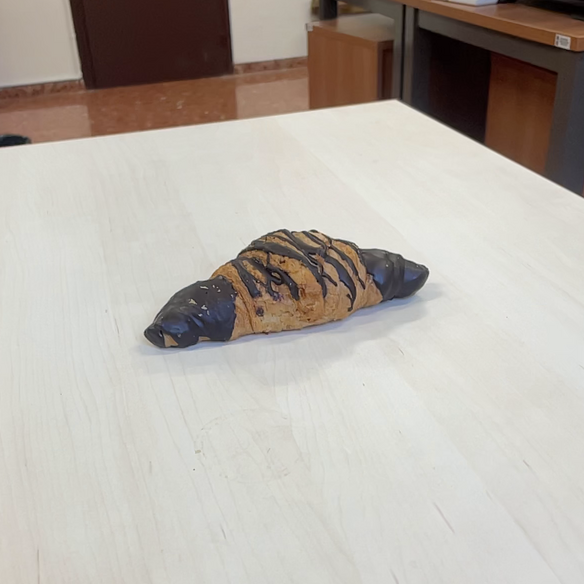}} &
        \subcaptionbox{\centering Chocolate Bomb}{\includegraphics[clip,width=0.13\linewidth]{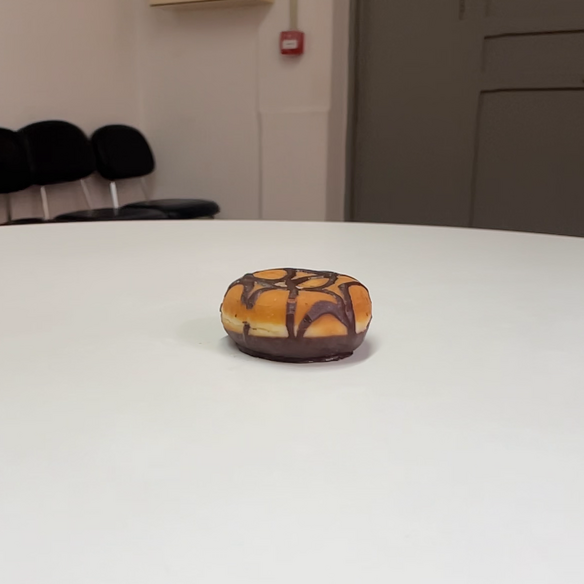}} &
        \subcaptionbox{\centering Chocolate Panettone}{\includegraphics[clip,width=0.13\linewidth]{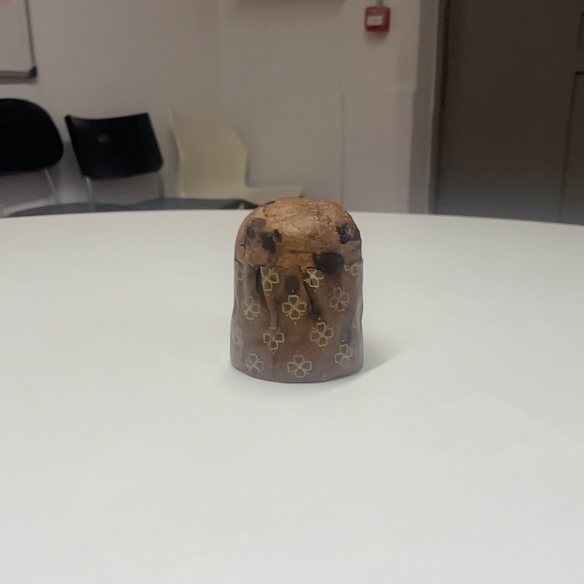}} &
        \subcaptionbox{\centering French Bread}{\includegraphics[clip,width=0.13\linewidth]{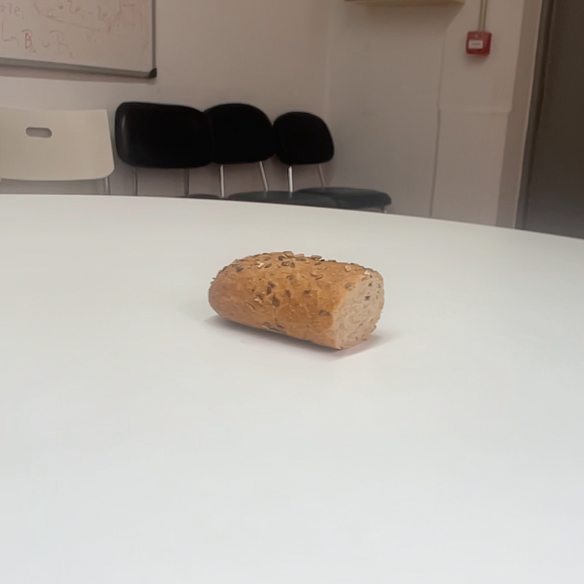}} &
        \subcaptionbox{\centering Beer Yellow Cane}{\includegraphics[clip,width=0.13\linewidth]{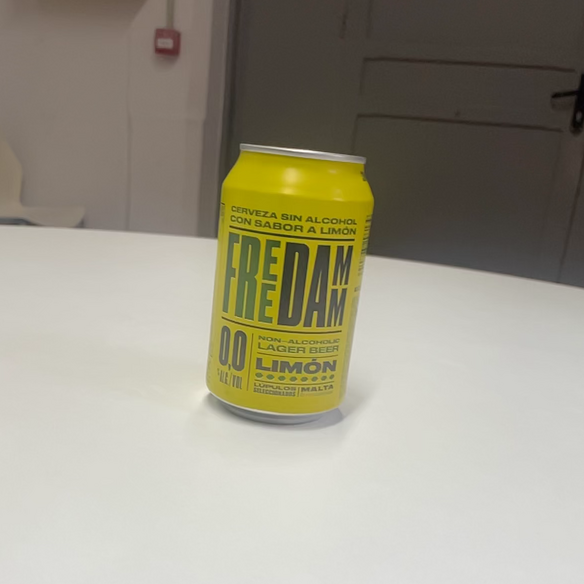}}
    \end{tabular}
    \caption{Foodkit dataset visual examples and their labels.}
    \label{fig:VolE_dataset}
\end{figure}

We introduce the foodkit dataset, a new and comprehensive dataset designed for estimating food volume. This dataset was created to address the limitations of existing datasets, which often include only a limited number of scenes, depend on reference information or CAD models, or have fixed capturing environments \cite{thames2021nutrition5k}. We aimed to develop a dataset that closely resembles free-motion real-world scenarios, allowing for volume estimation using smartphone cameras without needing reference objects. The Foodkit dataset comprises 21 diverse food objects, carefully selected to encompass a wide range of shapes, sizes, and textures commonly encountered in dietary assessment applications, as shown in Fig.~\ref{fig:VolE_dataset}. Each food item in the dataset is accompanied by ground truth measurements of volume and mass, providing a robust foundation for validating and benchmarking volume estimation techniques.

\begin{figure}[htb]
    \centering
    \includegraphics[width=0.7\linewidth]{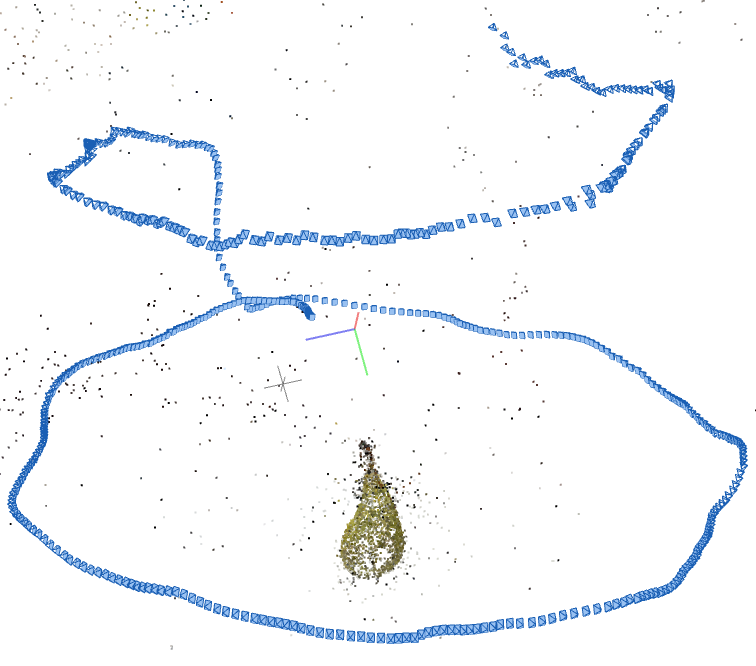}
     \caption{Visualization of the captured scene for a pear from the Foodkit dataset. The central pear is surrounded by 360 camera locations, represented as small camera icons, demonstrating comprehensive scene capture.}
    \label{fig:VolE_dataset_vc}
\end{figure}

To overcome the limitations of existing datasets, we developed an innovative approach combining AR technology with traditional 3D reconstruction techniques. The data collection process involved capturing each food object in a 360-degree view using an ARKit-based mobile application. This application captured dense sets of images extracted from the video along with ARKit-estimated 3D coordinates for each frame, providing crucial spatial information, as shown in Fig.~\ref{fig:VolE_dataset_vc}. Additionally, segmented masks were created for each image to isolate the food object from the background using FoodMem~\cite{almughrabi2024foodmem}. A key advantage of using ARKit-based capturing is that it eliminates the need for reference objects to scale the reconstructed model to real-world dimensions. The 3D coordinates provided by ARKit can be used to accurately refine the camera poses and subsequently scale both the camera poses and the output 3D reconstruction to real-world measurements. For each food item, we obtained precise ground truth measurements using the water displacement method \cite{displacement} for volume and a weighing scale for mass. To ensure volume accuracy, we performed the water displacement method five times and averaged the results. The water was at room temperature, and the error margin of ±5 mL comes from the scaler. The final Foodkit dataset for each of the 21 food objects includes: (a) Food video scene; (b) image sets sampled from the video; (c) ARKit-estimated 3D coordinates per image; (d) masks per image; (e) metadata such as ground truth volume and mass measurements, as shown in Table~\ref{tab:food_measurements}.

Our comprehensive dataset provides researchers with a valuable resource for developing and evaluating food volume estimation techniques, particularly those leveraging AR and computer vision technologies. By addressing the limitations of existing datasets and closely mimicking real-world capturing scenarios, our dataset contributes significantly to the advancement of 3D food volume estimation research.

\begin{table}[h]
\centering
\caption{Mass and volume measurements of various food items. To ensure the quality of the volumes, we repeated the water displacement method 5 times, and we took the average. The water temperature is room temperature. The error margin is ±5 ml.}
\begin{tabular}{|l|c|c|}
\hline
Item & Mass (g) & Volume (ml±5) \\
\hline
Apple & 150 & 175 \\
Orange & 250 & 200 \\
Aguacate & 125 & 85 \\
Lemon & 22.5 & 140 \\
Donut & 175 & 245 \\
Durrum & 200 & 200 \\
Pear & 250 & 170 \\
Chocolate Cake & 175 & 195 \\
Chocolate Croissant & 150 & 275 \\
Samosa & 150 & 145 \\
Apple Pie & 100 & 135 \\
Chocolate Bomb & 90 & 200 \\
Empanadilla & 110 & 95 \\
Falafel & 75 & 48 \\
French Bread & 75 & 163 \\
Paxoco Mini & 75 & 150 \\
Napolitanas & 175 & 233 \\
Capsicum & 250 & 320 \\
Mini Chocolate Panettone & 150 & 293 \\
Banana & 250 & 150 \\
Beer Yellow Cane & 375 & 350 \\
\hline
\end{tabular}

\label{tab:food_measurements}
\end{table}

\subsubsection{MTF Dataset}

\begin{figure}[htb]
    \centering
    \captionsetup[subfigure]{labelformat=empty, font=tiny}
    \setlength{\tabcolsep}{1pt}
    \begin{tabular}{ccccccc}
        \subcaptionbox{\centering Strawberry}{\includegraphics[clip,width=0.13\linewidth]{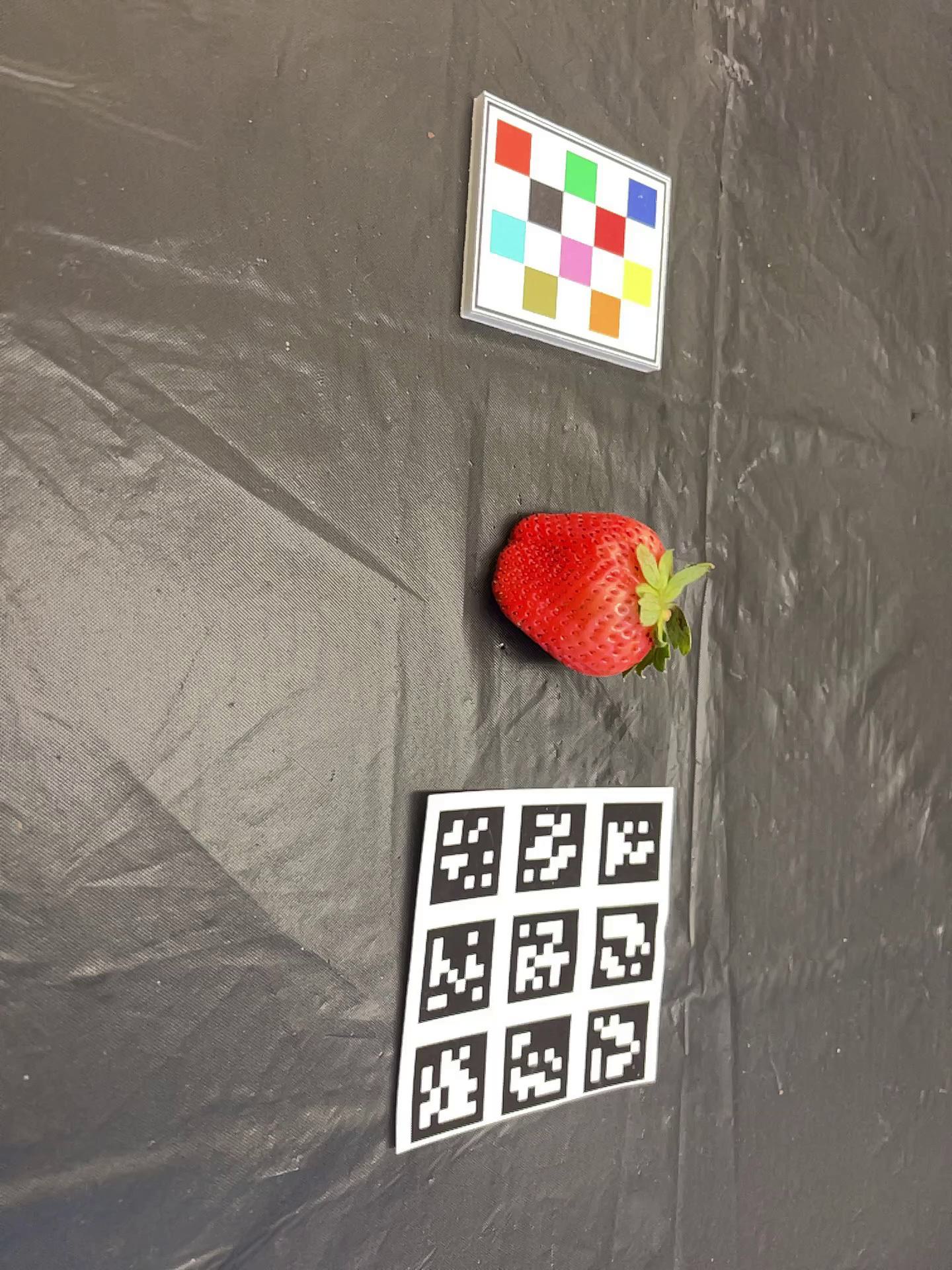}} &
        \subcaptionbox{\centering Cinnamon Bun}{\includegraphics[clip,width=0.13\linewidth]{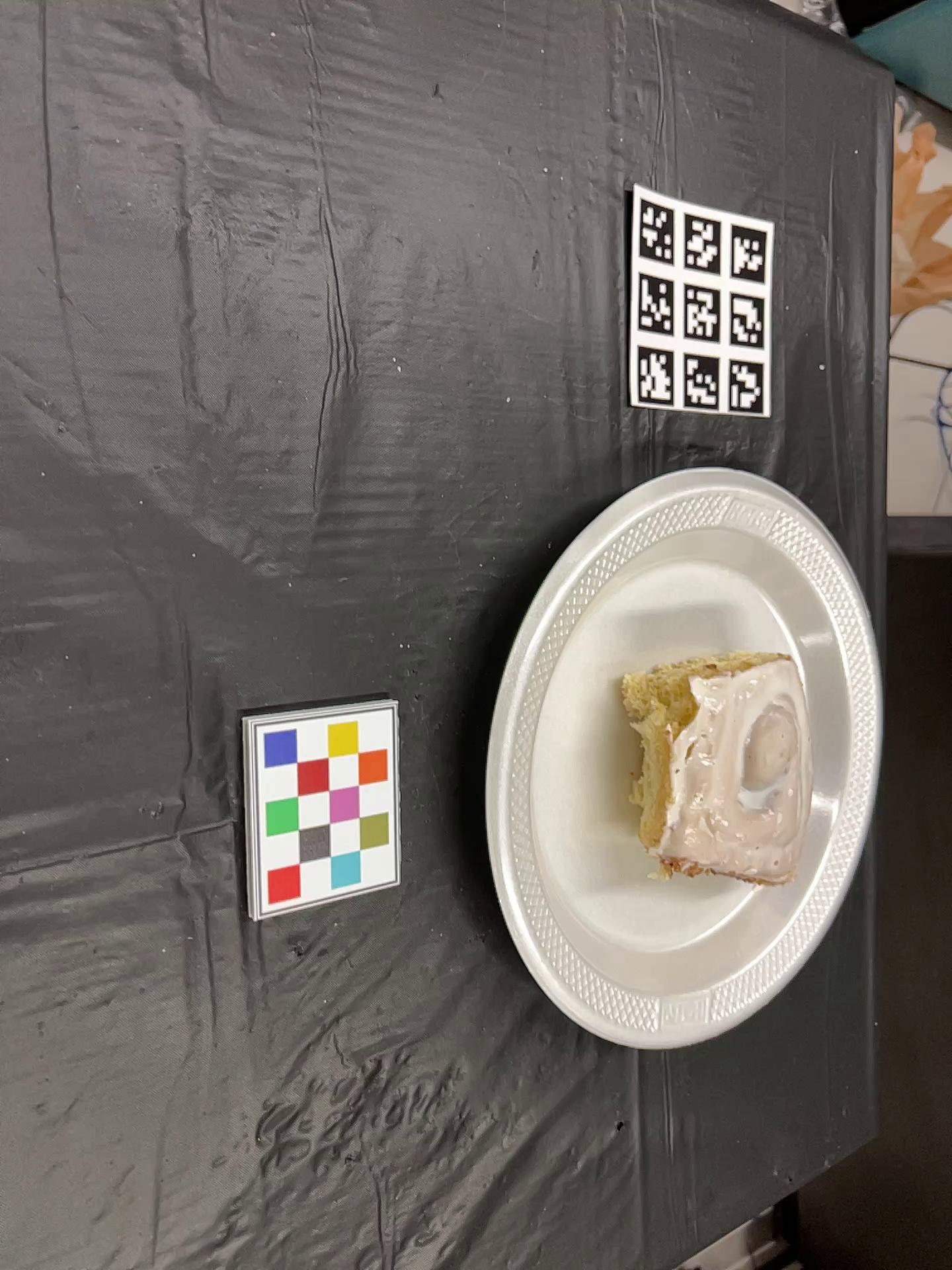}} &
        \subcaptionbox{\centering Pork Rib}{\includegraphics[clip,width=0.13\linewidth]{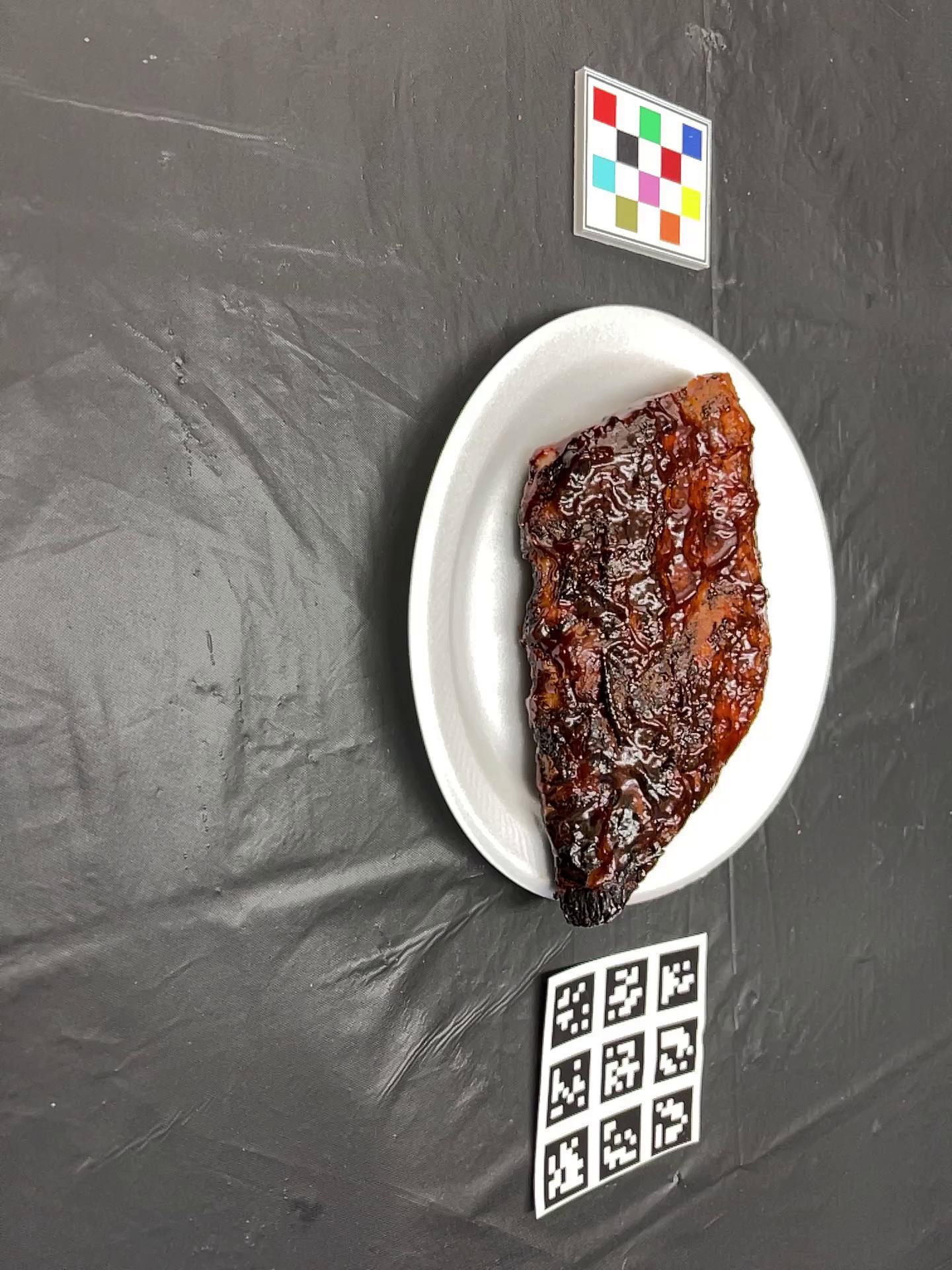}} &
        \subcaptionbox{\centering Corn}{\includegraphics[clip,width=0.13\linewidth]{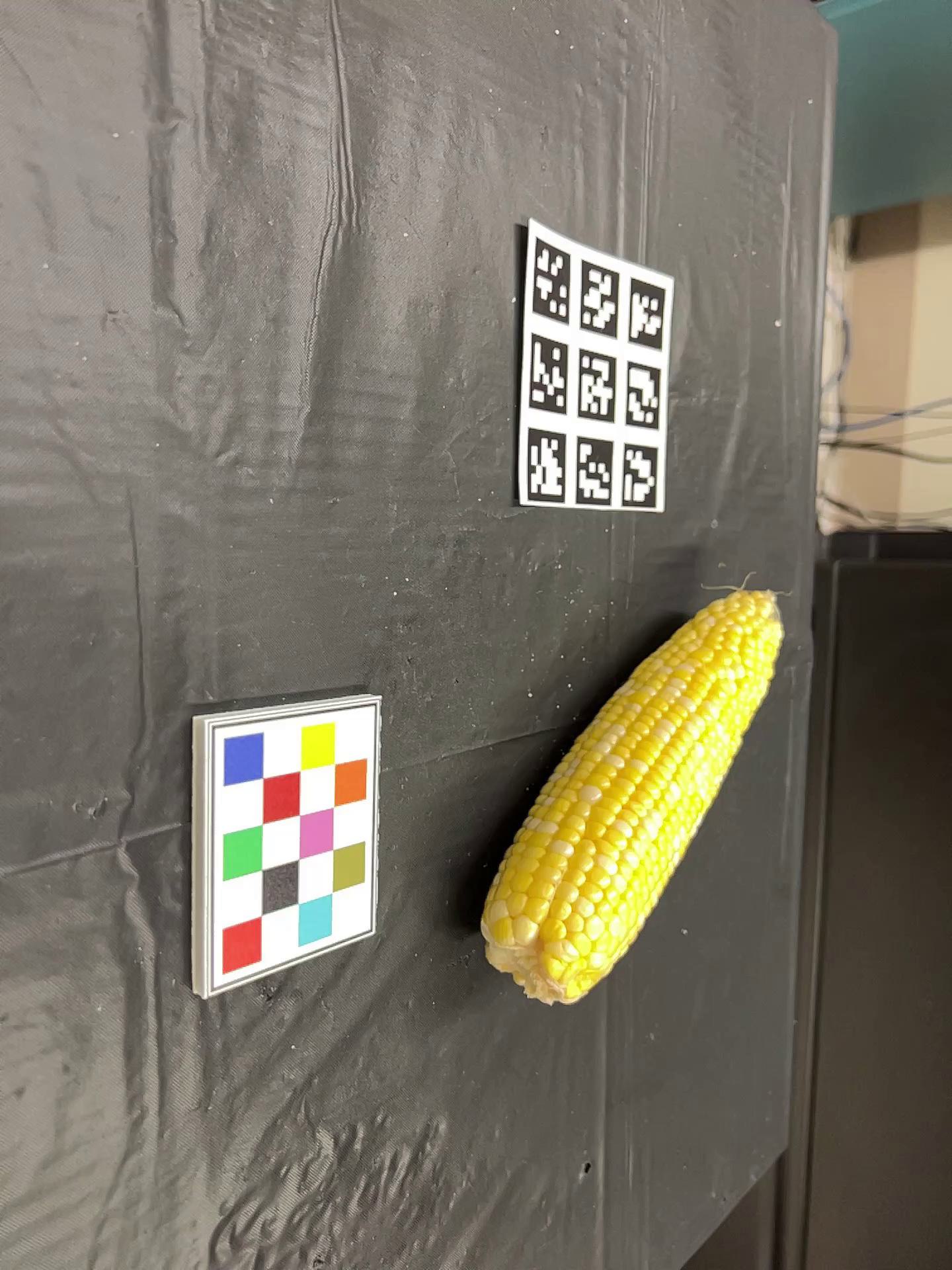}} &
        \subcaptionbox{\centering French Toast}{\includegraphics[clip,width=0.13\linewidth]{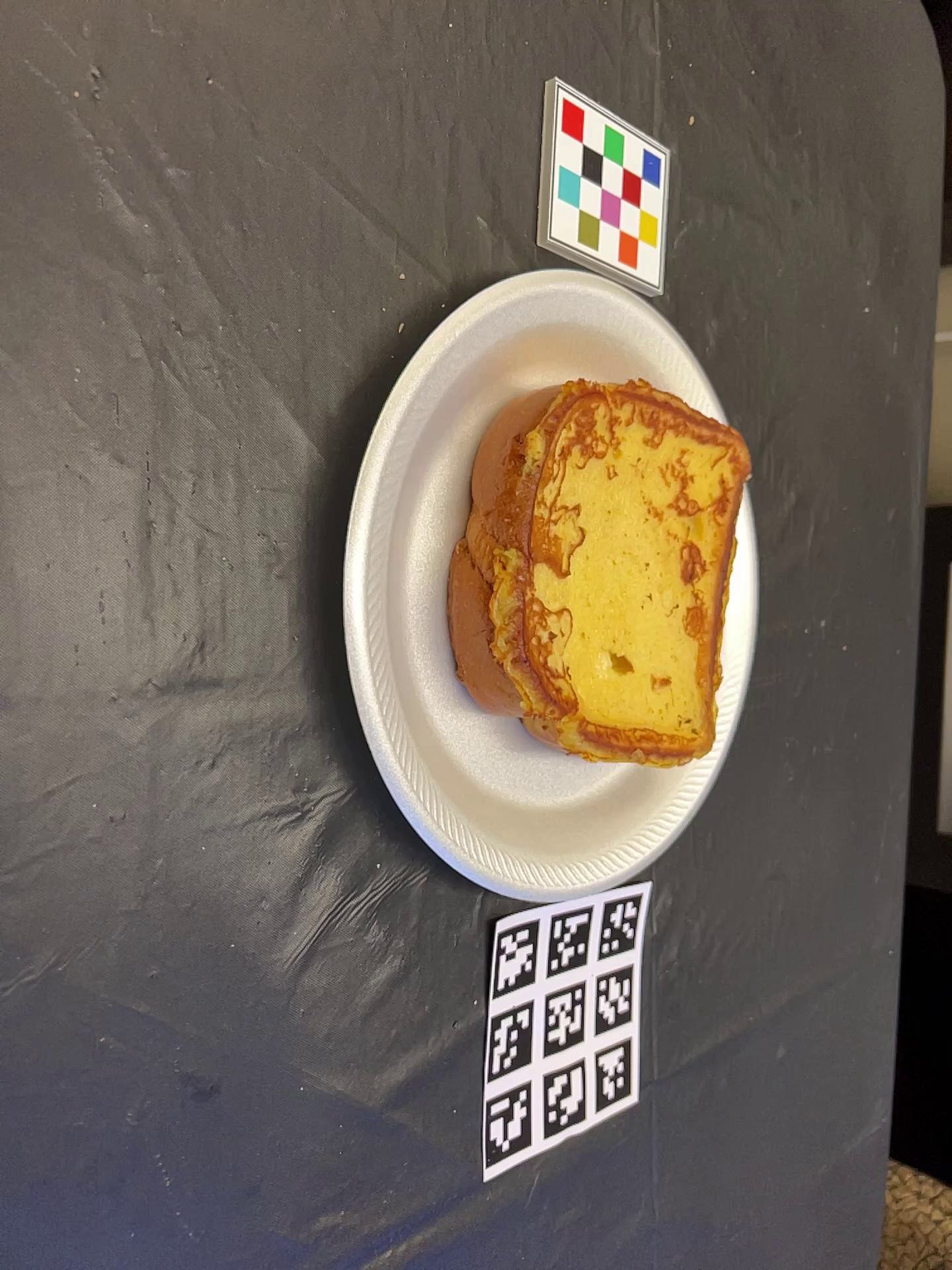}} &
        \subcaptionbox{\centering Sandwich}{\includegraphics[clip,width=0.13\linewidth]{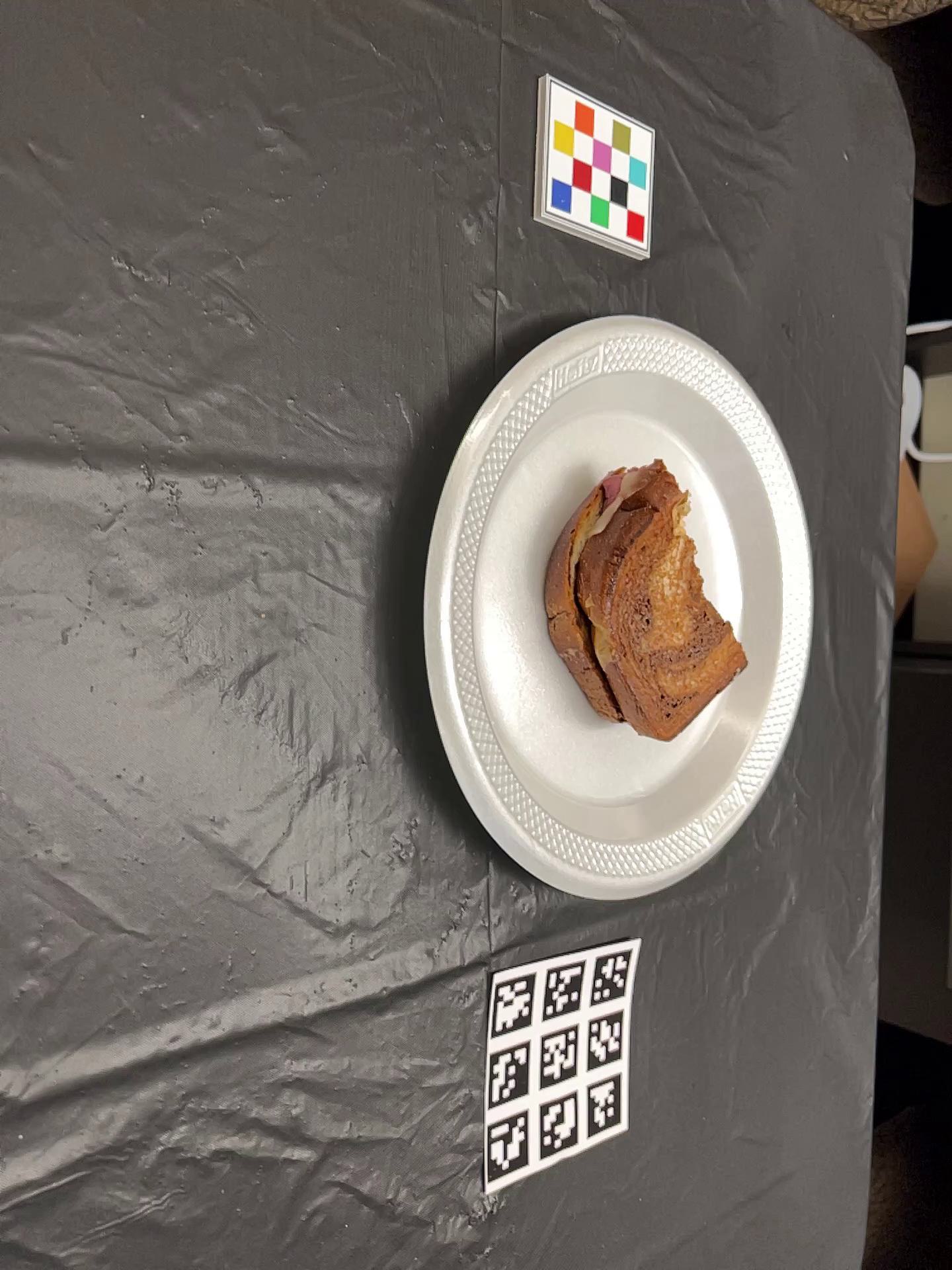}} &
        \subcaptionbox{\centering Burger}{\includegraphics[clip,width=0.13\linewidth]{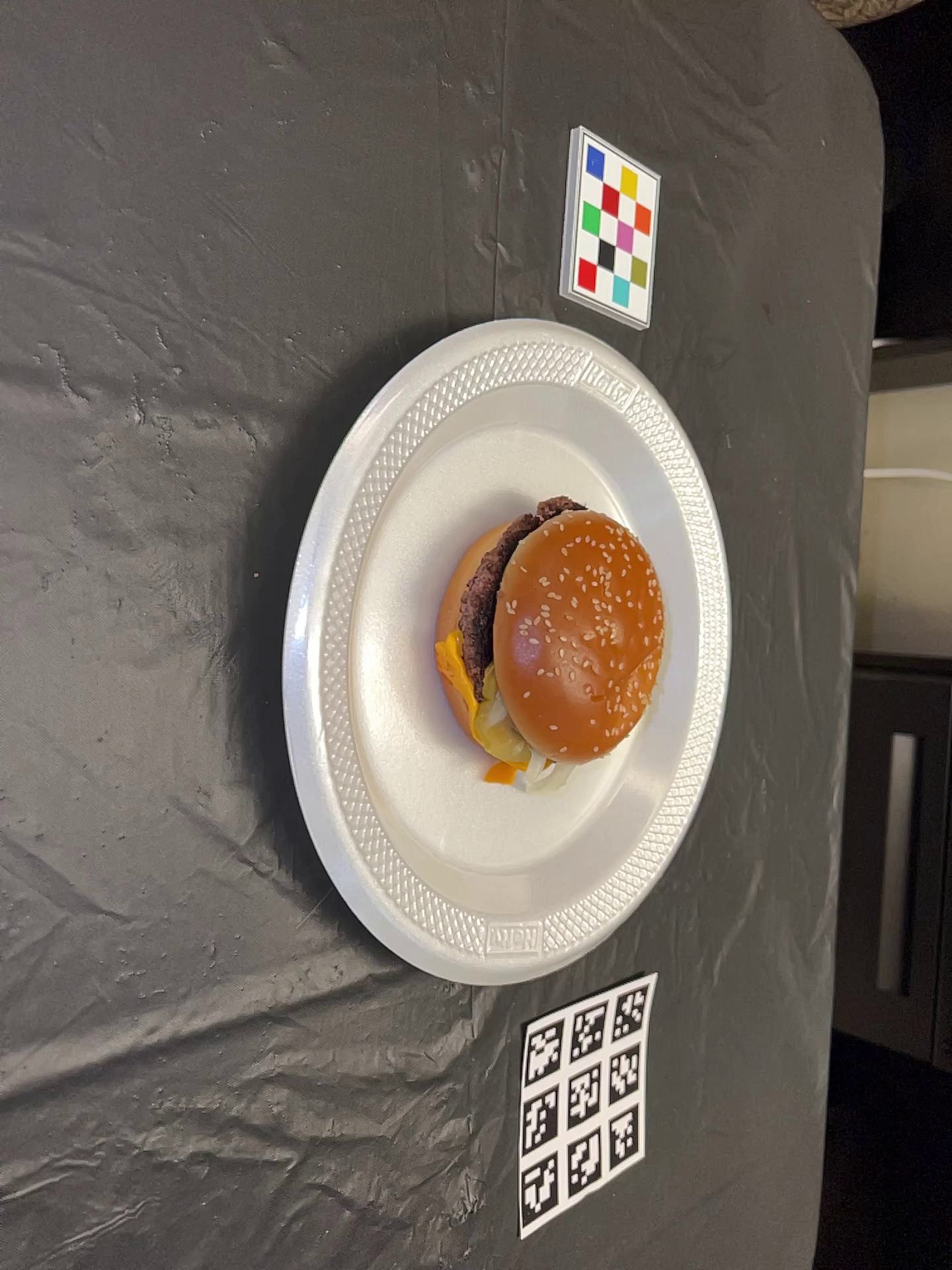}}
    \end{tabular}
    \\
    \begin{tabular}{cccccc}
        \subcaptionbox{\centering Cake}{\includegraphics[clip,width=0.13\linewidth]{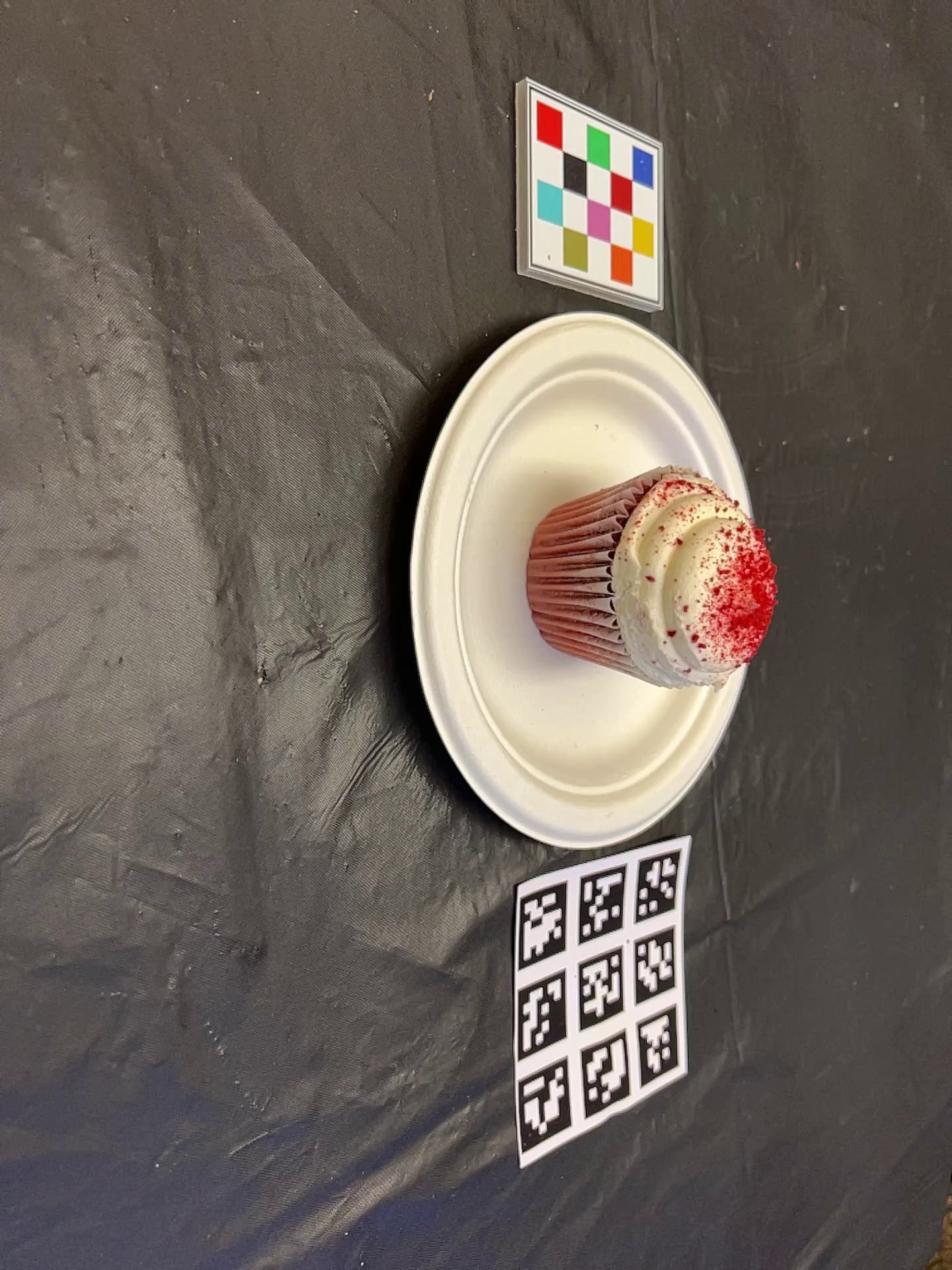}} &
        \subcaptionbox{\centering Blueberry Muffin}{\includegraphics[clip,width=0.13\linewidth]{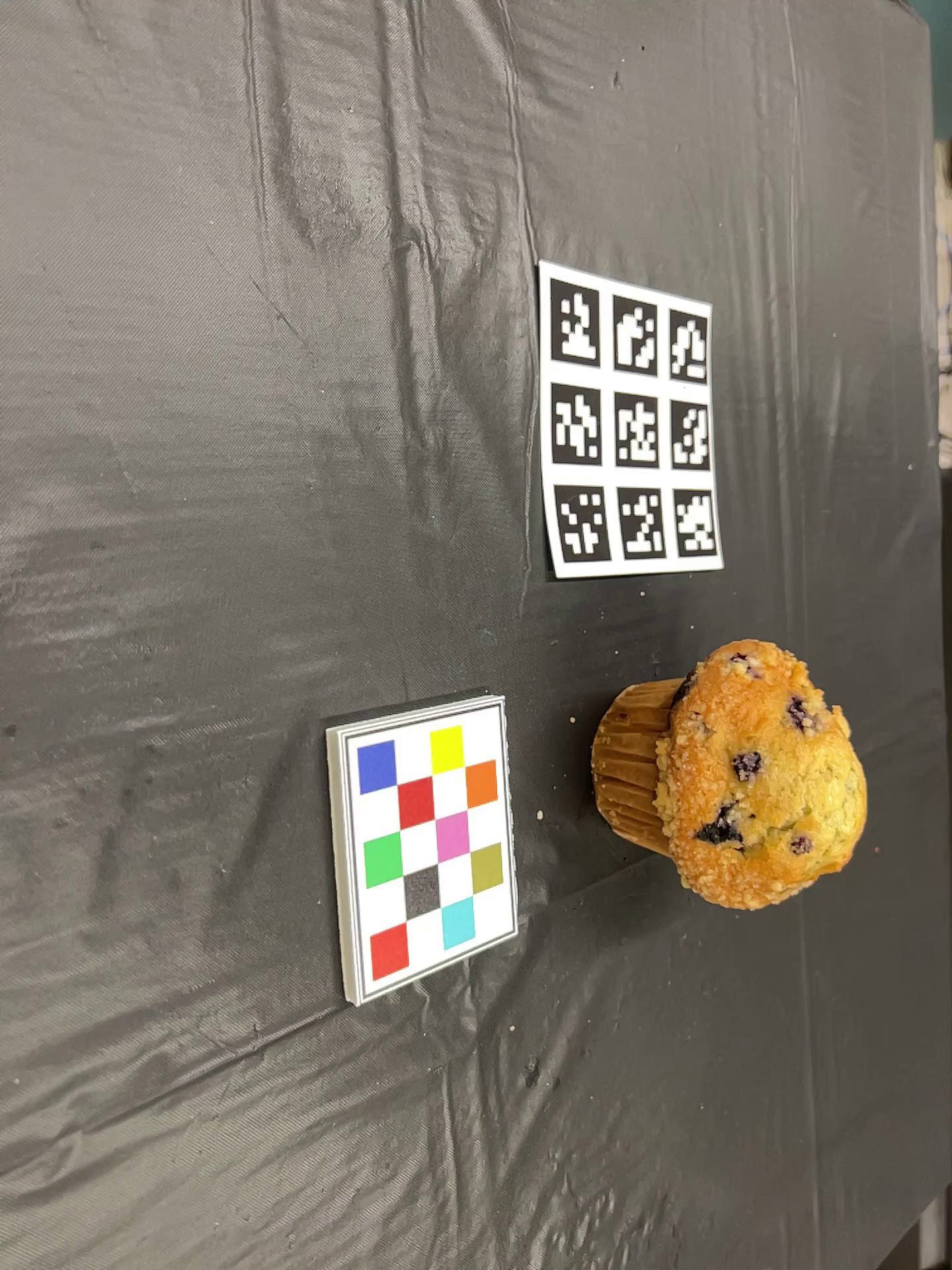}} &
        \subcaptionbox{\centering Banana}{\includegraphics[clip,width=0.13\linewidth]{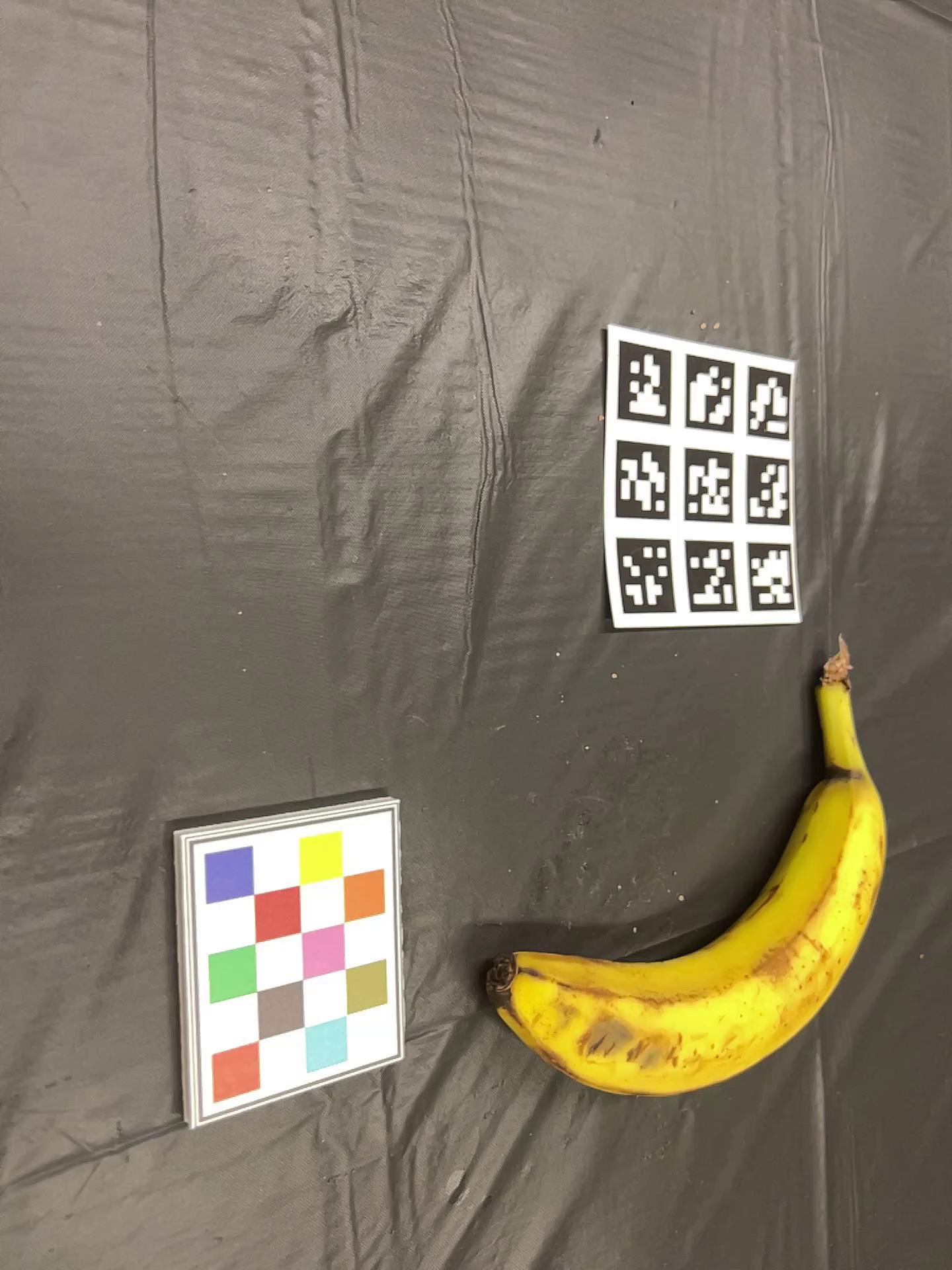}} &
        \subcaptionbox{\centering Salmon}{\includegraphics[clip,width=0.13\linewidth]{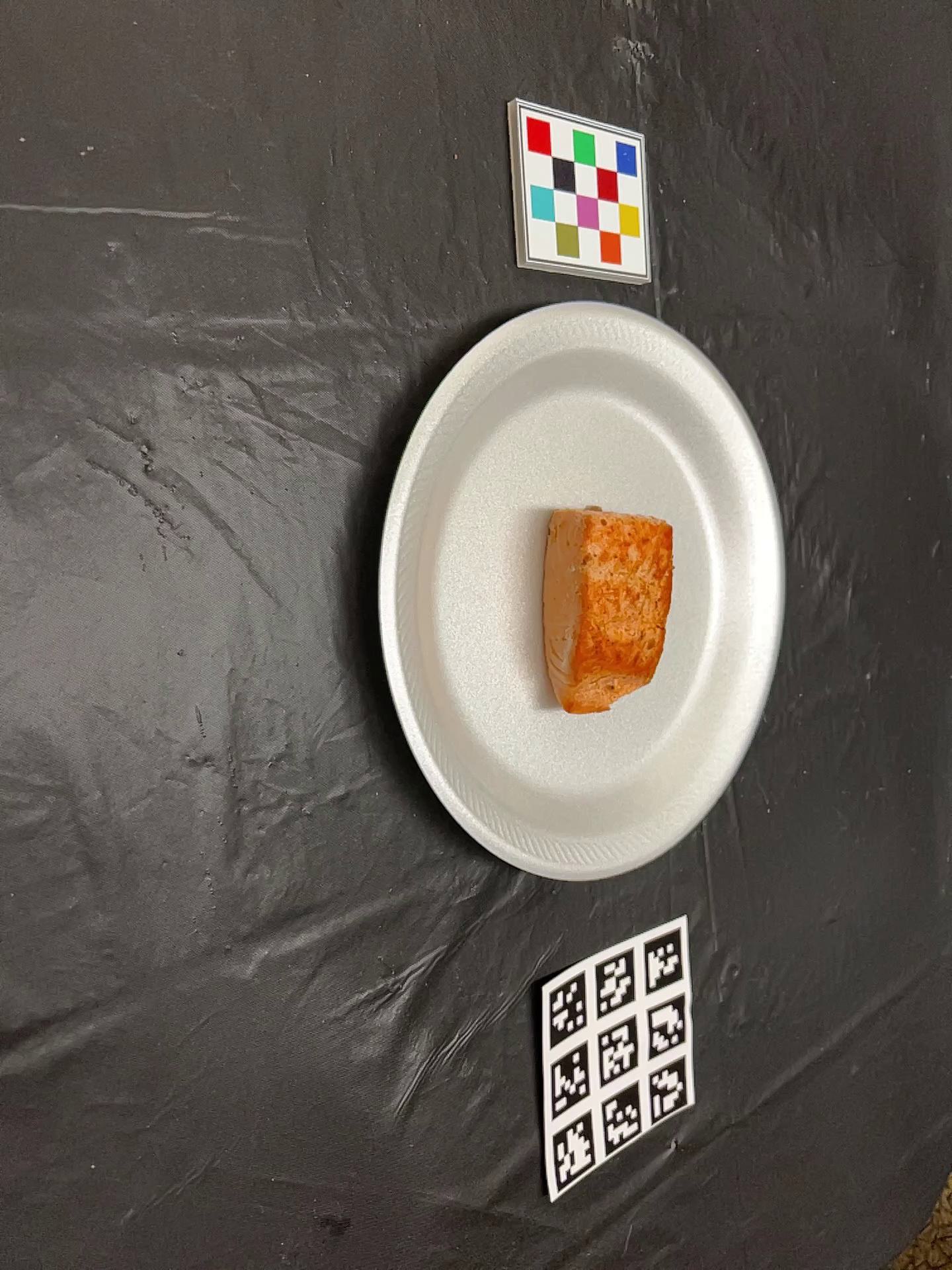}} &
        \subcaptionbox{\centering Burrito}{\includegraphics[clip,width=0.13\linewidth]{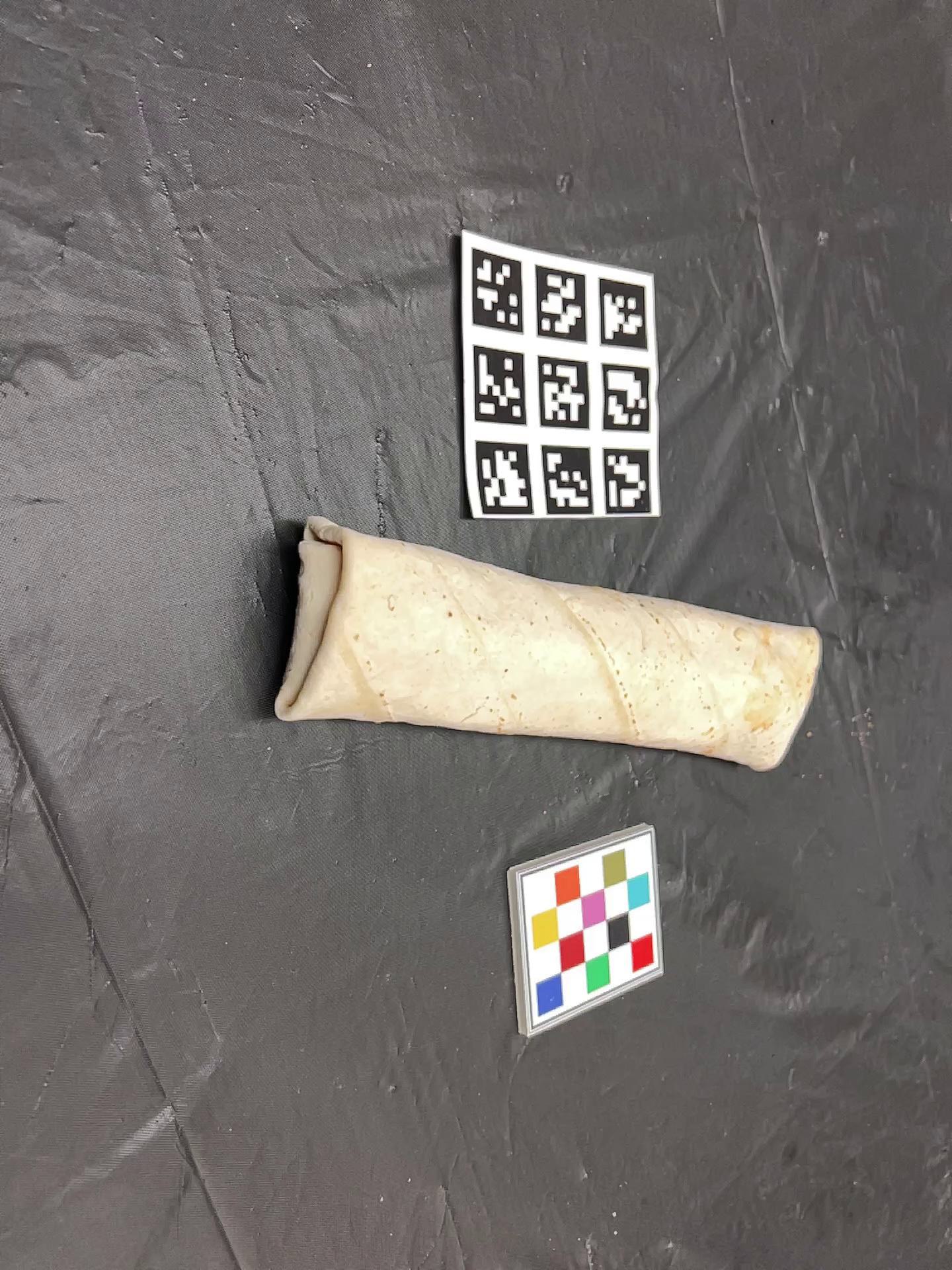}} &
        \subcaptionbox{\centering Hotdog}{\includegraphics[clip,width=0.13\linewidth]{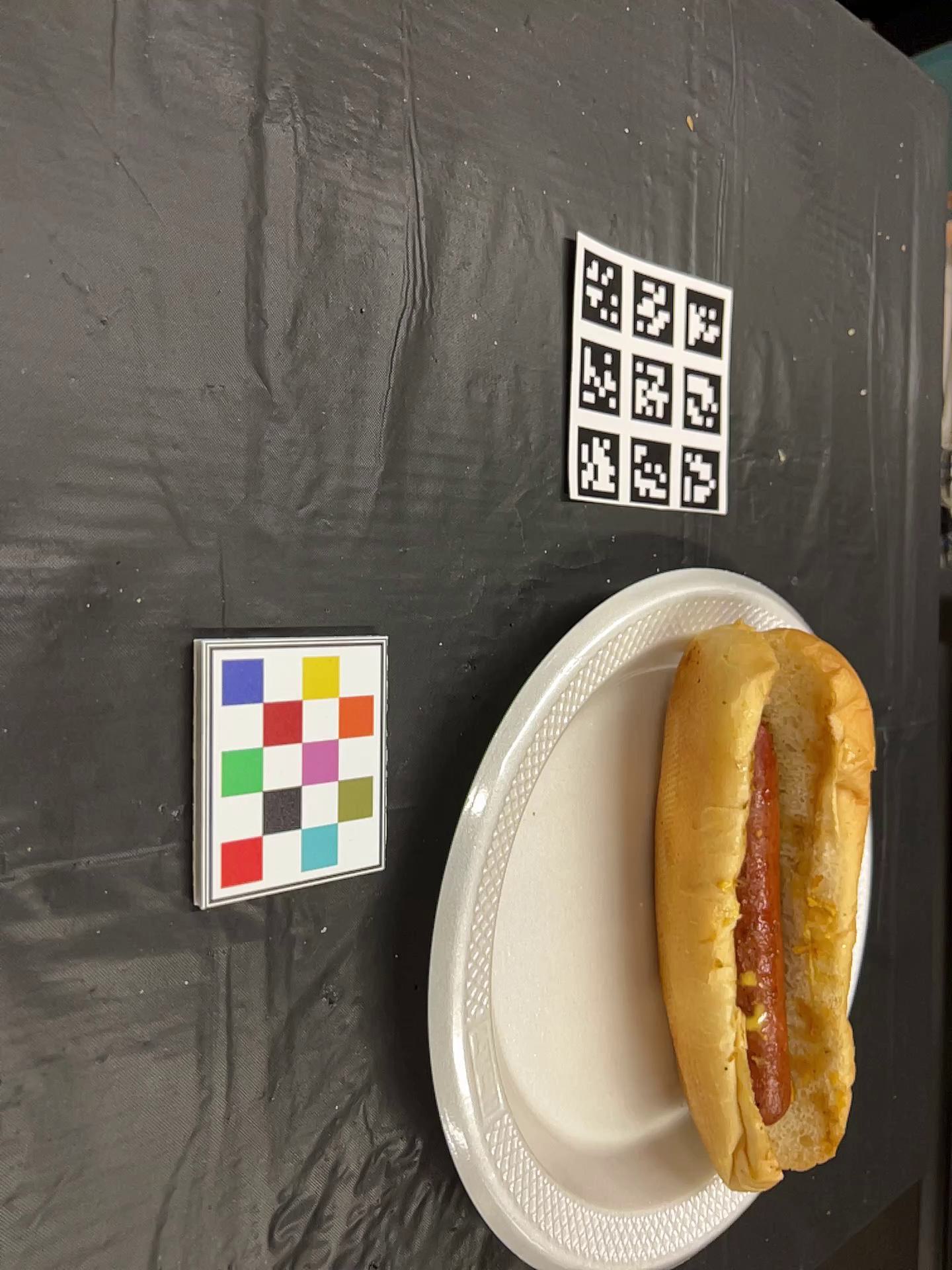}}
    \end{tabular}
    \caption{Sample images from the MTF dataset showcase a diverse range of food items from multiple angles across 13 categories. These include fruits (Strawberry, Banana), baked goods (Cinnamon Bun, French Toast, Blueberry Muffin, Cake), meats (Pork Rib, Salmon), and prepared dishes (Corn, Sandwich, Burger, Burrito, Hotdog), highlighting the variety in texture, color, and composition.}
    \label{fig:MTF_dataset}
\end{figure}

The MTF dataset~\cite{he2024metafood} is designed to evaluate food volume estimation techniques across varying levels of complexity. It comprises 20 food scenes strategically categorized into three difficulty tiers: easy, medium, and hard. The easy category encompasses the first eight scenes, each containing approximately 200 images, providing a rich set of multi-view data for each food item. The medium category consists of seven scenes, each offering around 30 images, presenting a more challenging scenario with fewer viewpoints. The hard category, representing the most demanding test cases, includes the remaining scenes, each containing only a single image. Each image in the dataset is accompanied by corresponding food masks and depth images, enhancing the dataset's utility for various computer vision tasks. The scenes are carefully constructed to include a single food object, a reference board (such as a chessboard), and QR code papers surrounding the object. This setup facilitates accurate scale estimation and provides contextual information for the food items. We focused on the \textbf{easy and medium scenes} for our evaluation, as our framework processes more than one image. Therefore, to evaluate our proposed framework on the MTF dataset, we must consider scenes with more than one image, as shown in Fig.~\ref{fig:MTF_dataset}. Additionally, we used the reference board to scale the reconstructed scenes to their original size.

\subsubsection{DTU Dataset}
\begin{figure}[htb]
    \centering
    \captionsetup[subfigure]{labelformat=empty, font=tiny}
    \setlength{\tabcolsep}{1pt}
    \begin{tabular}{ccccccc}
        \subcaptionbox{\centering Scan 24}{\includegraphics[clip,width=0.19\linewidth]{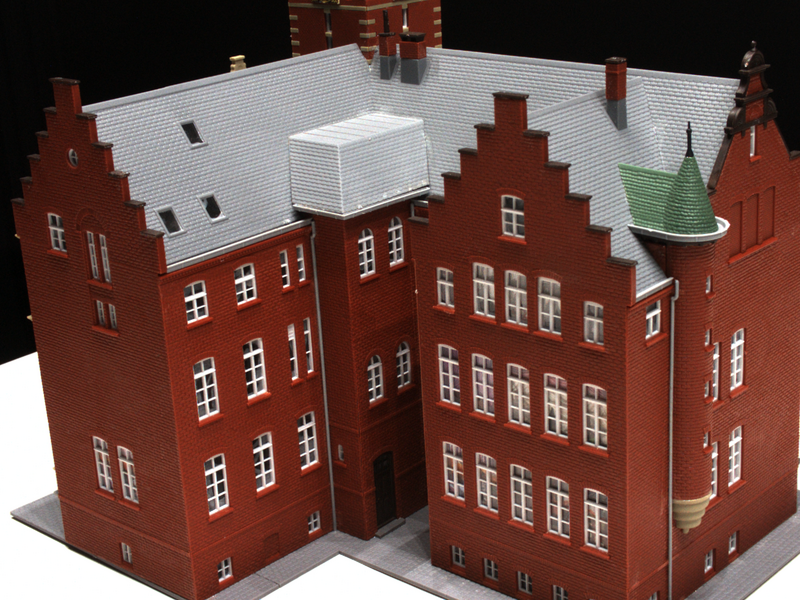}} &
        \subcaptionbox{\centering Scan 37}{\includegraphics[clip,width=0.19\linewidth]{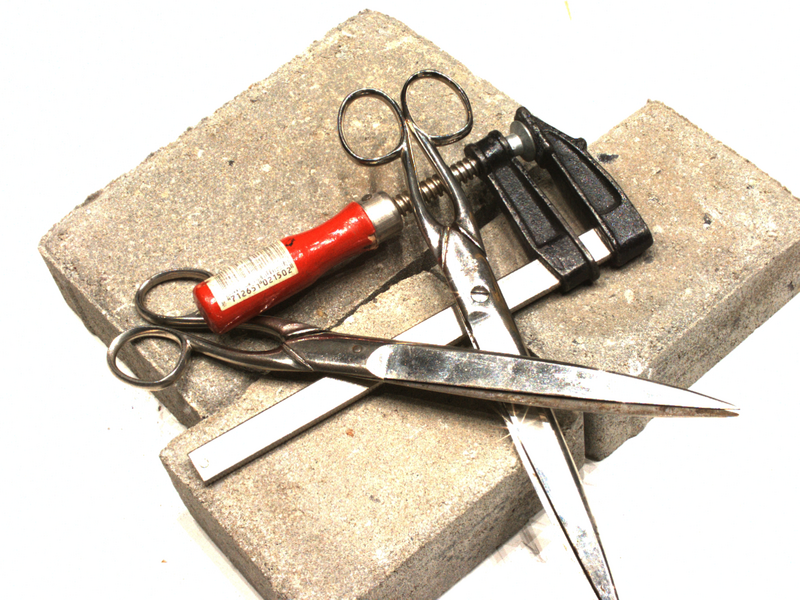}} &
        \subcaptionbox{\centering Scan 40}{\includegraphics[clip,width=0.19\linewidth]{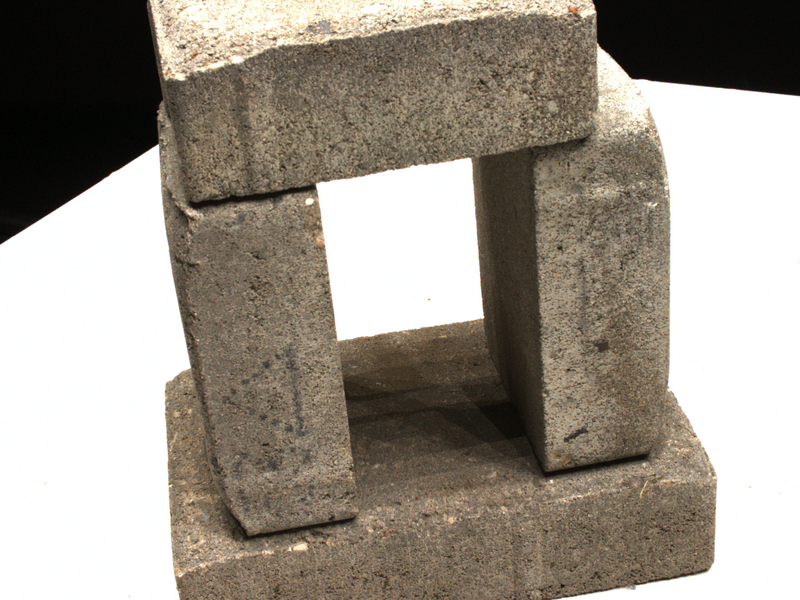}} &
        \subcaptionbox{\centering Scan 55}{\includegraphics[clip,width=0.19\linewidth]{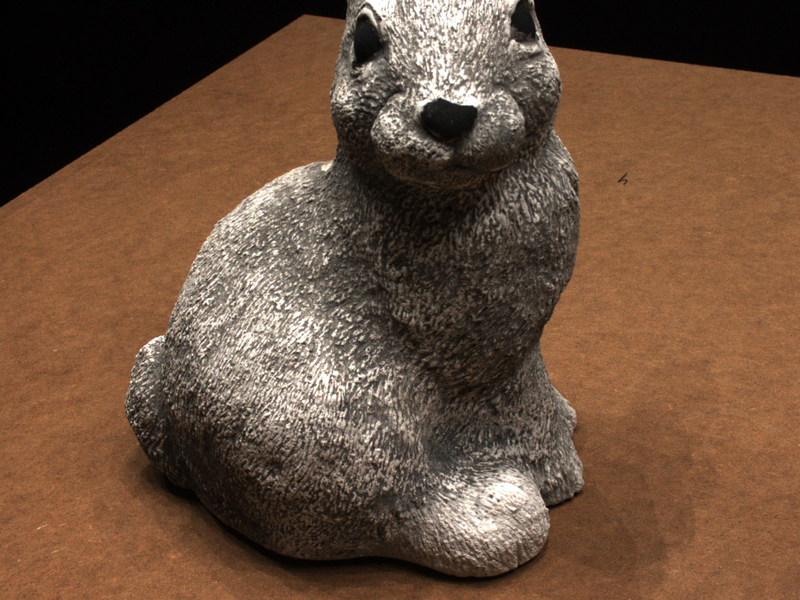}} &
        \subcaptionbox{\centering Scan 63}{\includegraphics[clip,width=0.19\linewidth]{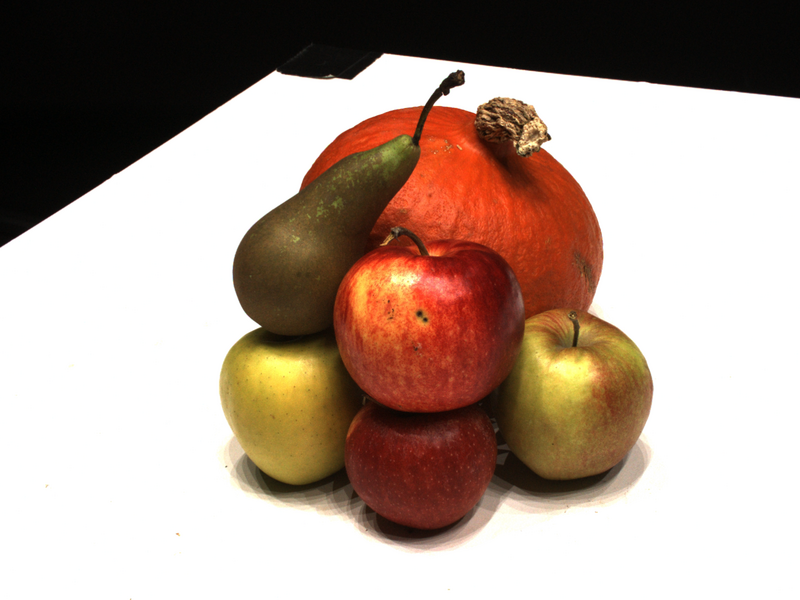}} 
        \\
        \subcaptionbox{\centering Scan 65}{\includegraphics[clip,width=0.19\linewidth]{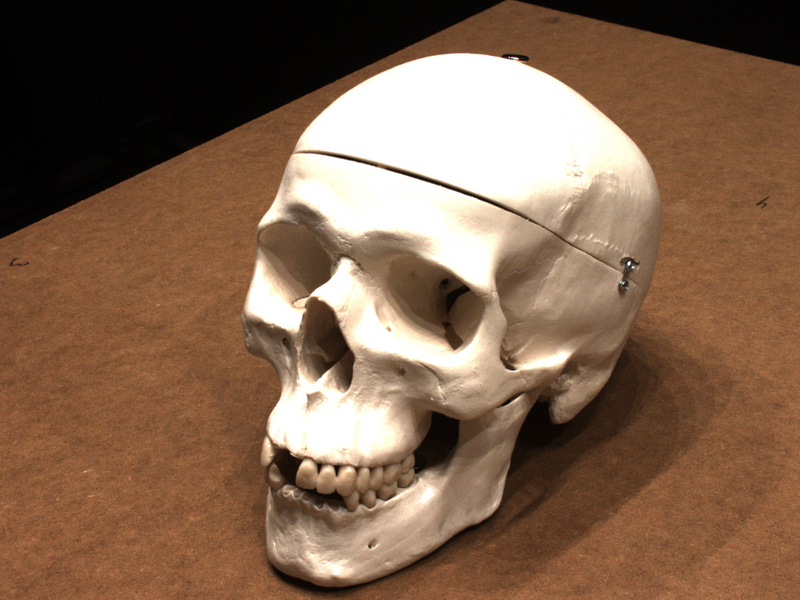}} &
        \subcaptionbox{\centering Scan 69}{\includegraphics[clip,width=0.19\linewidth]{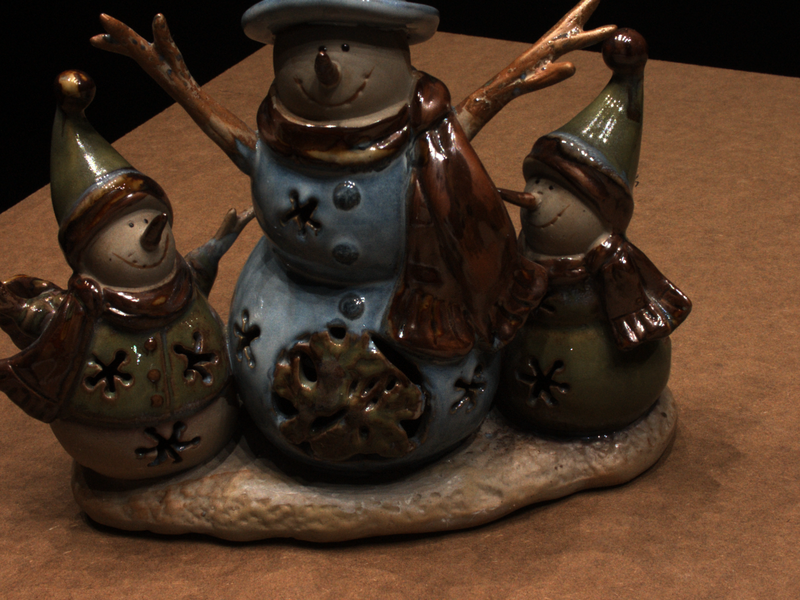}} &
        \subcaptionbox{\centering Scan 83}{\includegraphics[clip,width=0.19\linewidth]{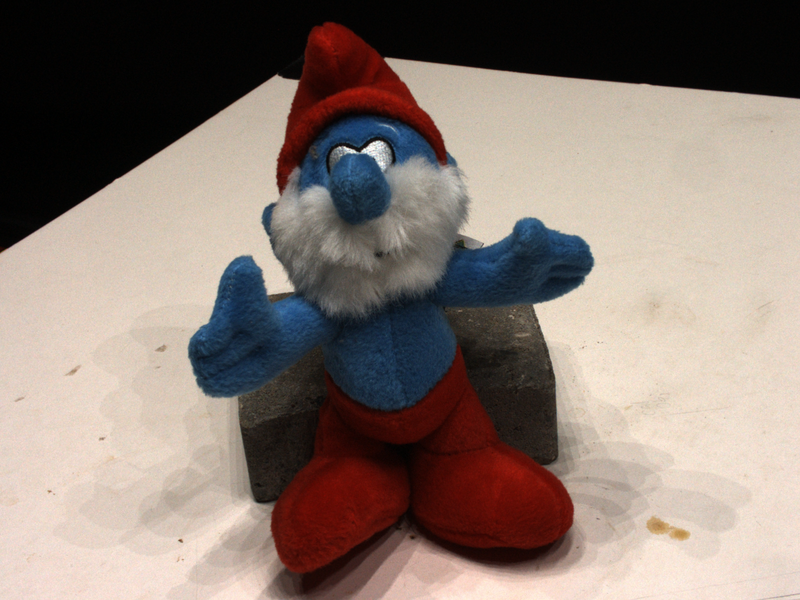}} &
        \subcaptionbox{\centering Scan 97}{\includegraphics[clip,width=0.19\linewidth]{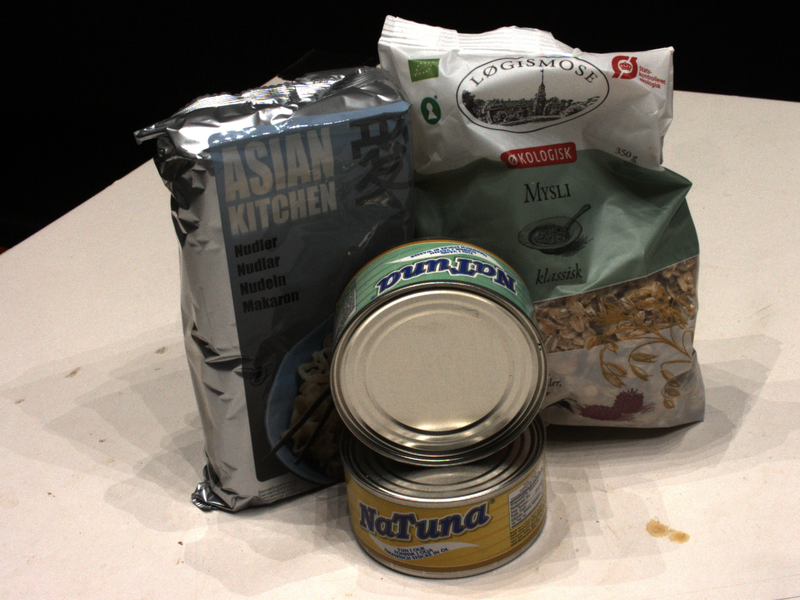}} &
        \subcaptionbox{\centering Scan 105}{\includegraphics[clip,width=0.19\linewidth]{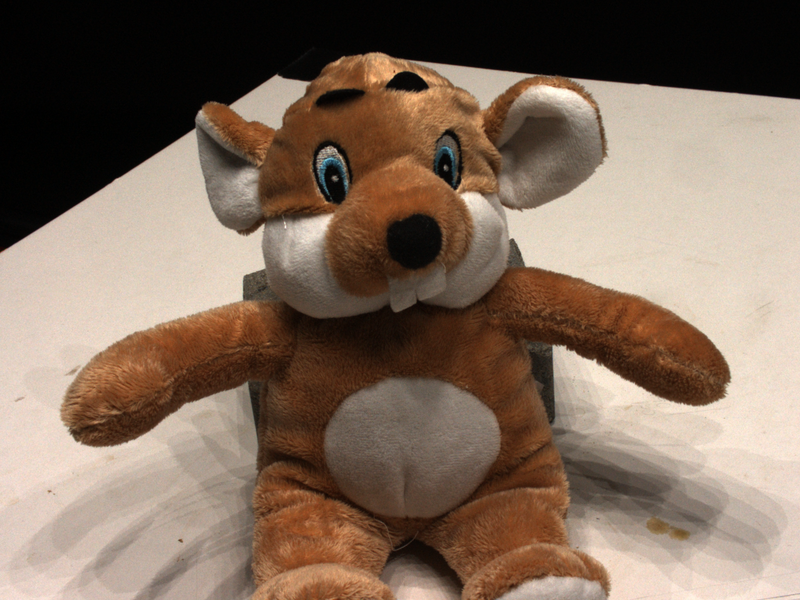}} 
        \\
        \subcaptionbox{\centering Scan 106}{\includegraphics[clip,width=0.19\linewidth]{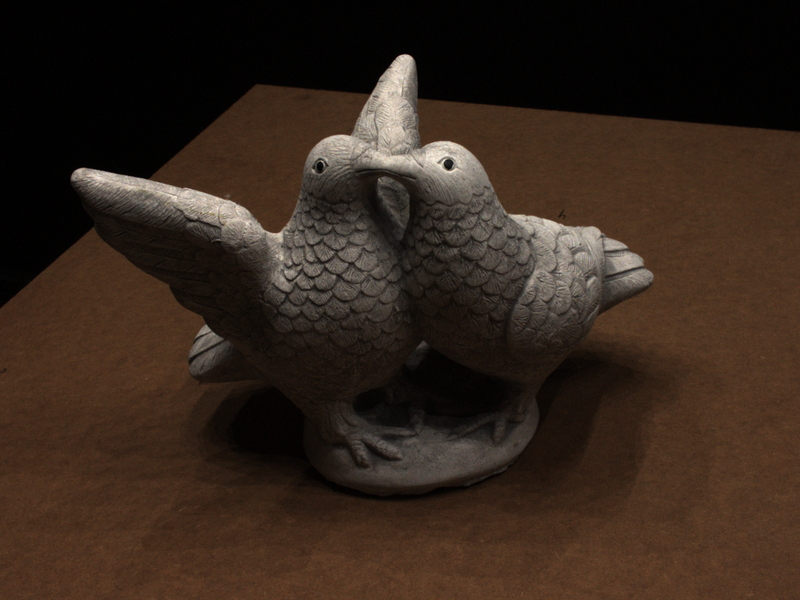}} &
        \subcaptionbox{\centering Scan 110}{\includegraphics[clip,width=0.19\linewidth]{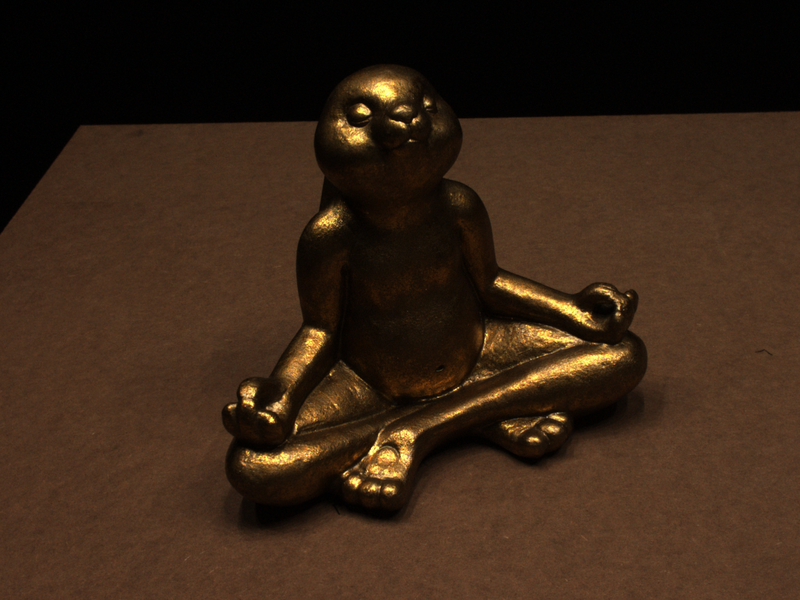}} &
        \subcaptionbox{\centering Scan 114}{\includegraphics[clip,width=0.19\linewidth]{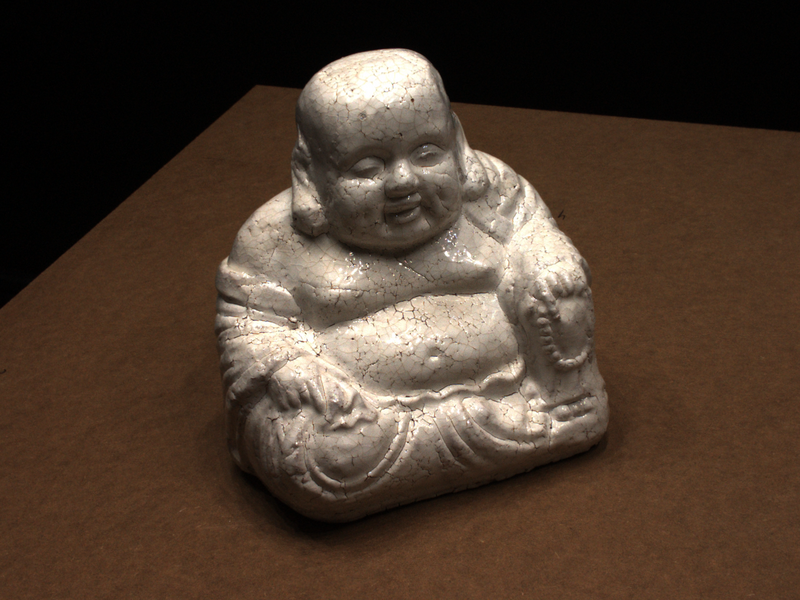}} &
        \subcaptionbox{\centering Scan 118}{\includegraphics[clip,width=0.19\linewidth]{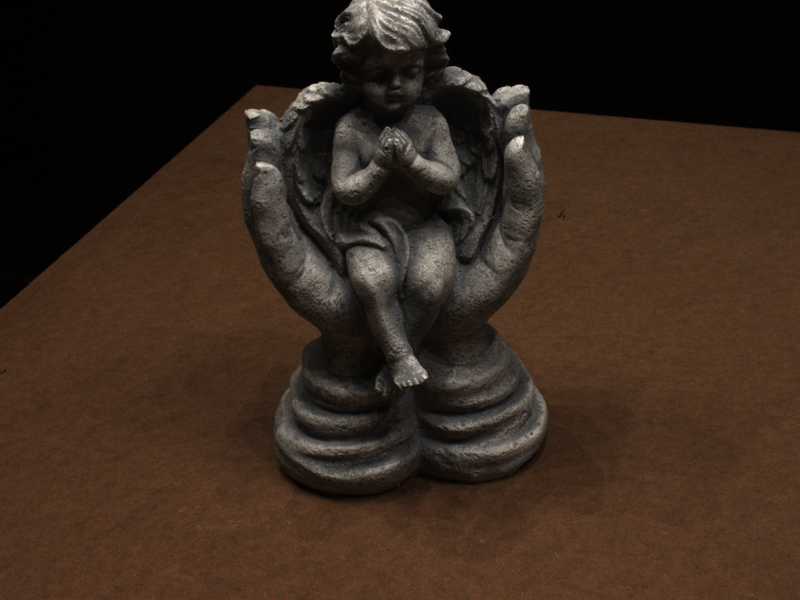}} &
        \subcaptionbox{\centering Scan 122}{\includegraphics[clip,width=0.19\linewidth]{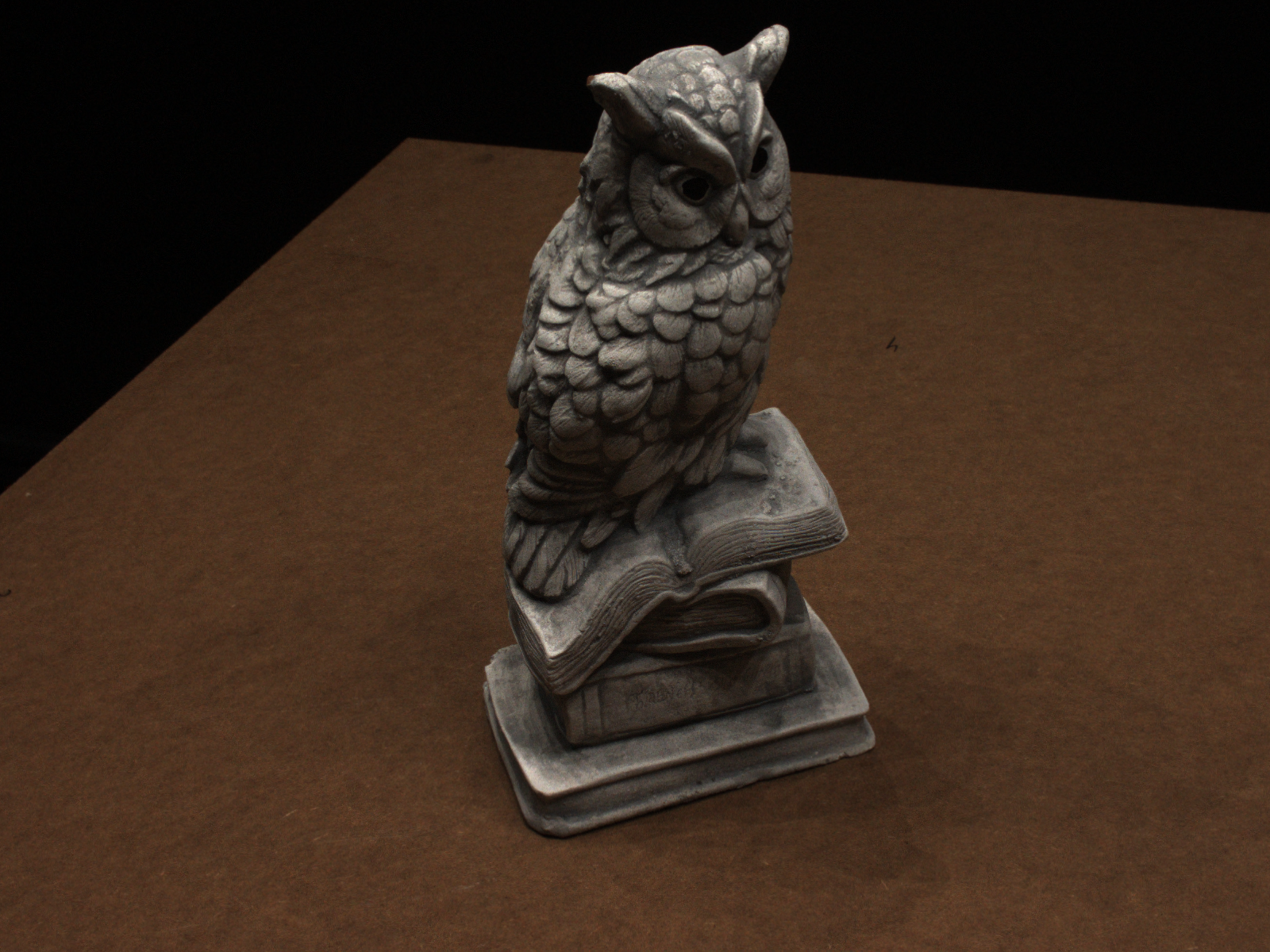}} 
    \end{tabular}
    \caption{Visualization of selected scans from the DTU dataset, which showcase diverse 3D models captured under controlled conditions using a robotic setup. Each subfigure represents a distinct scan (e.g., Scan 24, Scan 37, etc.), highlighting variations in geometry and structure.}
    \label{fig:DTU_dataset}
\end{figure}

The DTU dataset~\cite{jensen2014large} is a comprehensive benchmark for multi-view stereo reconstruction, consisting of 128 scenes captured in controlled laboratory settings. Each scene was scanned from 49 or 64 fixed camera positions under seven lighting conditions, using structured light scanners to generate ground truth point clouds. While not specifically designed for food items, the dataset offers various objects and scenes to evaluate 3D reconstruction methods. The official evaluation uses the Chamfer distance metric to assess point cloud accuracy and completeness. Recent studies have focused on a subset of 15 diverse scenes selected to represent the dataset's variety, with corresponding segmentation masks provided for each chosen scene. Therefore, to evaluate our proposed framework on the DTU dataset, we also selected the same subset of 15 diverse scenes used in recent studies. This approach allows for a fair comparison with state-of-the-art methods and comprehensively assesses our framework's performance across a representative dataset sample, as shown in Fig.~\ref{fig:DTU_dataset}.

\subsection{Comparative Analysis}

We evaluate our framework against state-of-the-art methods in volume estimation and 3D reconstruction, selecting them based on their relevance, recency, and reproducibility. The comparison includes COLMAP~\cite{schoenberger2016sfm}, NeuS~\cite{wang2021neus}, InstantNGP~\cite{mueller2022instant}, InstantNSR~\cite{zhao2022human}, NeuS2~\cite{neus2}, VolETA~\cite{almughrabi2024voleta}, ININ~\cite{he2024metafood}, and FoodR.~\cite{he2024metafood}, which represent a range of diverse approaches. This selection allows for a comprehensive assessment of VolE's performance across different aspects of the task. 

In this section, we provide a detailed qualitative and quantitative analysis to highlight the strengths and limitations of each method in the context of volume estimation and 3D reconstruction. The "Qualitative Results" section emphasizes the superior detail and geometric accuracy of the proposed VolE framework in reconstructing food items, illustrated through comparisons with other methods across various datasets. Simultaneously, the "Quantitative Results" section offers performance metrics, including MAPE and CD, showcasing VolE's effectiveness in volume estimation and 3D reconstruction accuracy compared to other state-of-the-art approaches across different food types. We aim to provide insights into their capabilities and trade-offs by systematically evaluating these methods.

\subsubsection{Qualitative Results}
\begin{figure*}[h]
    \centering
        \begin{subfigure}[b]{0.13\linewidth}
            \centering
                \includegraphics[clip,width=0.45\linewidth]{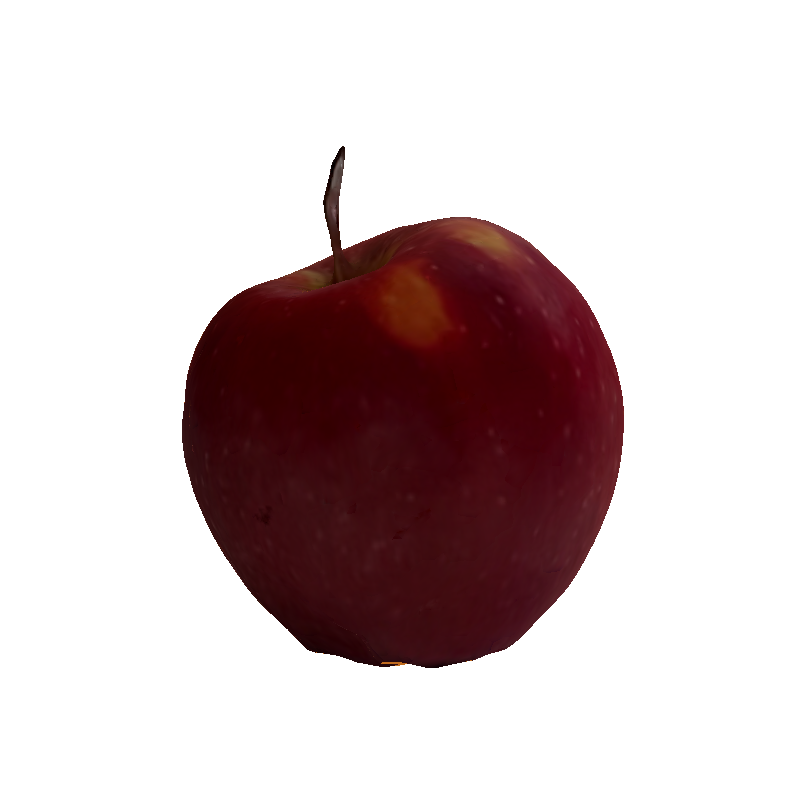}
                \includegraphics[clip,width=0.45\linewidth]{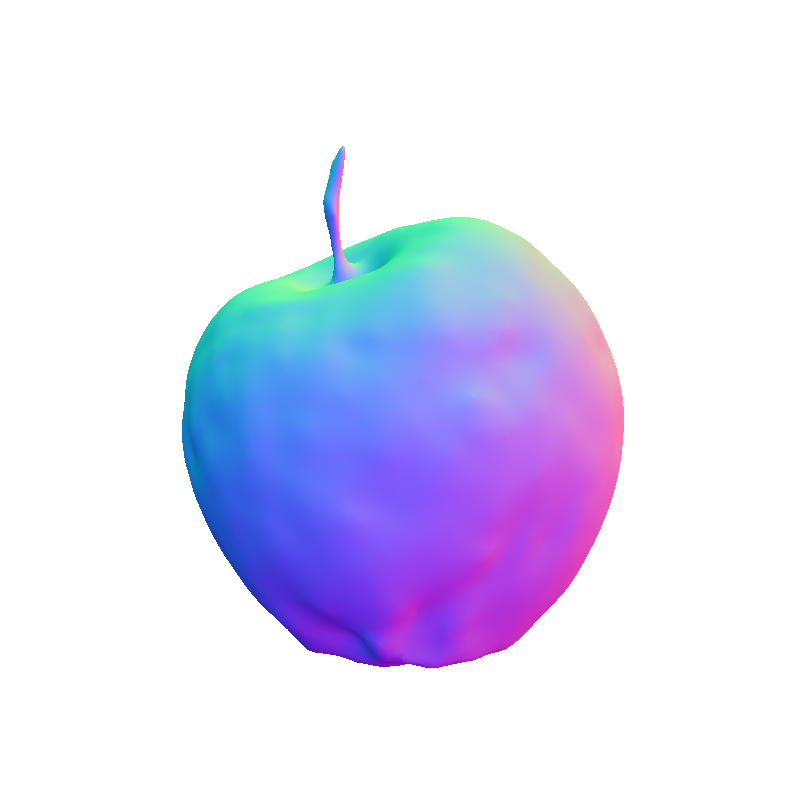}
            \caption{Apple}
        \end{subfigure}
        \begin{subfigure}[b]{0.13\linewidth}
            \centering
                \includegraphics[clip,width=0.45\linewidth]{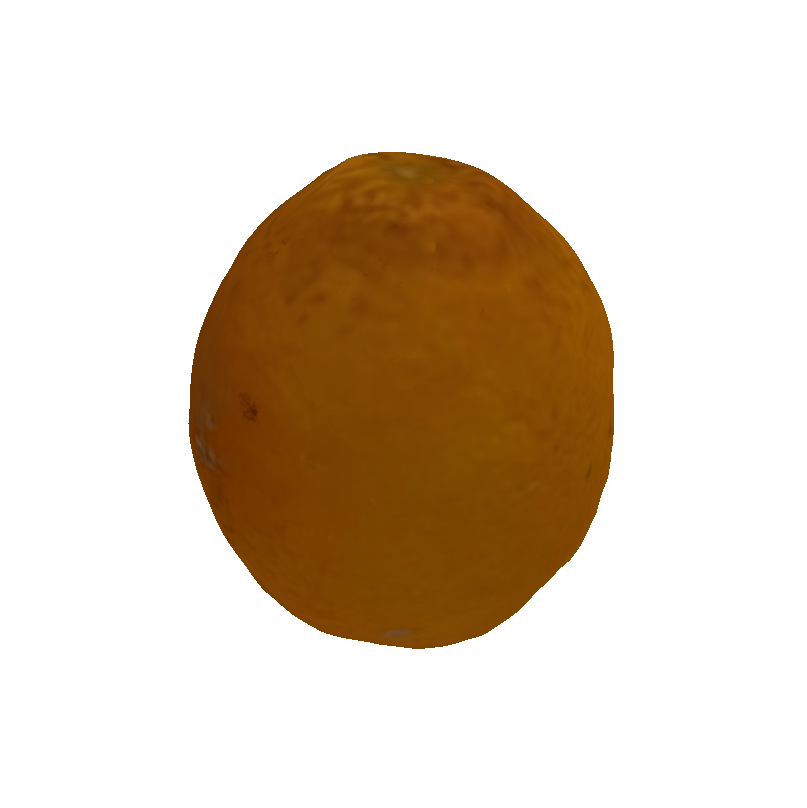}
                \includegraphics[clip,width=0.45\linewidth]{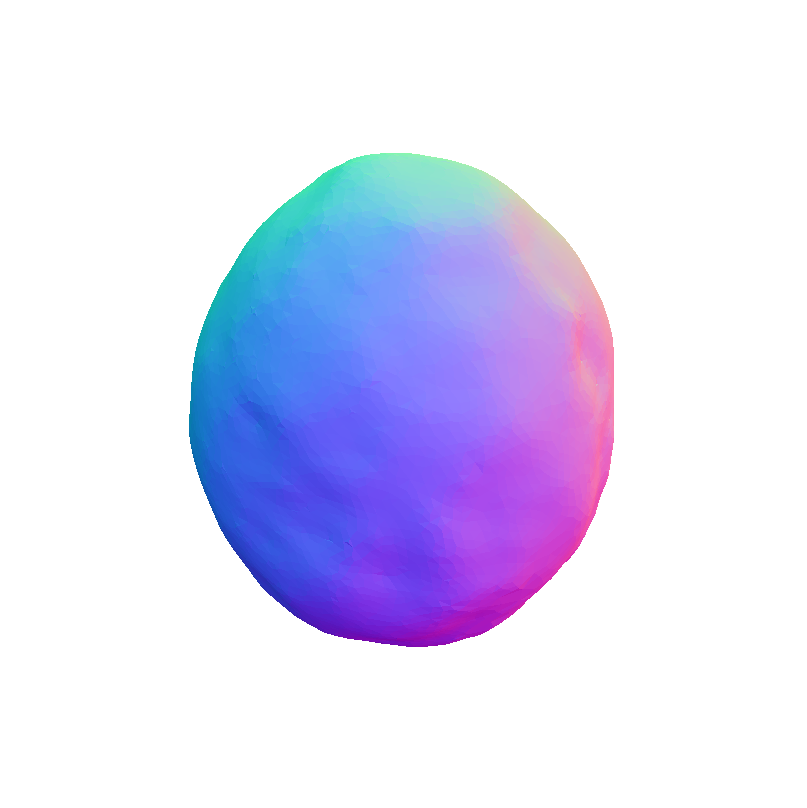}
            \caption{Orange}
        \end{subfigure}
        \begin{subfigure}[b]{0.13\linewidth}
            \centering
                \includegraphics[clip,width=0.45\linewidth]{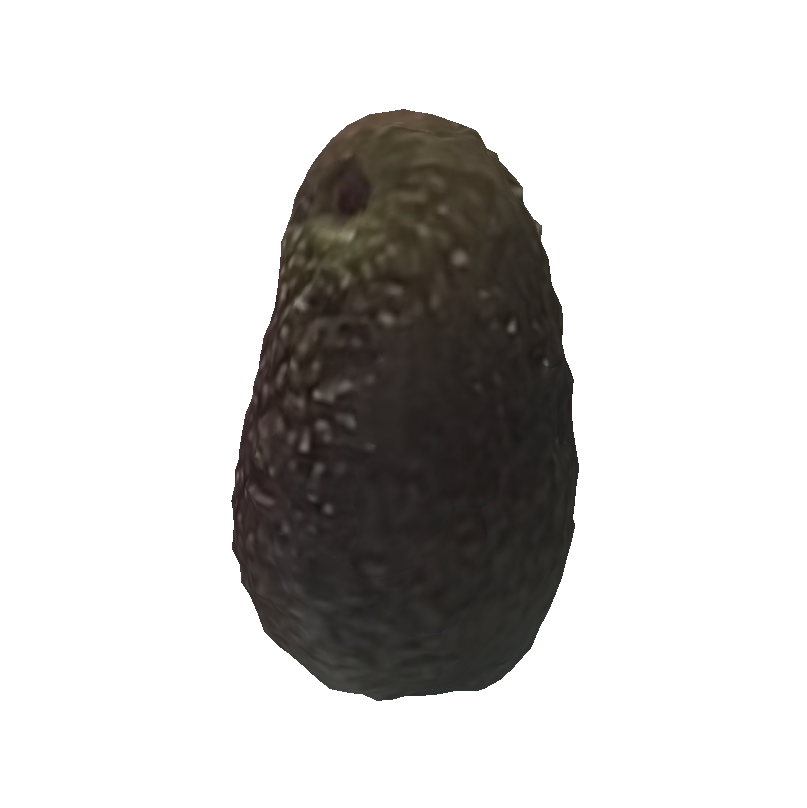}
                \includegraphics[clip,width=0.45\linewidth]{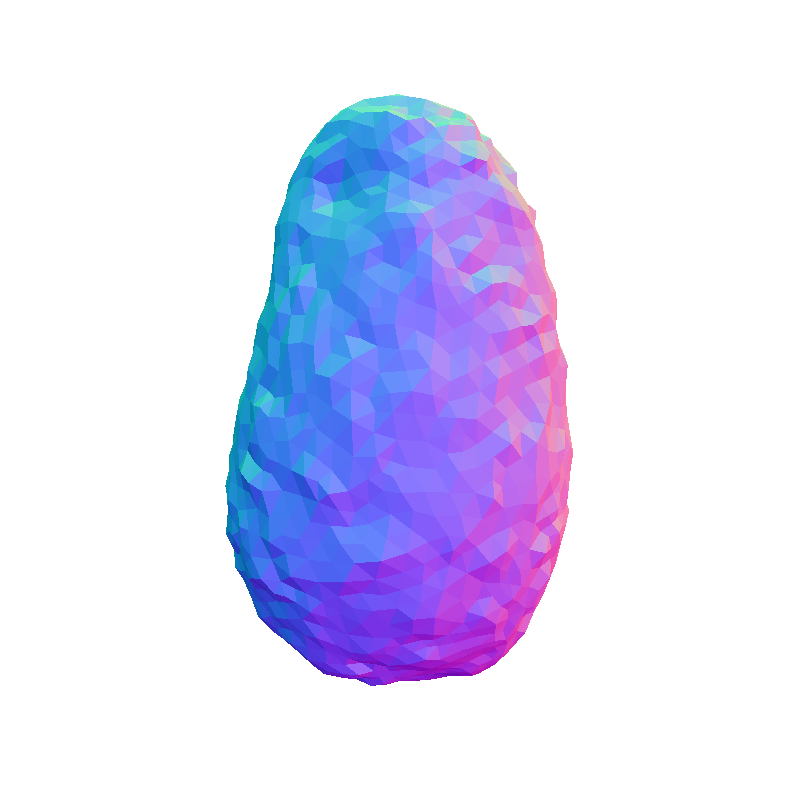}
            \caption{Aguacate}
        \end{subfigure}
        \begin{subfigure}[b]{0.13\linewidth}
             \centering
                \includegraphics[clip,width=0.45\linewidth]{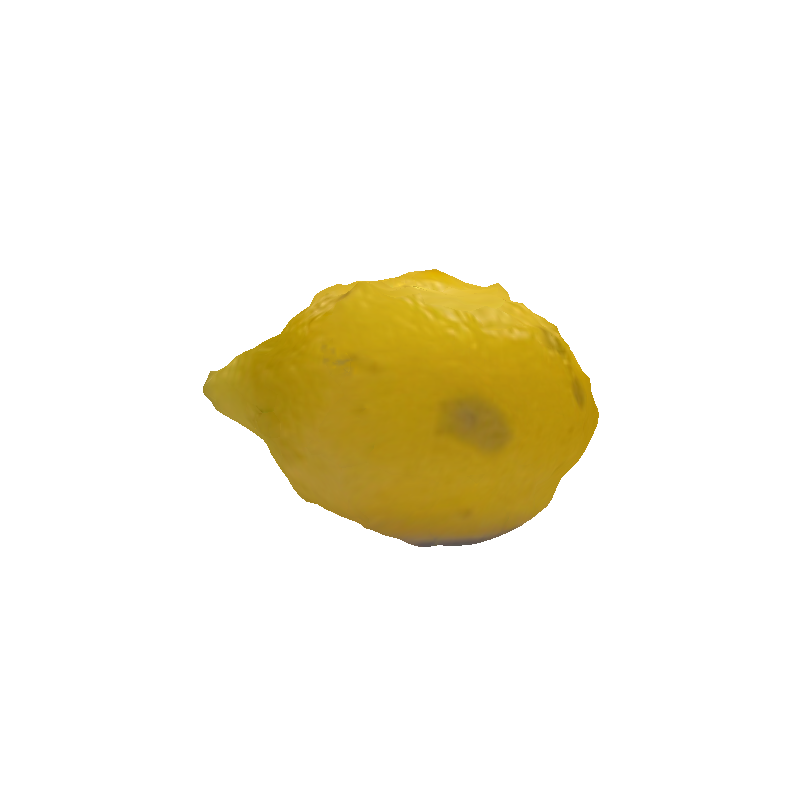}
                \includegraphics[clip,width=0.45\linewidth]{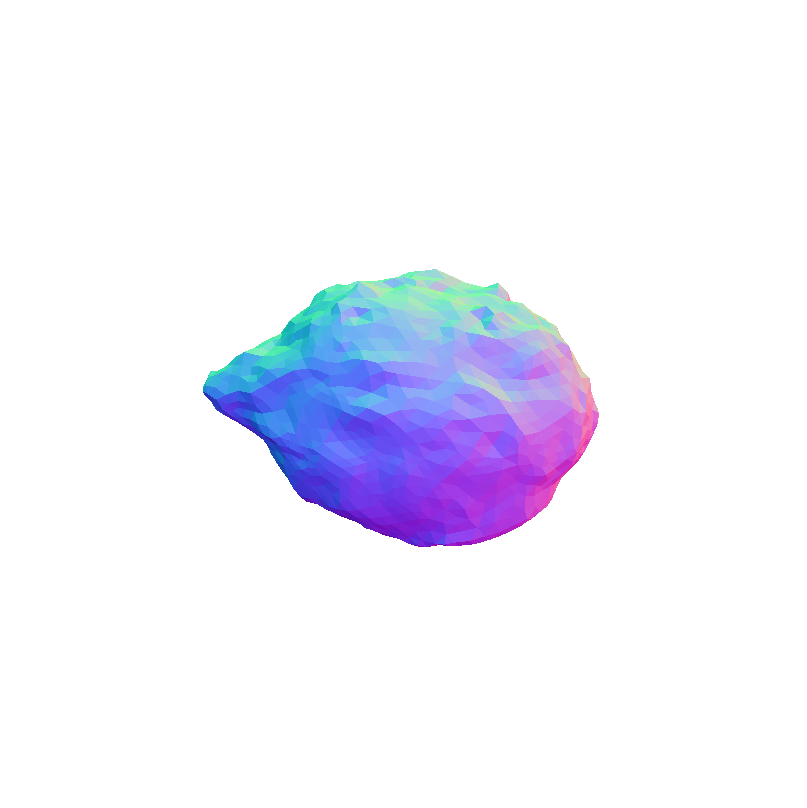}
            \caption{Lemon}
        \end{subfigure}
        \begin{subfigure}[b]{0.13\linewidth}
            \centering
                \includegraphics[clip,width=0.45\linewidth]{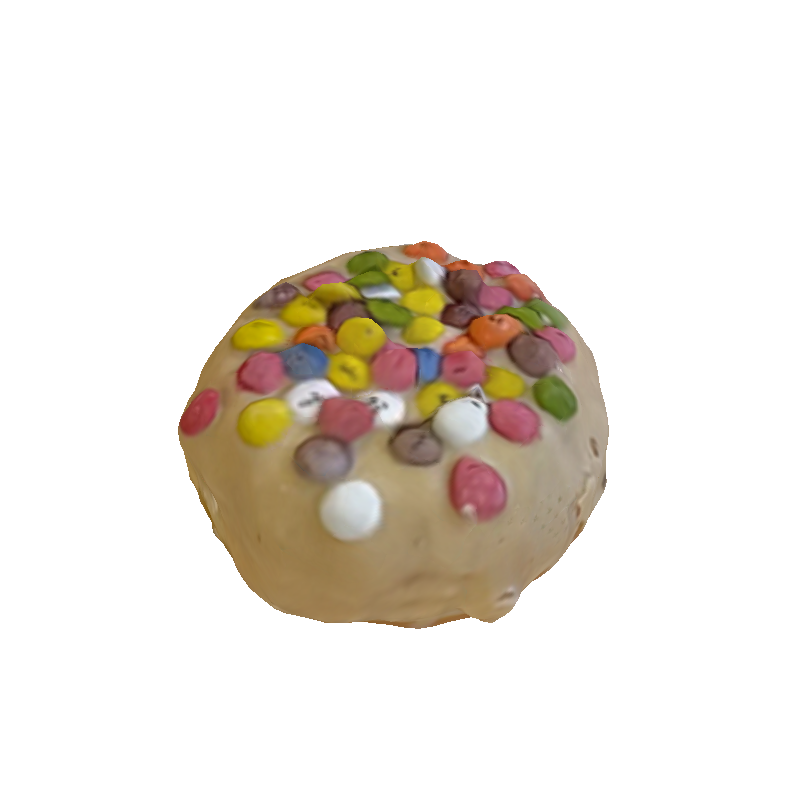}
                \includegraphics[clip,width=0.45\linewidth]{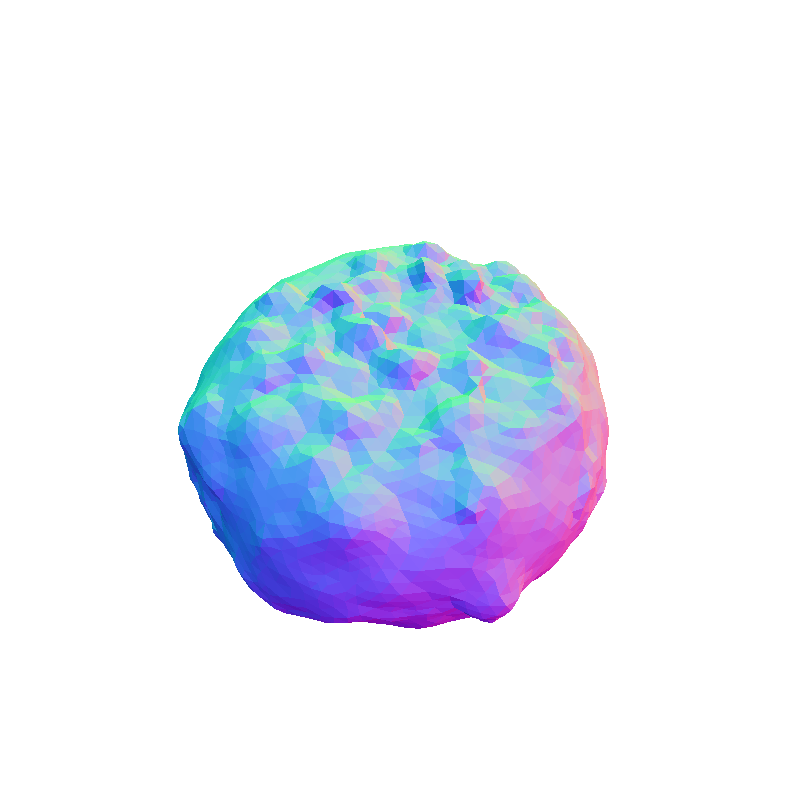}
            \caption{Donut}
        \end{subfigure}
        \begin{subfigure}[b]{0.13\linewidth}
            \centering
                \includegraphics[clip,width=0.45\linewidth]{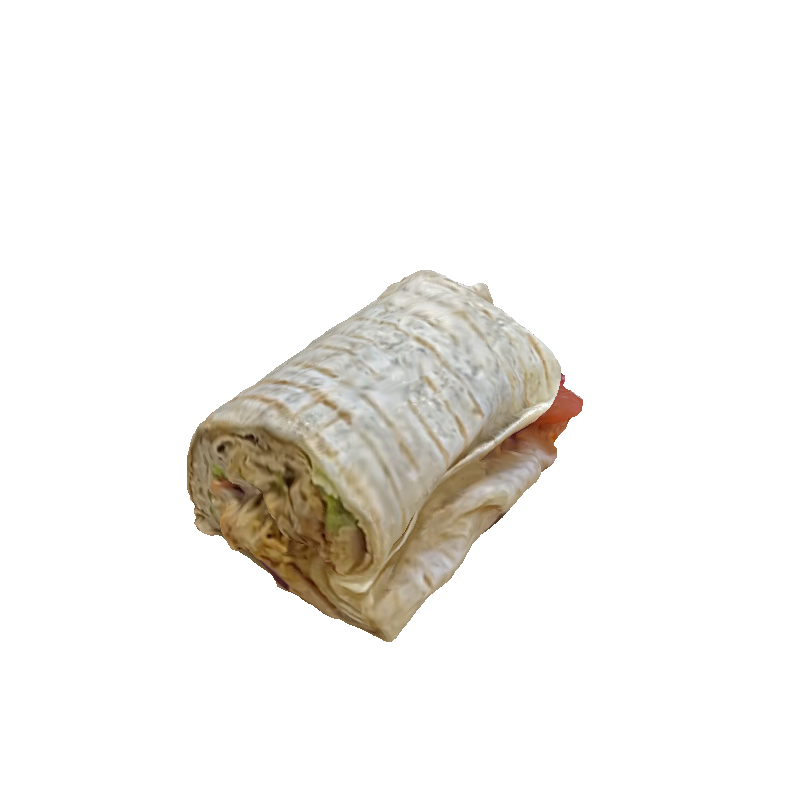}
                \includegraphics[clip,width=0.45\linewidth]{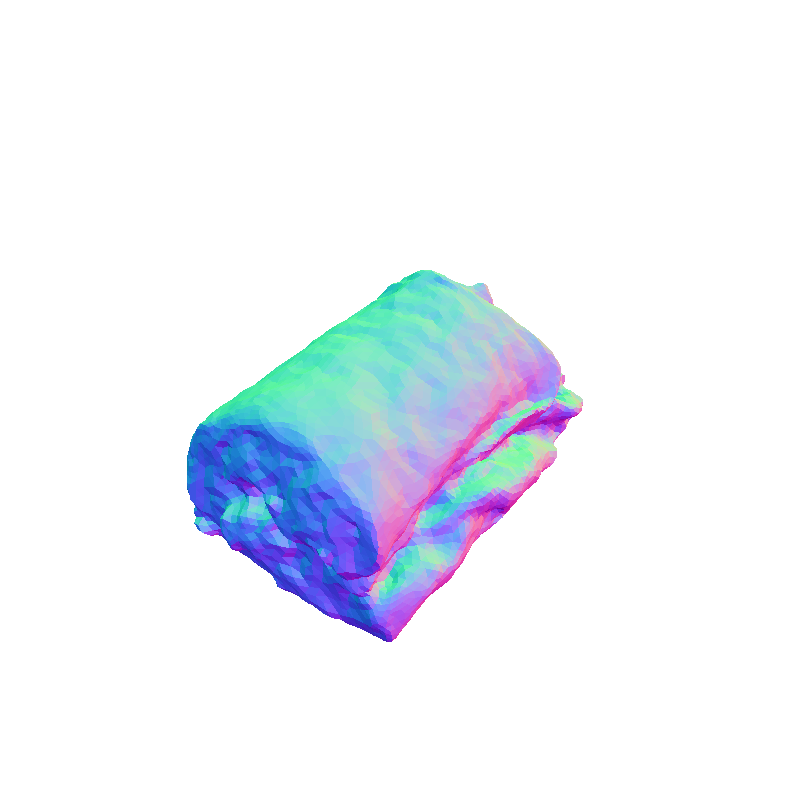}
            \caption{Durum}
        \end{subfigure}
        \begin{subfigure}[b]{0.13\linewidth}
            \centering
                \includegraphics[clip,width=0.45\linewidth]{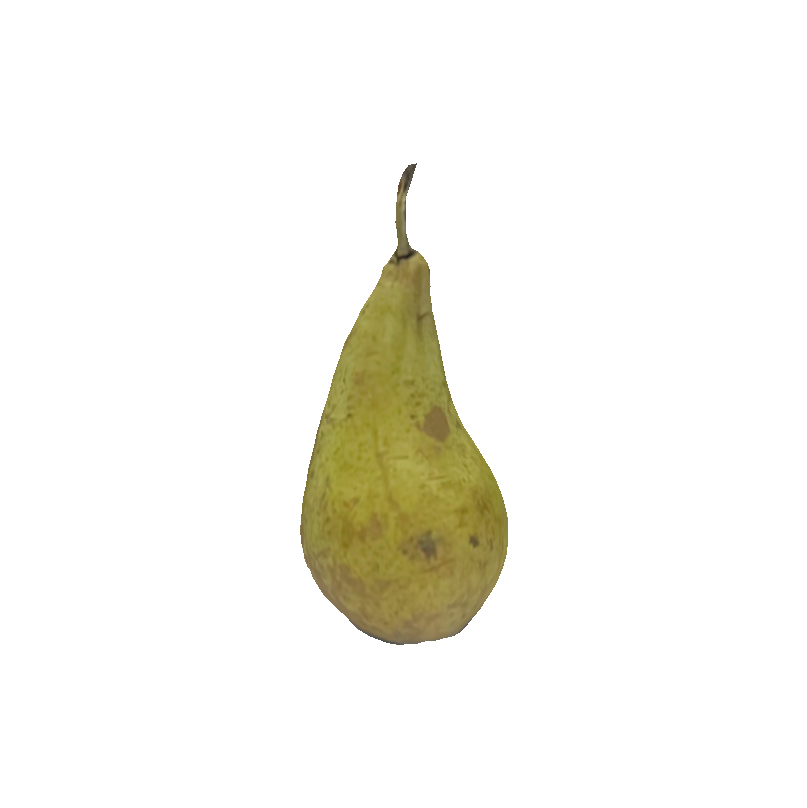}
                \includegraphics[clip,width=0.45\linewidth]{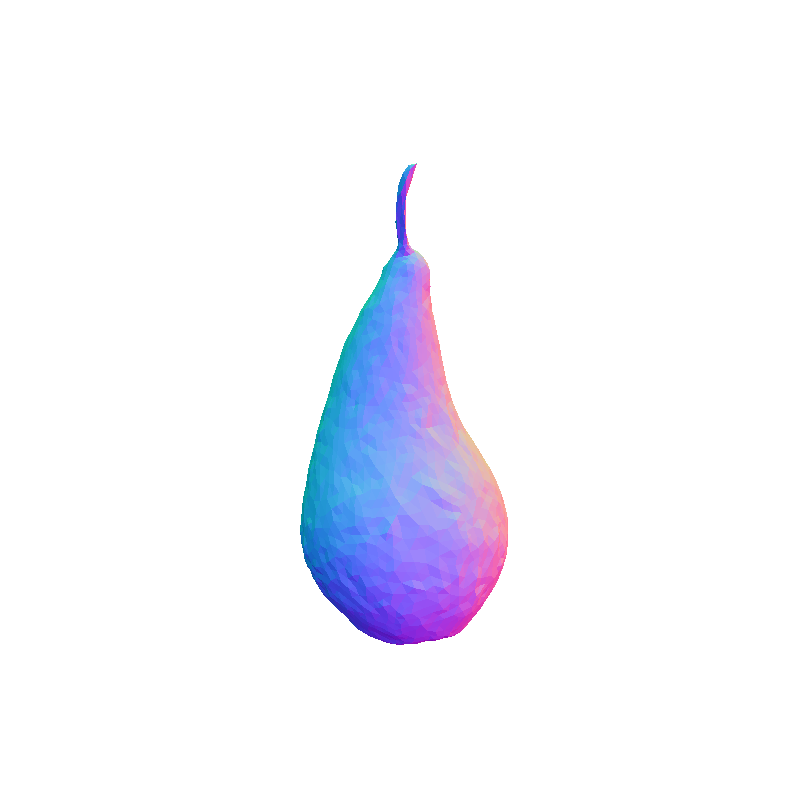}
            \caption{Pear}
        \end{subfigure}

        \begin{subfigure}[b]{0.13\linewidth}
            \centering
                \includegraphics[clip,width=0.45\linewidth]{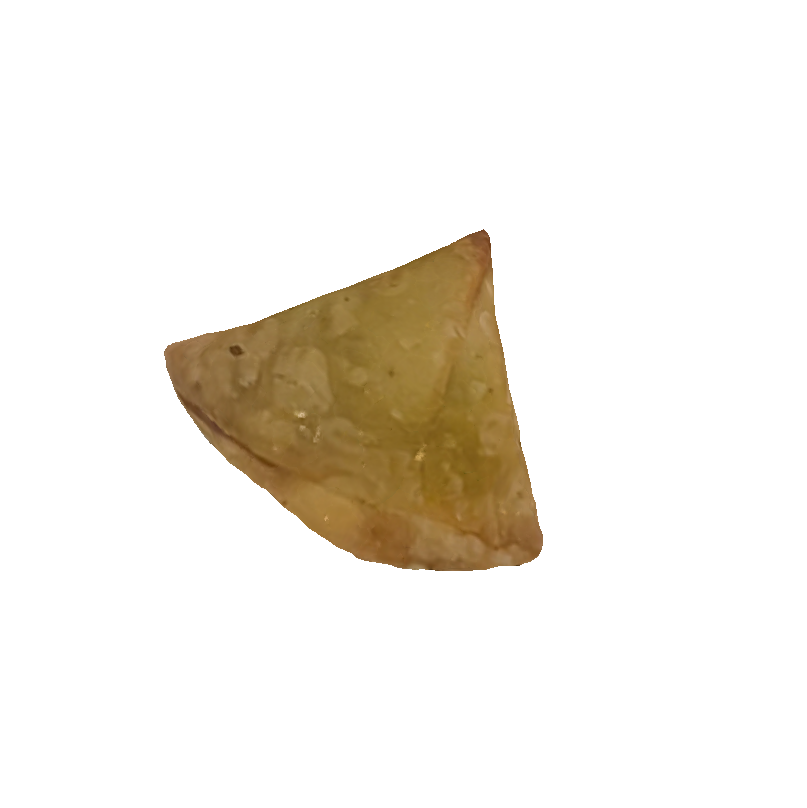}
                \includegraphics[clip,width=0.45\linewidth]{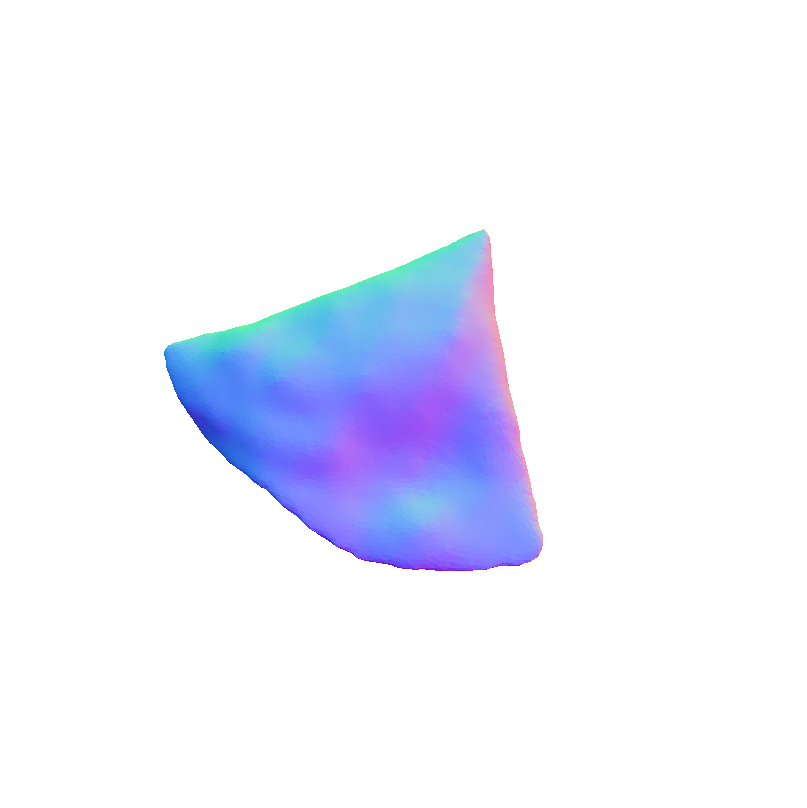}
            \caption{Apple}
        \end{subfigure}
        \begin{subfigure}[b]{0.13\linewidth}
            \centering
                \includegraphics[clip,width=0.45\linewidth]{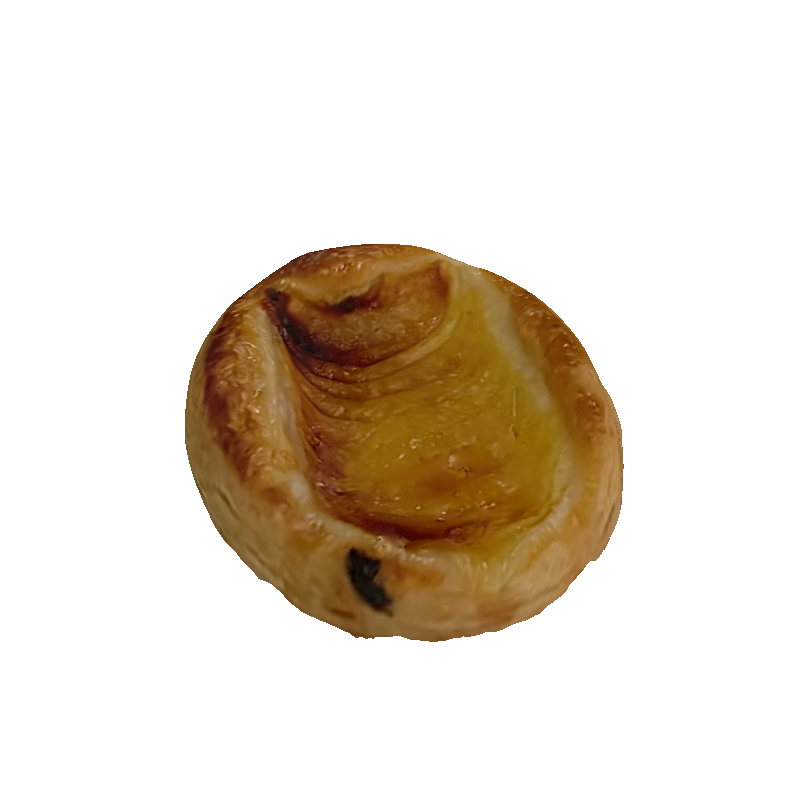}
                \includegraphics[clip,width=0.45\linewidth]{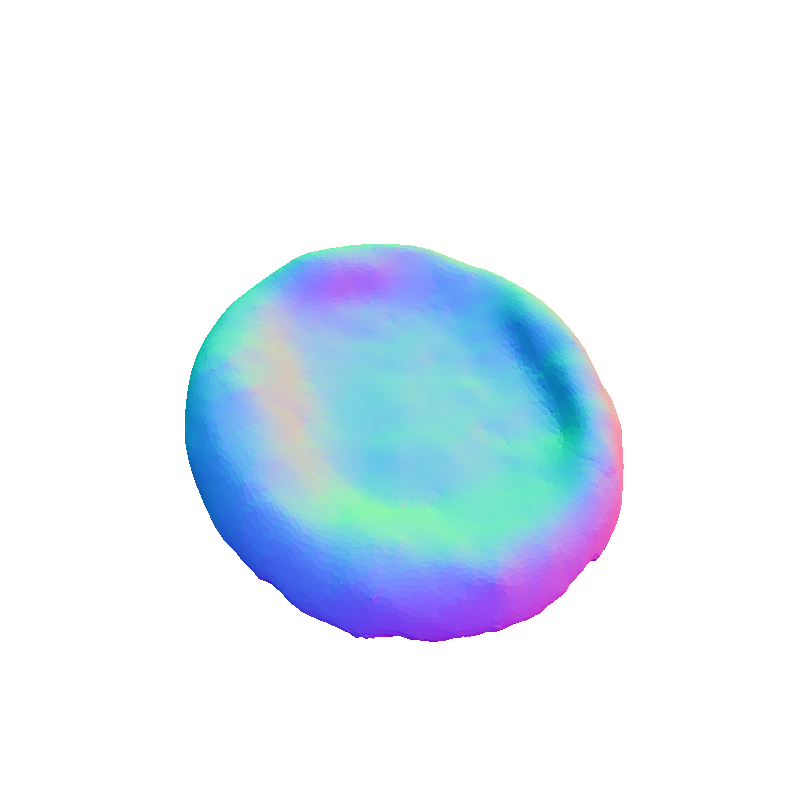}
            \caption{Orange}
        \end{subfigure}
        \begin{subfigure}[b]{0.13\linewidth}
            \centering
                \includegraphics[clip,width=0.45\linewidth]{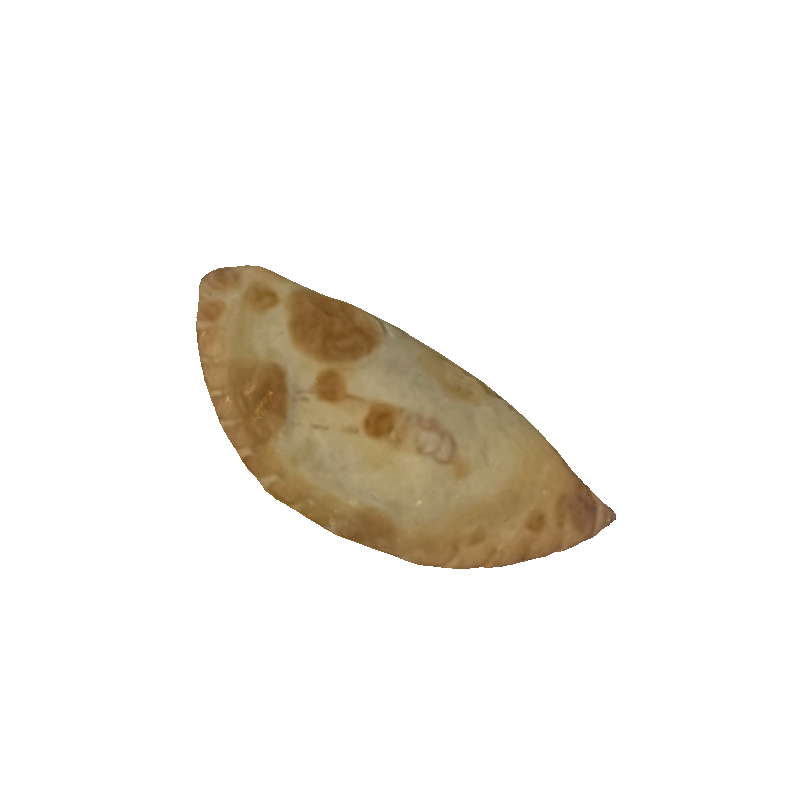}
                \includegraphics[clip,width=0.45\linewidth]{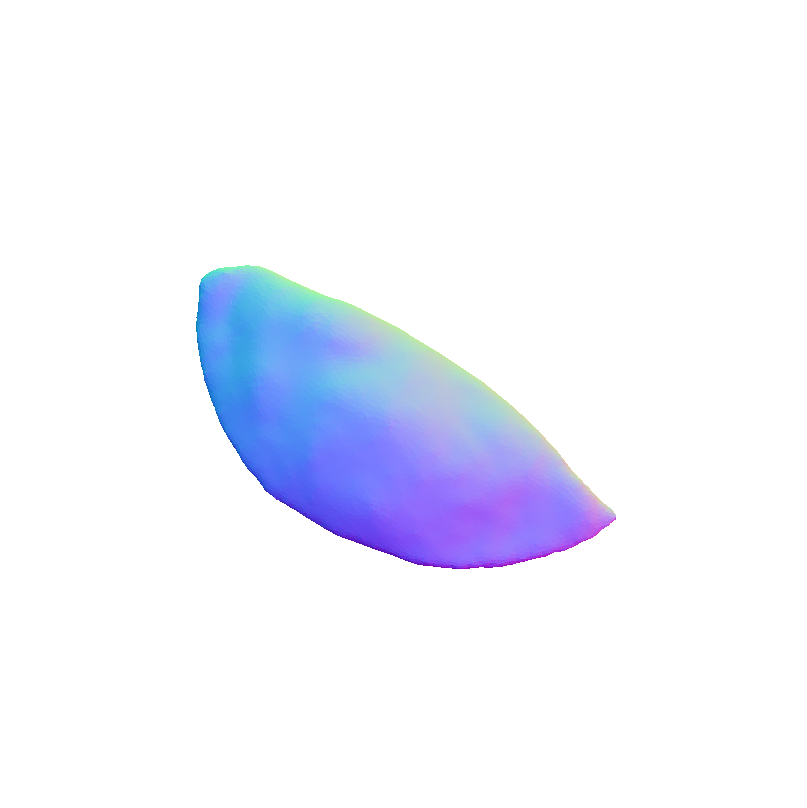}
            \caption{Aguacate}
        \end{subfigure}
        \begin{subfigure}[b]{0.13\linewidth}
             \centering
                \includegraphics[clip,width=0.45\linewidth]{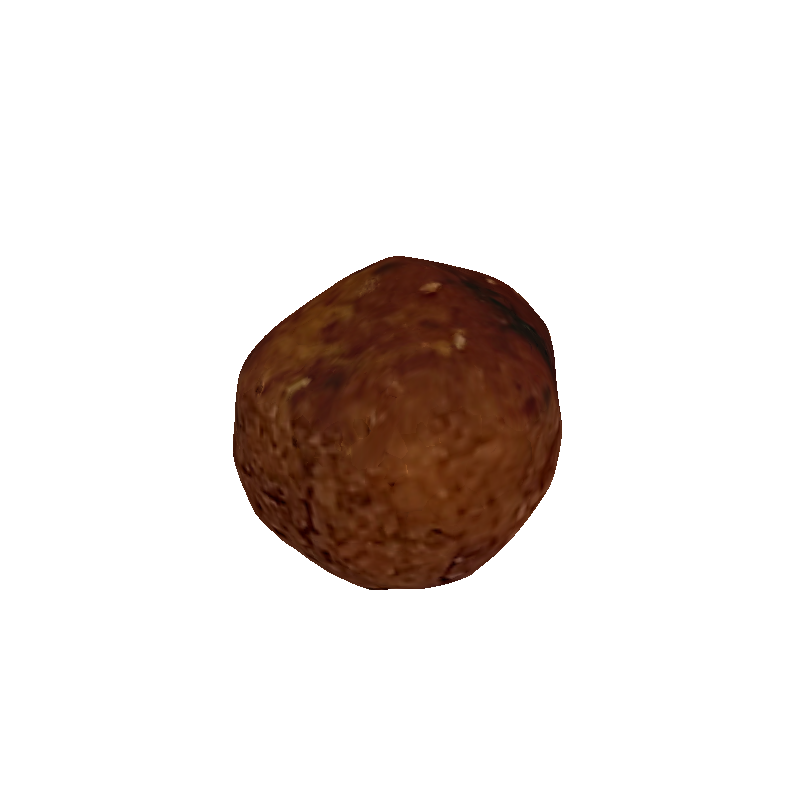}
                \includegraphics[clip,width=0.45\linewidth]{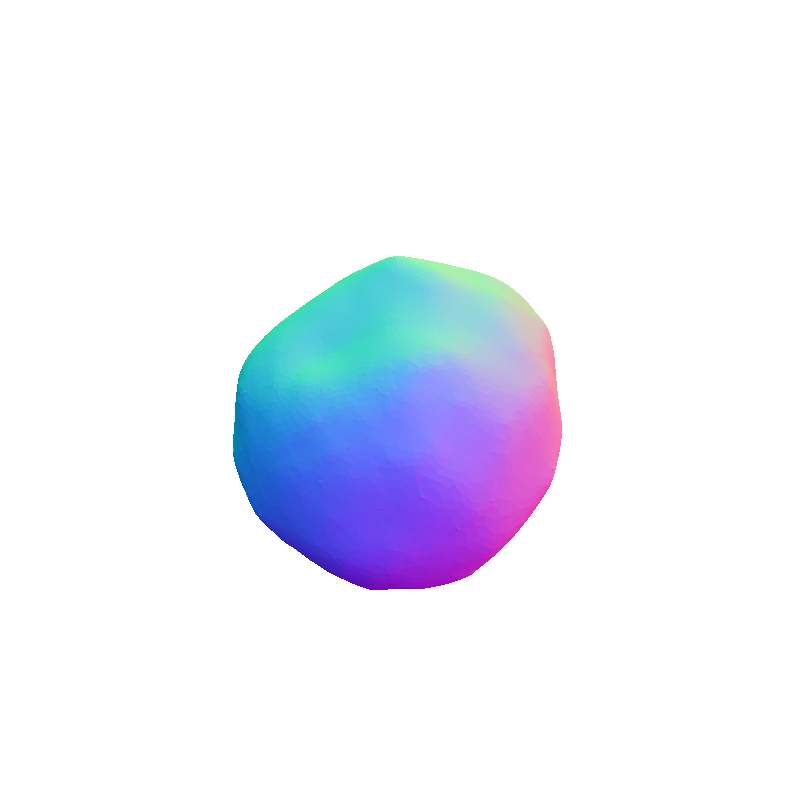}
            \caption{Lemon}
        \end{subfigure}
        \begin{subfigure}[b]{0.13\linewidth}
            \centering
                \includegraphics[clip,width=0.45\linewidth]{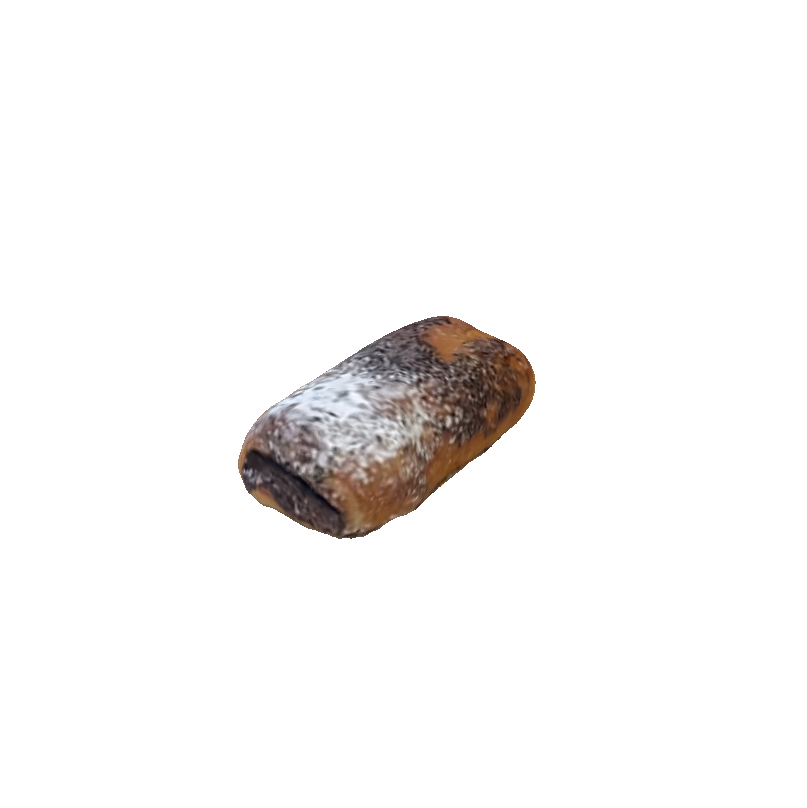}
                \includegraphics[clip,width=0.45\linewidth]{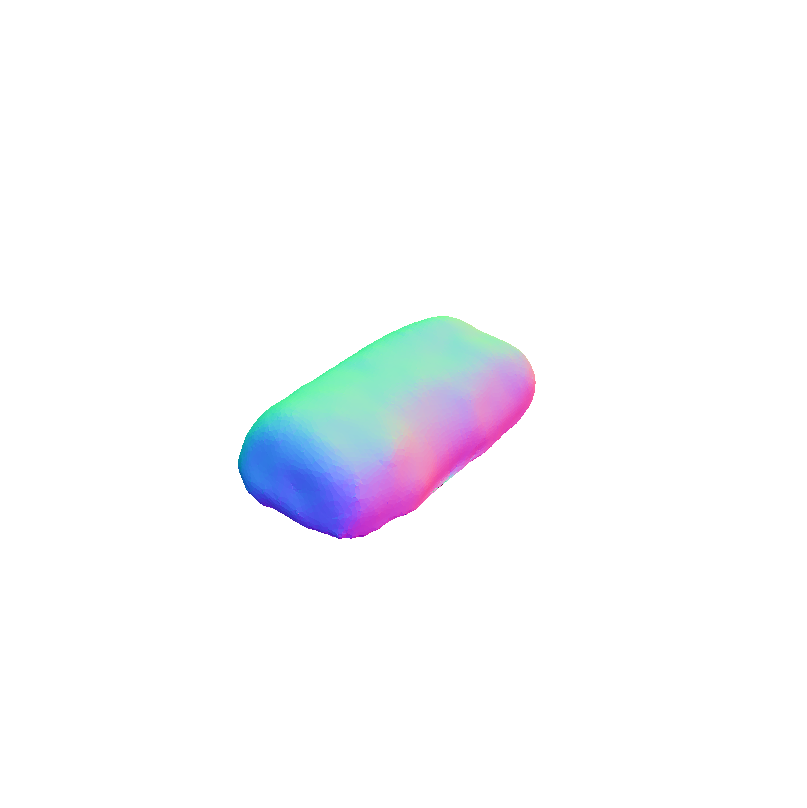}
            \caption{Donut}
        \end{subfigure}
        \begin{subfigure}[b]{0.13\linewidth}
            \centering
                \includegraphics[clip,width=0.45\linewidth]{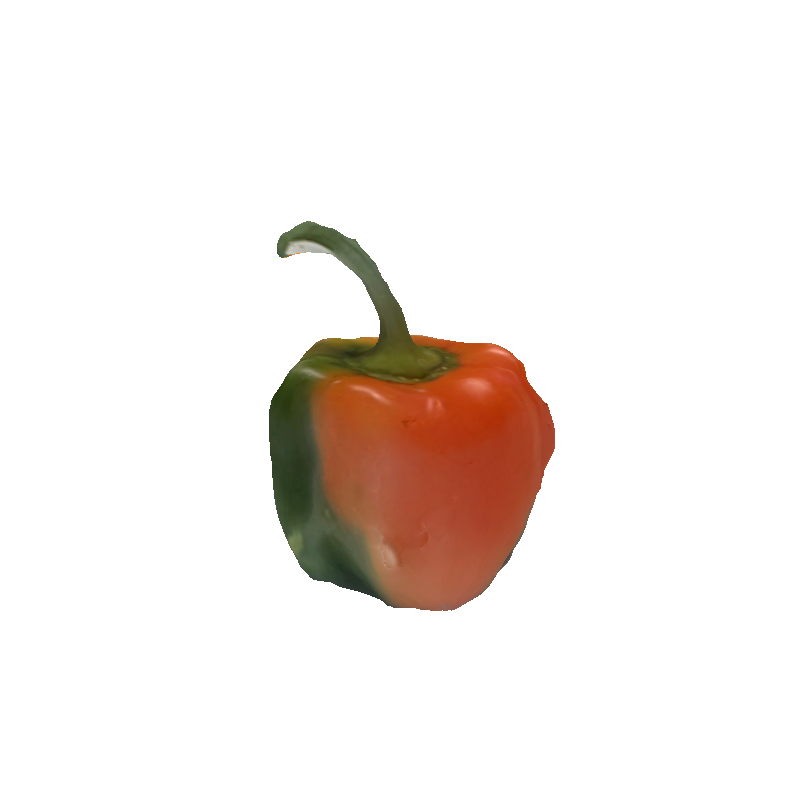}
                \includegraphics[clip,width=0.45\linewidth]{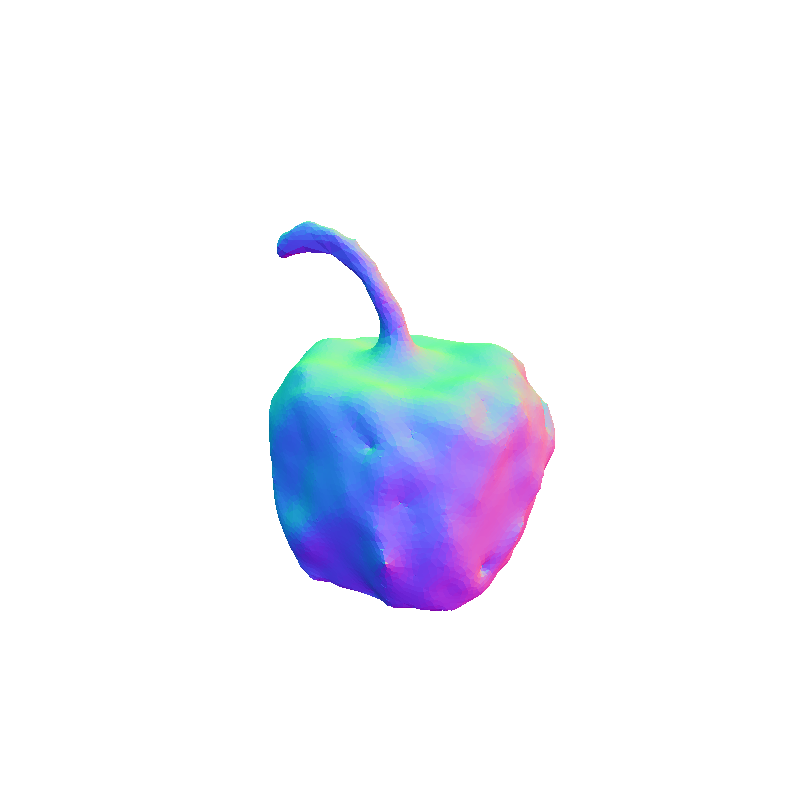}
            \caption{Durum}
        \end{subfigure}
        \begin{subfigure}[b]{0.13\linewidth}
            \centering
                \includegraphics[clip,width=0.45\linewidth]{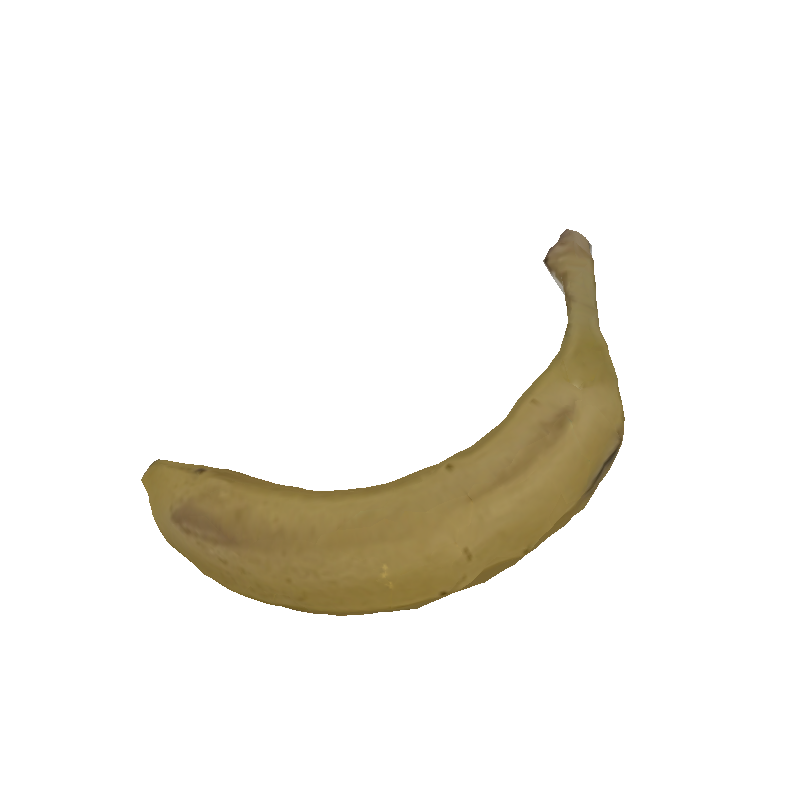}
                \includegraphics[clip,width=0.45\linewidth]{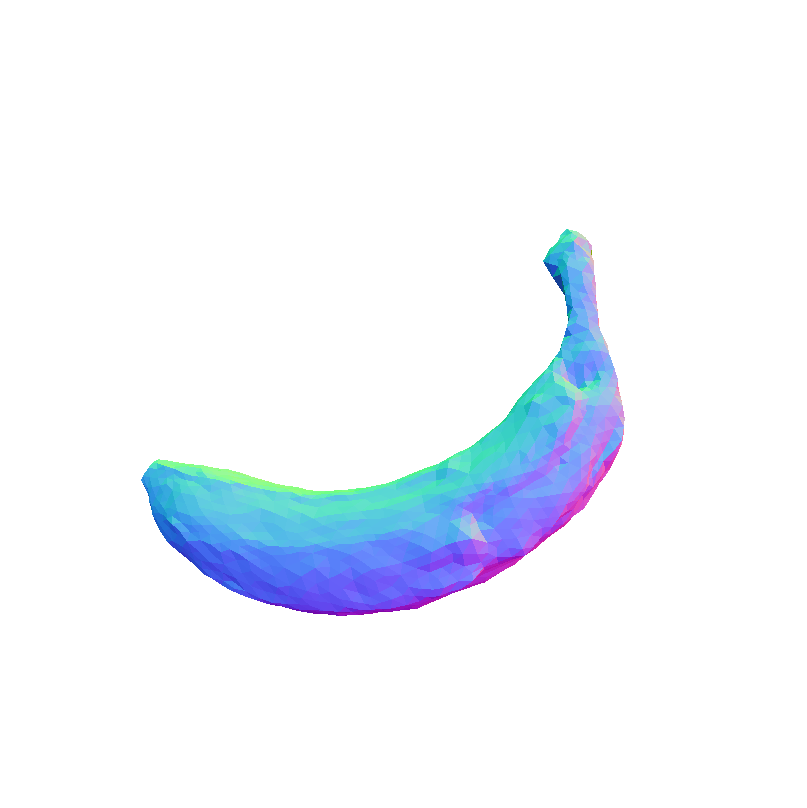}
            \caption{Pear}
        \end{subfigure}
        \begin{subfigure}[b]{0.13\linewidth}
            \centering
                \includegraphics[clip,width=0.45\linewidth]{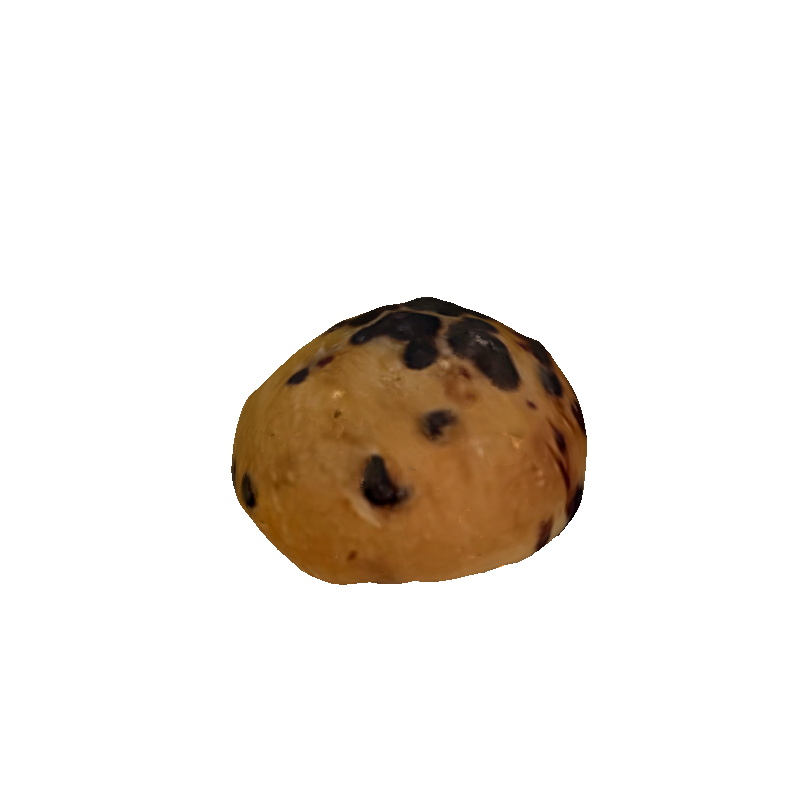}
                \includegraphics[clip,width=0.45\linewidth]{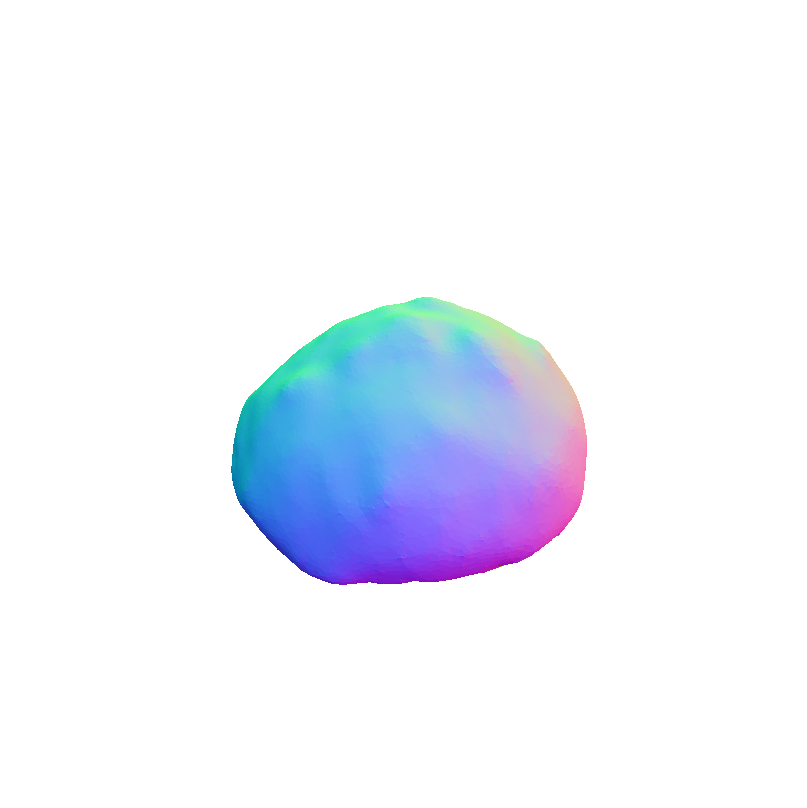}
            \caption{Apple}
        \end{subfigure}
        \begin{subfigure}[b]{0.13\linewidth}
            \centering
                \includegraphics[clip,width=0.45\linewidth]{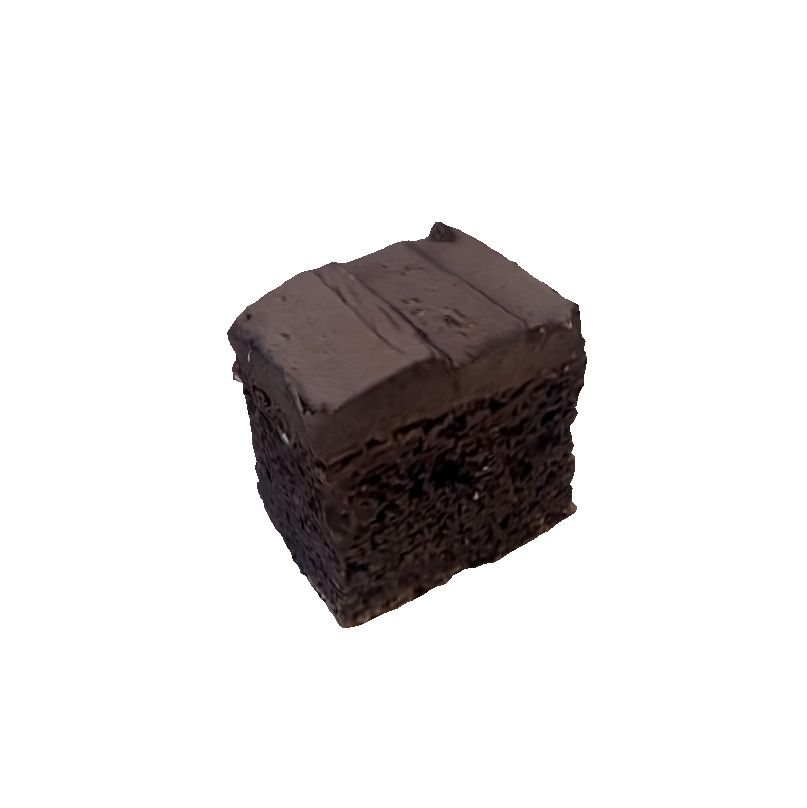}
                \includegraphics[clip,width=0.45\linewidth]{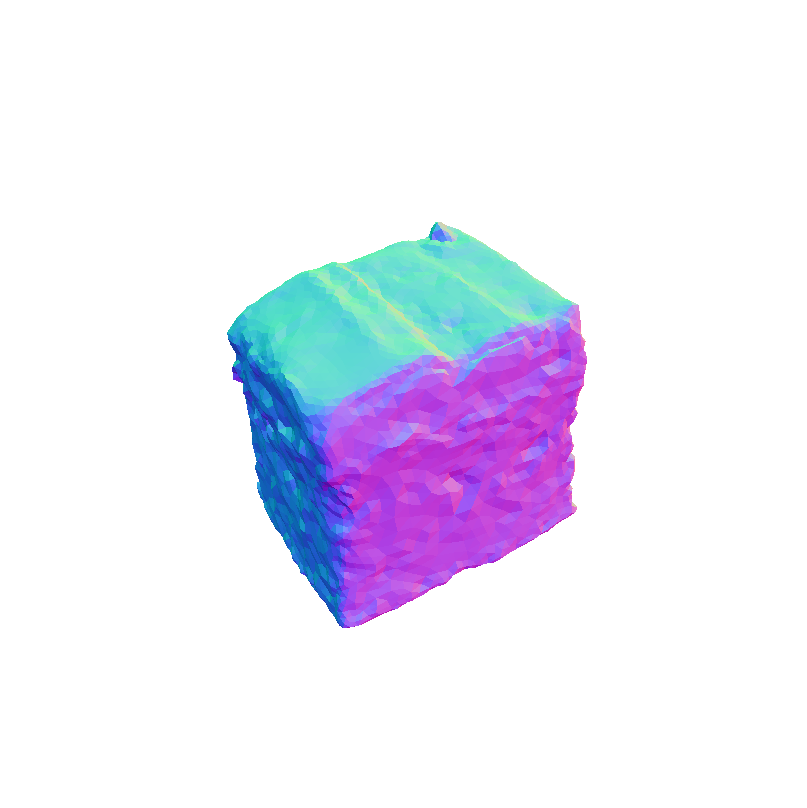}
            \caption{Orange}
        \end{subfigure}
        \begin{subfigure}[b]{0.13\linewidth}
            \centering
                \includegraphics[clip,width=0.45\linewidth]{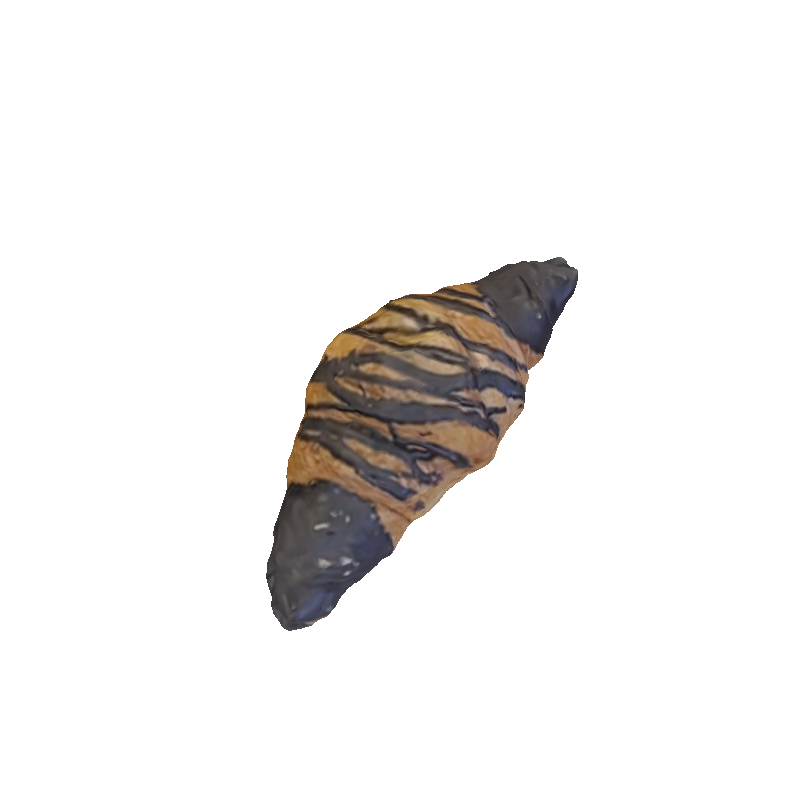}
                \includegraphics[clip,width=0.45\linewidth]{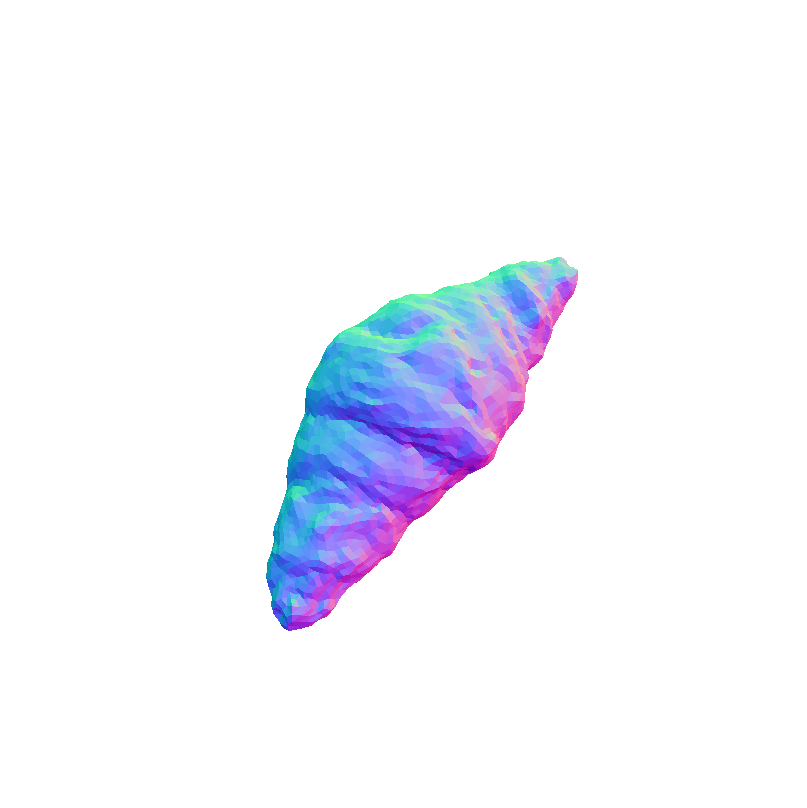}
            \caption{Aguacate}
        \end{subfigure}
        \begin{subfigure}[b]{0.13\linewidth}
             \centering
                \includegraphics[clip,width=0.45\linewidth]{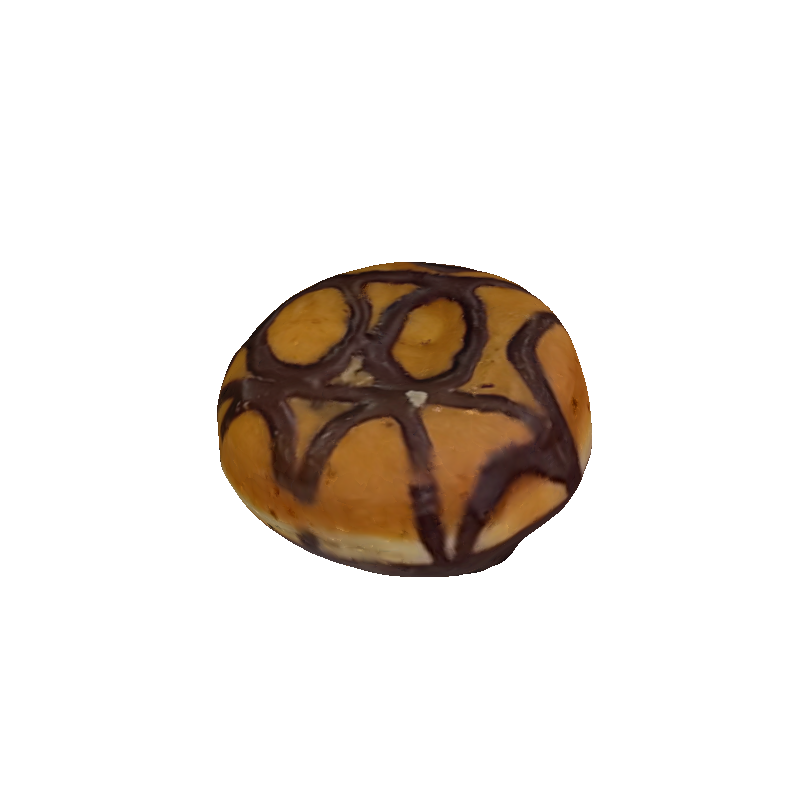}
                \includegraphics[clip,width=0.45\linewidth]{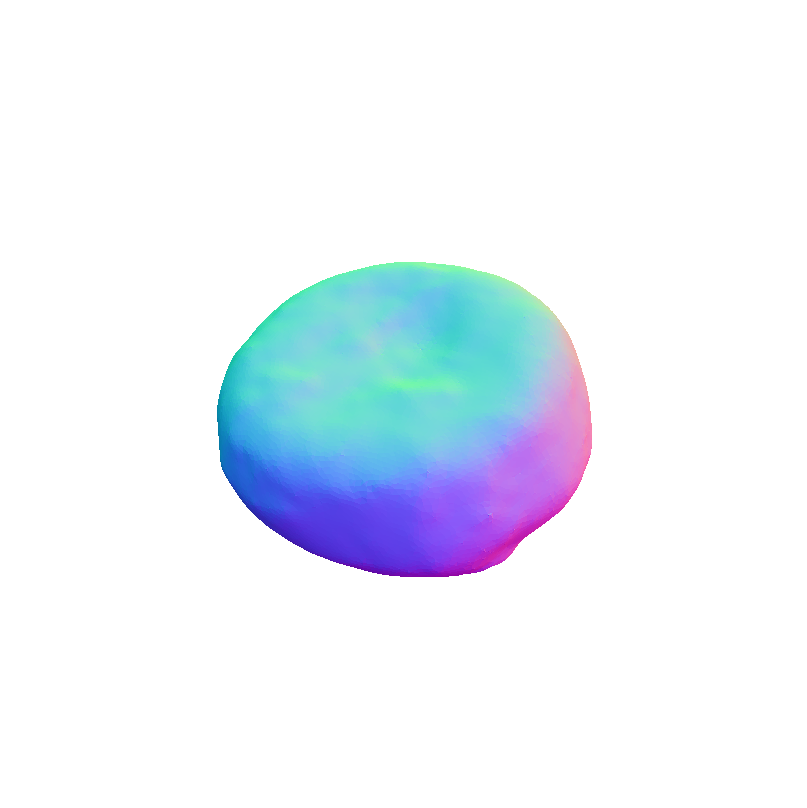}
            \caption{Lemon}
        \end{subfigure}
        \begin{subfigure}[b]{0.13\linewidth}
            \centering
                \includegraphics[clip,width=0.45\linewidth]{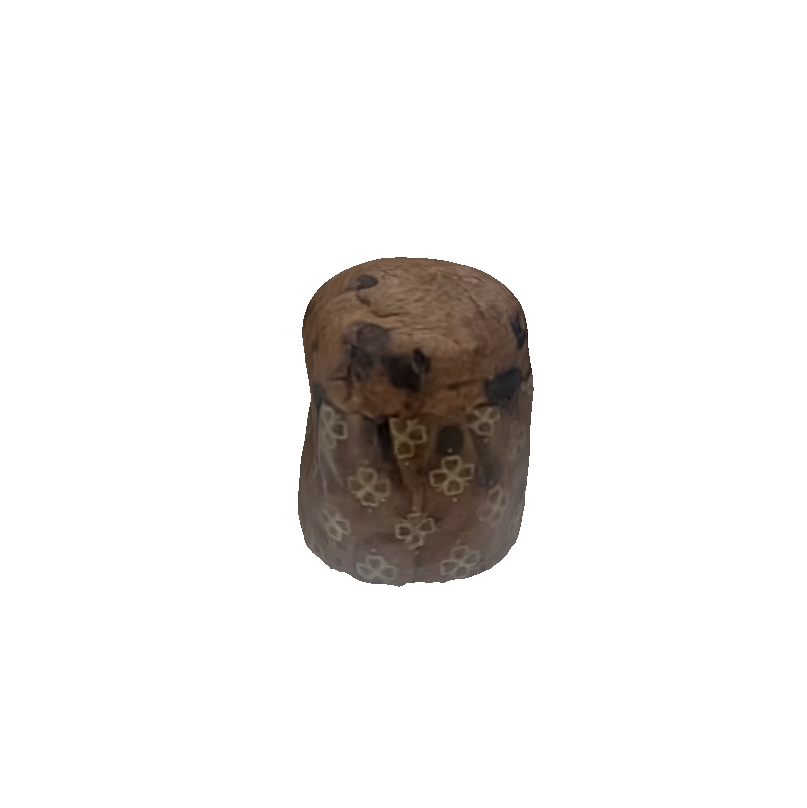}
                \includegraphics[clip,width=0.45\linewidth]{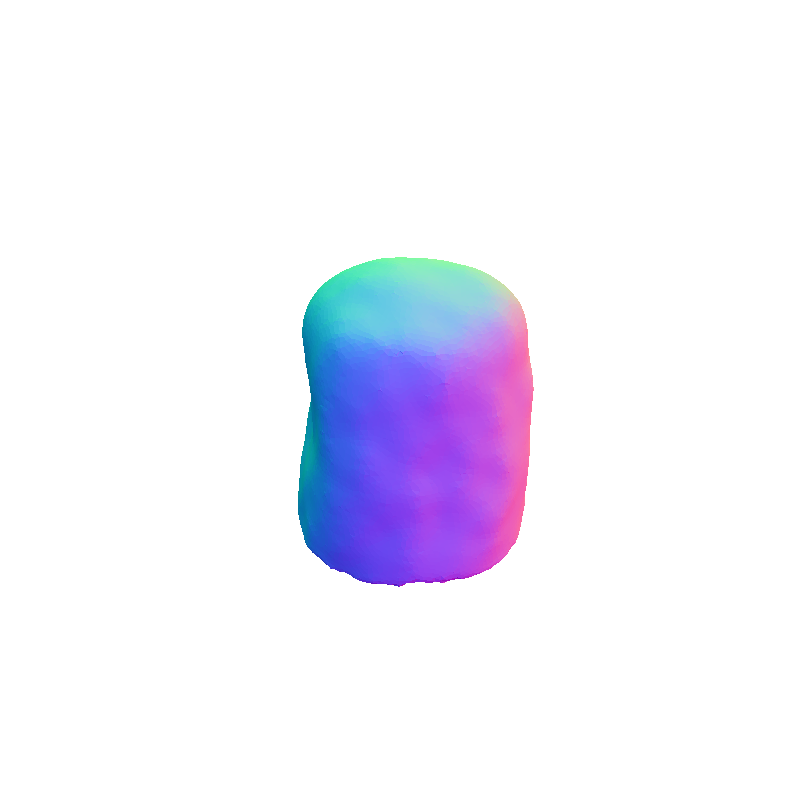}
            \caption{Donut}
        \end{subfigure}
        \begin{subfigure}[b]{0.13\linewidth}
            \centering
                \includegraphics[clip,width=0.45\linewidth]{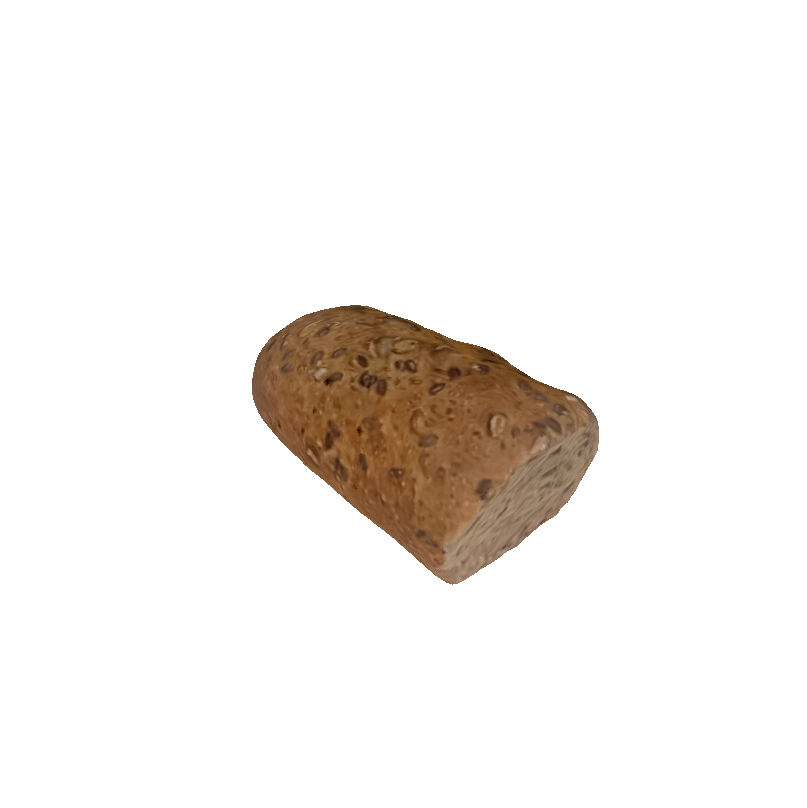}
                \includegraphics[clip,width=0.45\linewidth]{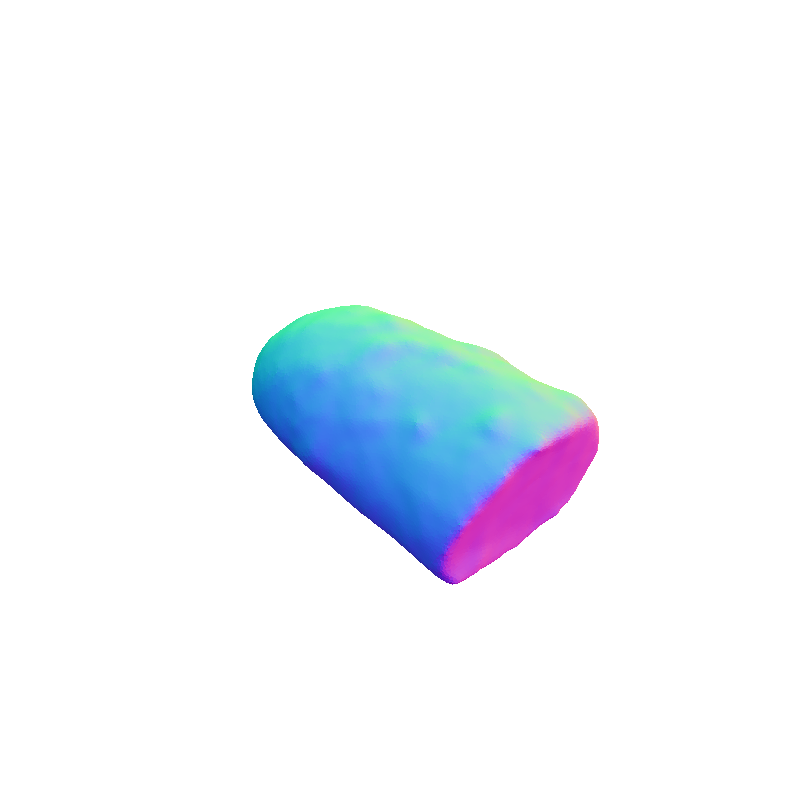}
            \caption{Durum}
        \end{subfigure}
        \begin{subfigure}[b]{0.13\linewidth}
            \centering
                \includegraphics[clip,width=0.45\linewidth]{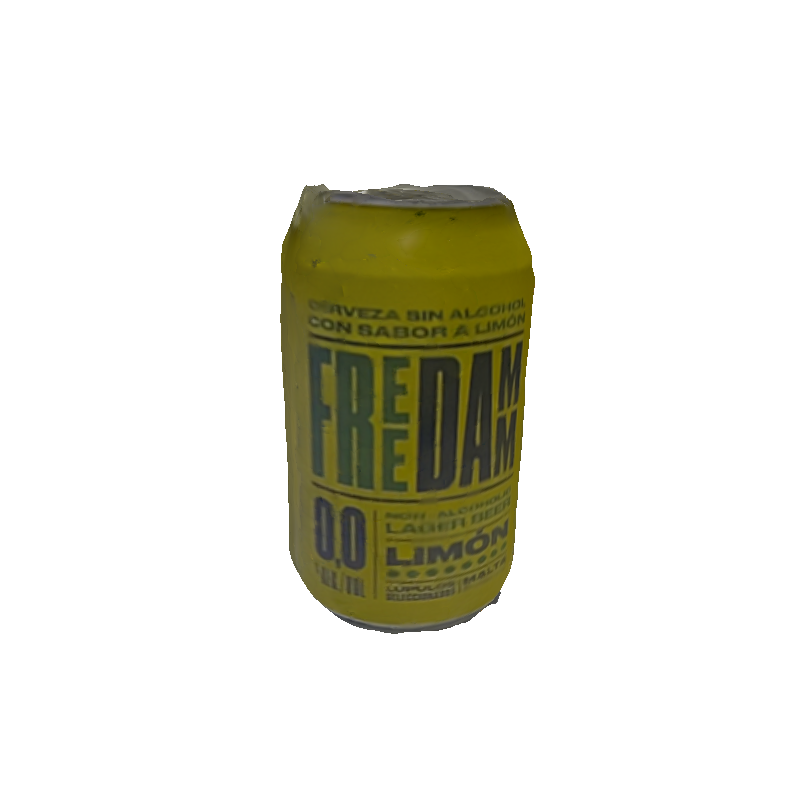}
                \includegraphics[clip,width=0.45\linewidth]{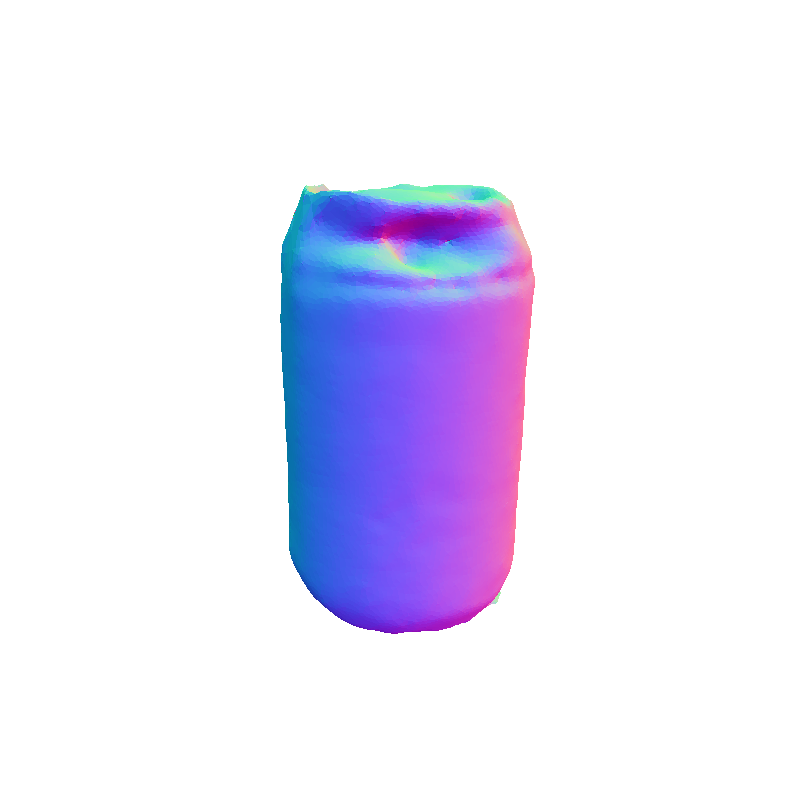}
            \caption{Pear}
        \end{subfigure}
    
    \caption{Visual representation of our framework results on the Foodkit dataset. Each food item is displayed with a textured mesh on the left and a shaded mesh on the right, emphasizing our framework's ability to accurately capture the 3D shape, texture, and structural details of various food items.}
    \label{fig:Results_VolE_dataset}
\end{figure*}

\begin{figure*}[htb]
    \centering
    \tiny
    \arrayrulecolor{black!30}
    \setlength{\tabcolsep}{1pt}
    \begin{tabular}{c|c|c|c|c|c|c|c|c|c|c|c|c|c}
        \hline
          & Strawberry & Cinnamon Bun & Pork Rib & Corn & French Toast & Sandwich & Burger & Cake & Blueberry Muffin & Banana & Salmon & Burrito & Hotdog        
        \\
        
        \hline
        \raisebox{+0.3cm}{\rotatebox[origin=l]{90}{\footnotesize{GT}}} &
        \includegraphics[clip,width=0.07\linewidth]{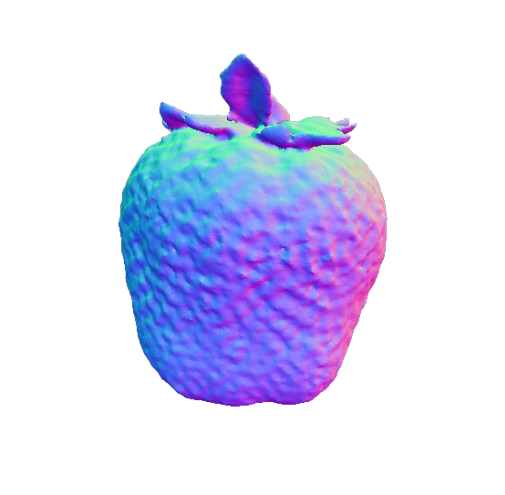} &
        \includegraphics[clip,width=0.07\linewidth]{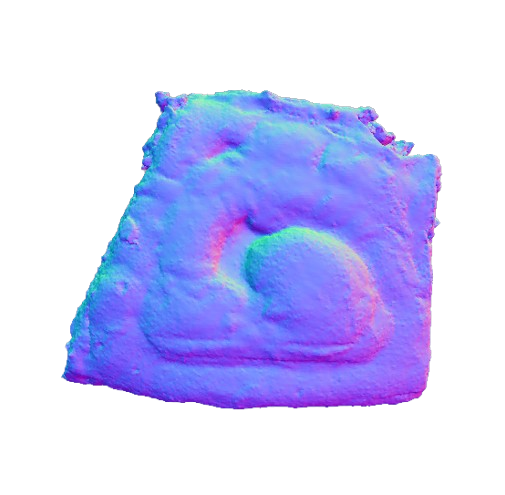} &
        \includegraphics[clip,width=0.07\linewidth]{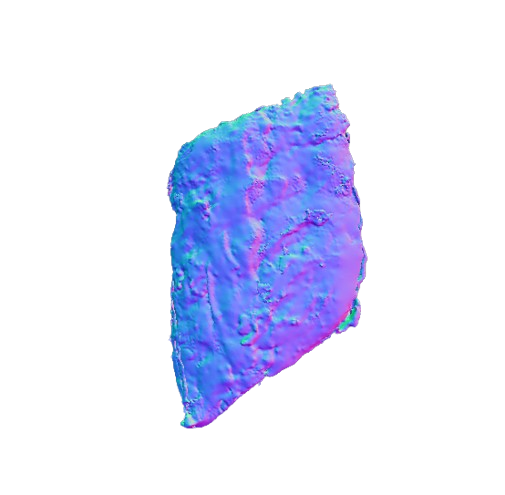} &
        \includegraphics[clip,width=0.07\linewidth]{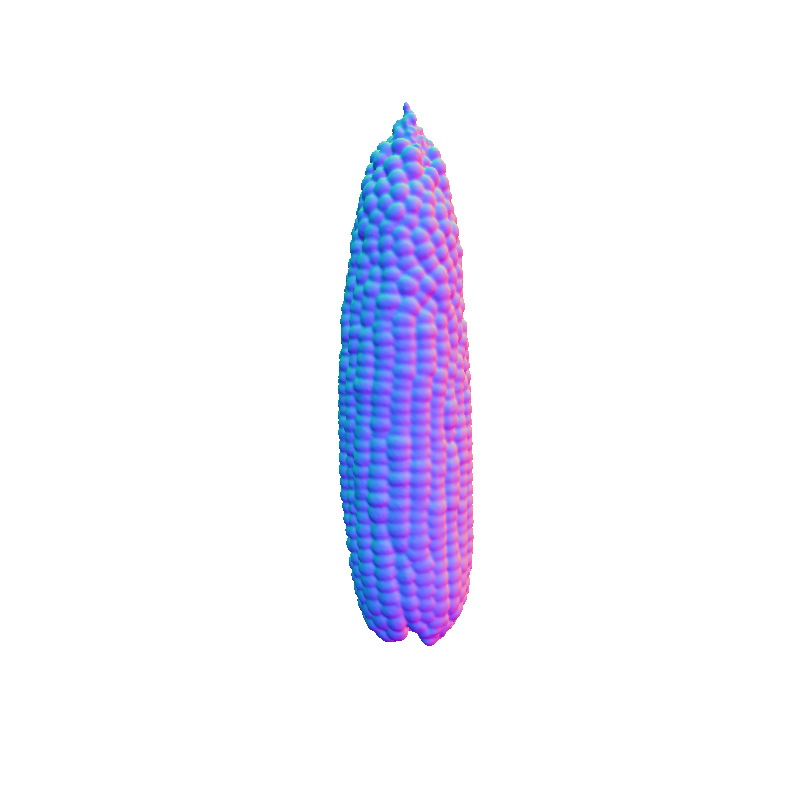} &
        \includegraphics[clip,width=0.07\linewidth]{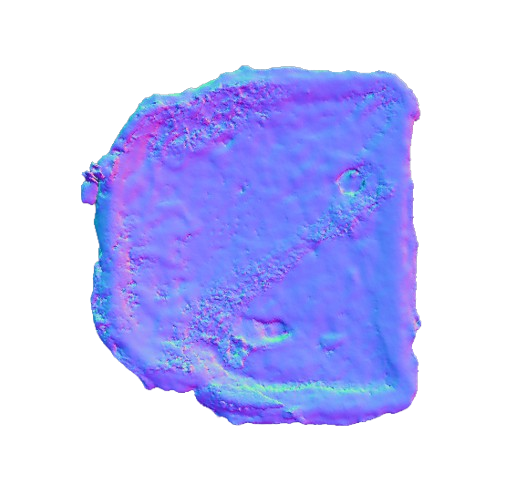} &
        \includegraphics[clip,width=0.07\linewidth]{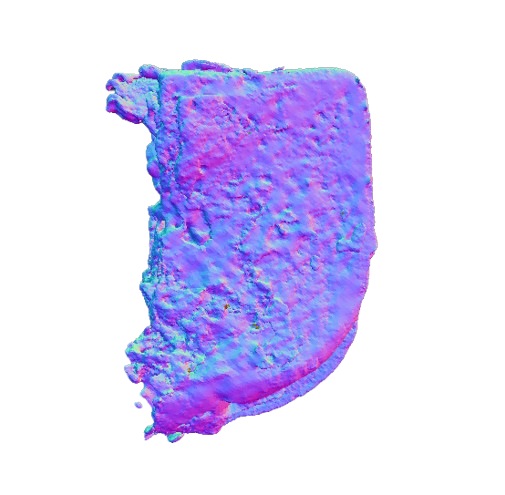} &
        \includegraphics[clip,width=0.07\linewidth]{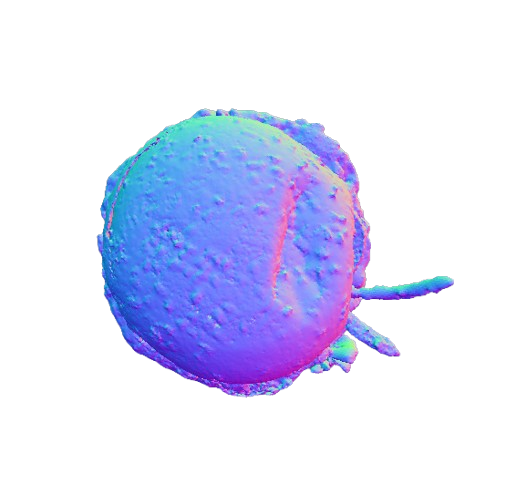} &
        \includegraphics[clip,width=0.07\linewidth]{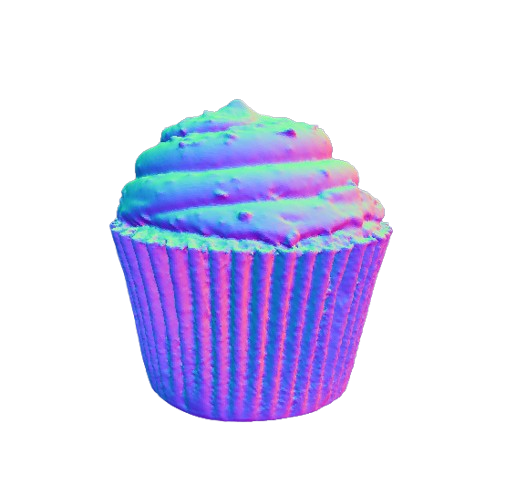} &
        \includegraphics[clip,width=0.07\linewidth]{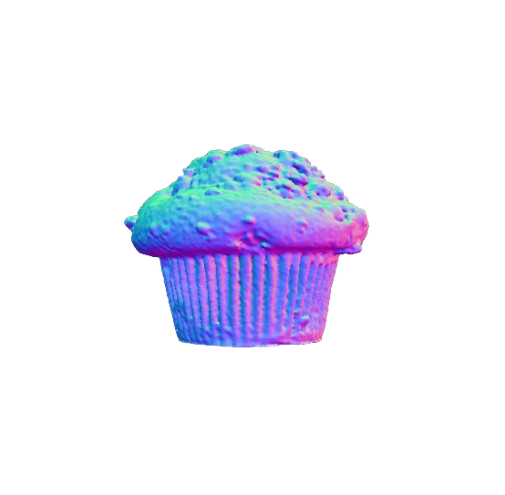} &
        \includegraphics[clip,width=0.07\linewidth]{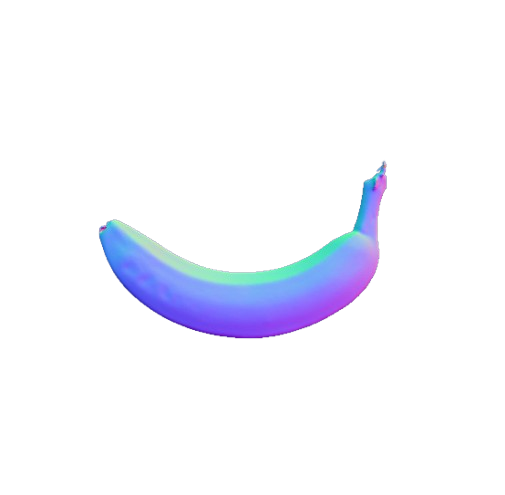} &
        \includegraphics[clip,width=0.07\linewidth]{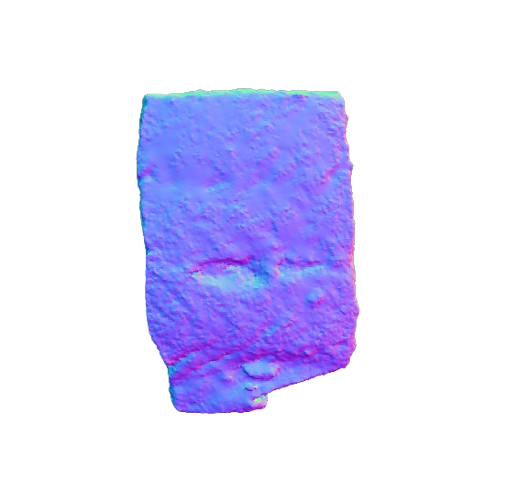} &
        \includegraphics[clip,width=0.07\linewidth]{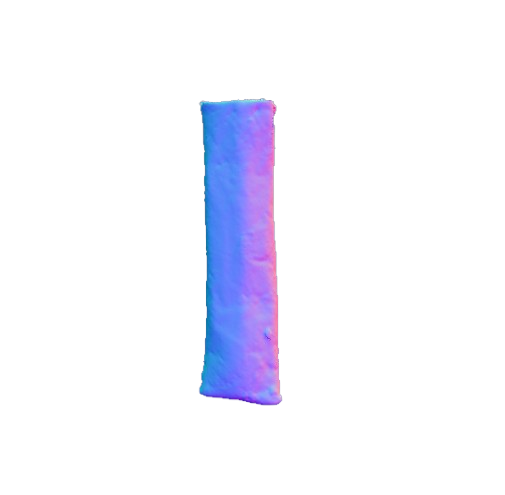} &
        \includegraphics[clip,width=0.07\linewidth]{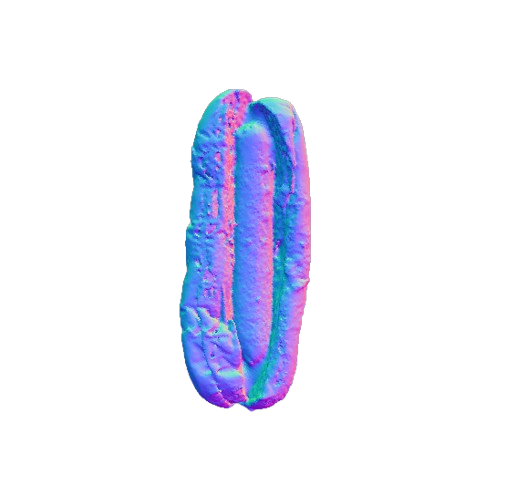}
        \\
          \rotatebox{90}{\footnotesize{VolETA~\cite{almughrabi2024voleta}}} &
        \includegraphics[clip,width=0.07\linewidth]{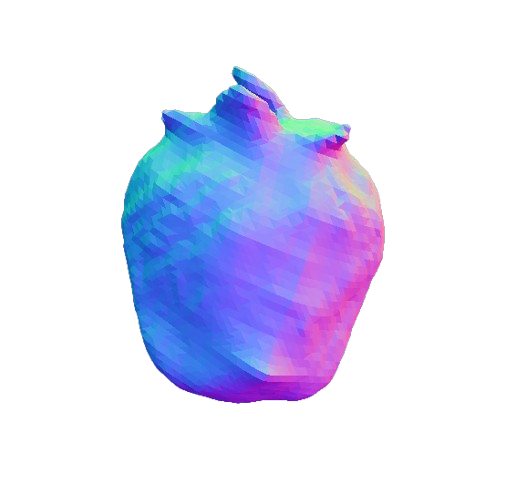} &
        \includegraphics[clip,width=0.07\linewidth]{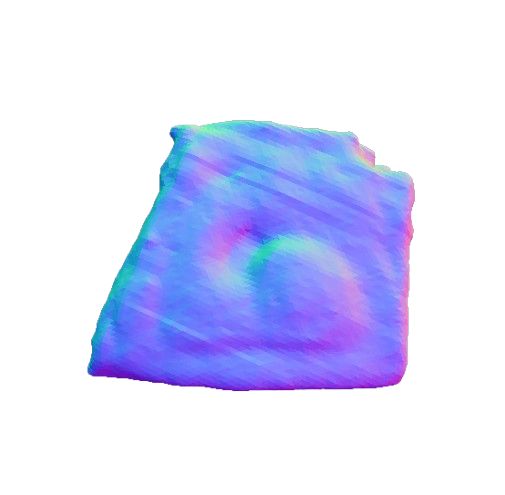} &
        \includegraphics[clip,width=0.07\linewidth]{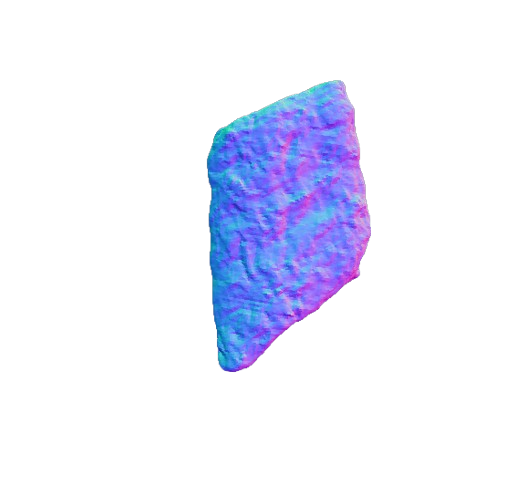} &
        \includegraphics[clip,width=0.07\linewidth]{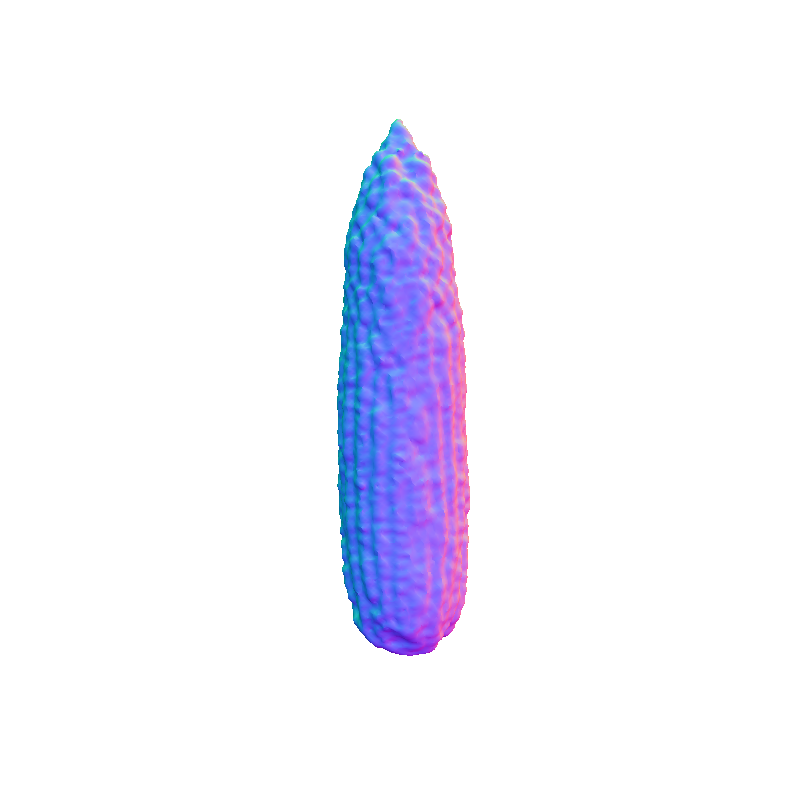} &
        \includegraphics[clip,width=0.07\linewidth]{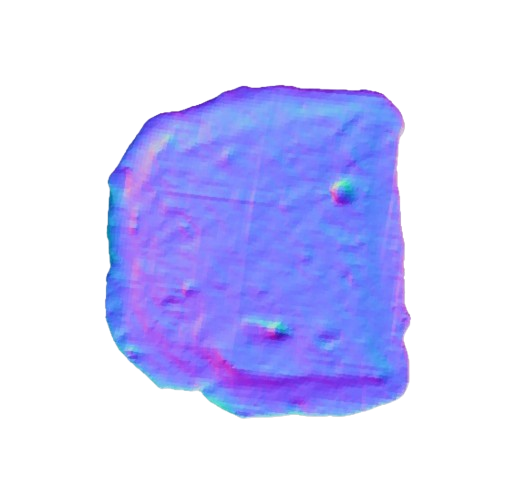} &
        \includegraphics[clip,width=0.07\linewidth]{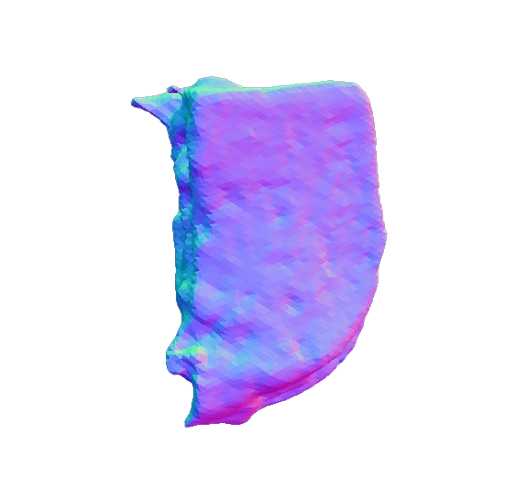} &
        \includegraphics[clip,width=0.07\linewidth]{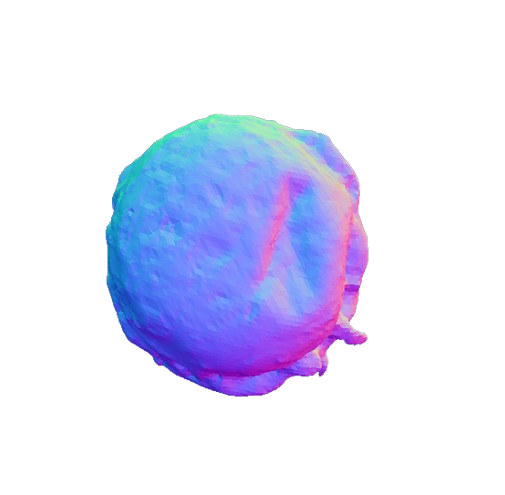} &
        \includegraphics[clip,width=0.07\linewidth]{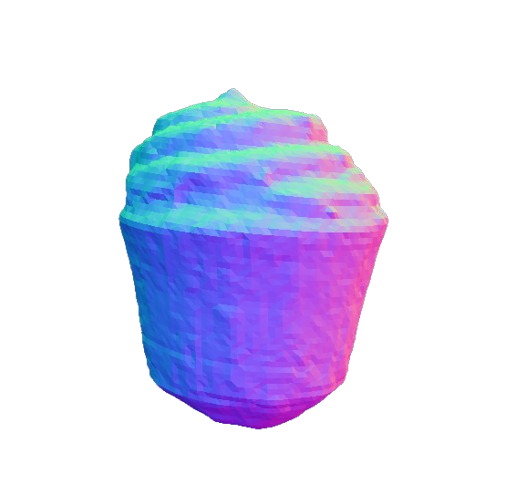} &
        \includegraphics[clip,width=0.07\linewidth]{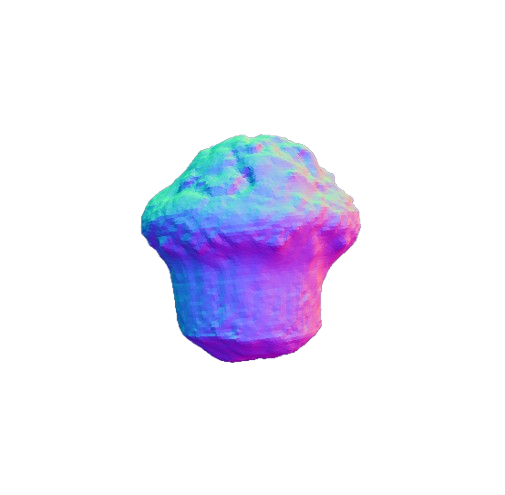} &
        \includegraphics[clip,width=0.07\linewidth]{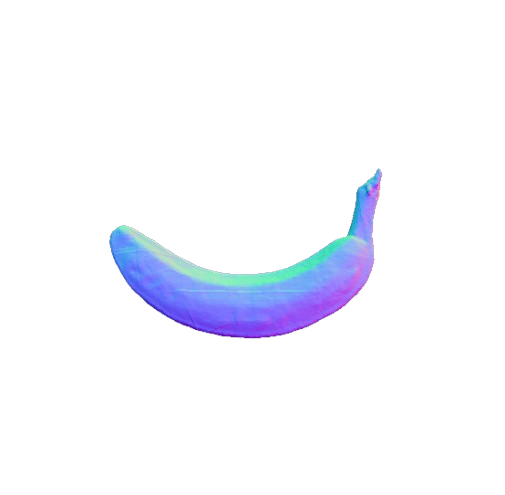} &
        \includegraphics[clip,width=0.07\linewidth]{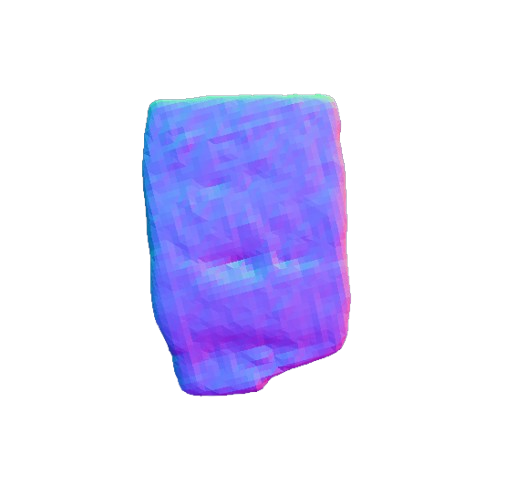} &
        \includegraphics[clip,width=0.07\linewidth]{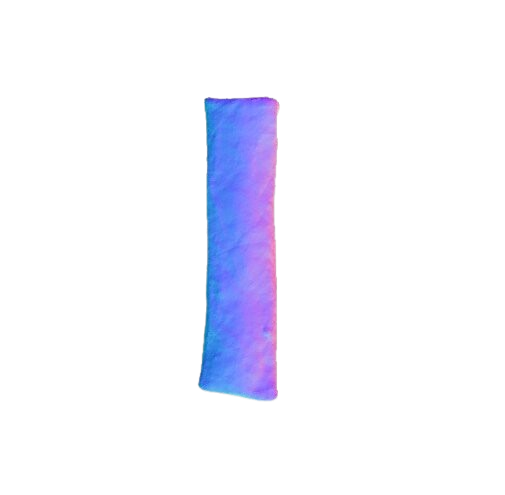} &
        \includegraphics[clip,width=0.07\linewidth]{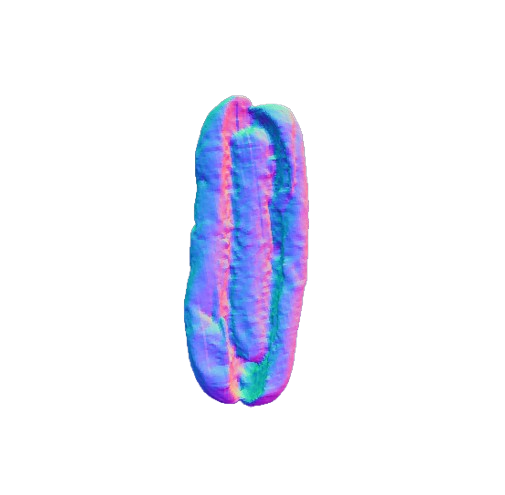}
        \\
        \raisebox{+0.2cm}{\rotatebox[origin=l]{90}{\footnotesize{Ours}}} &
        \includegraphics[clip,width=0.07\linewidth]{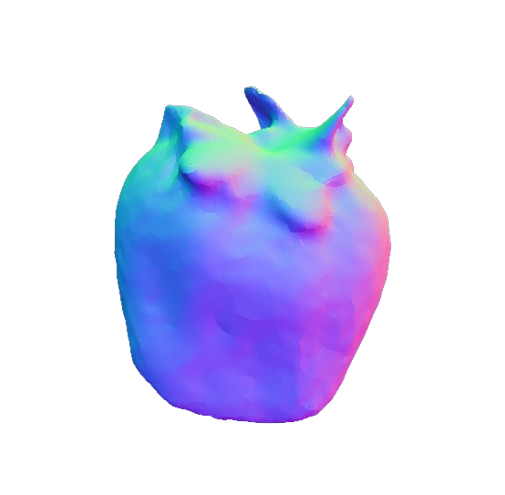} &
        \includegraphics[clip,width=0.07\linewidth]{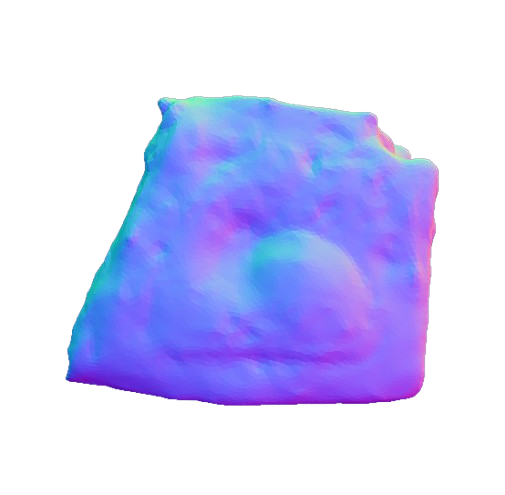} &
        \includegraphics[clip,width=0.07\linewidth]{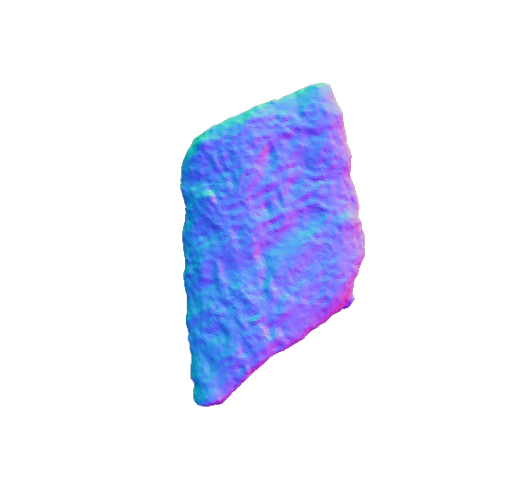} &
        \includegraphics[clip,width=0.07\linewidth]{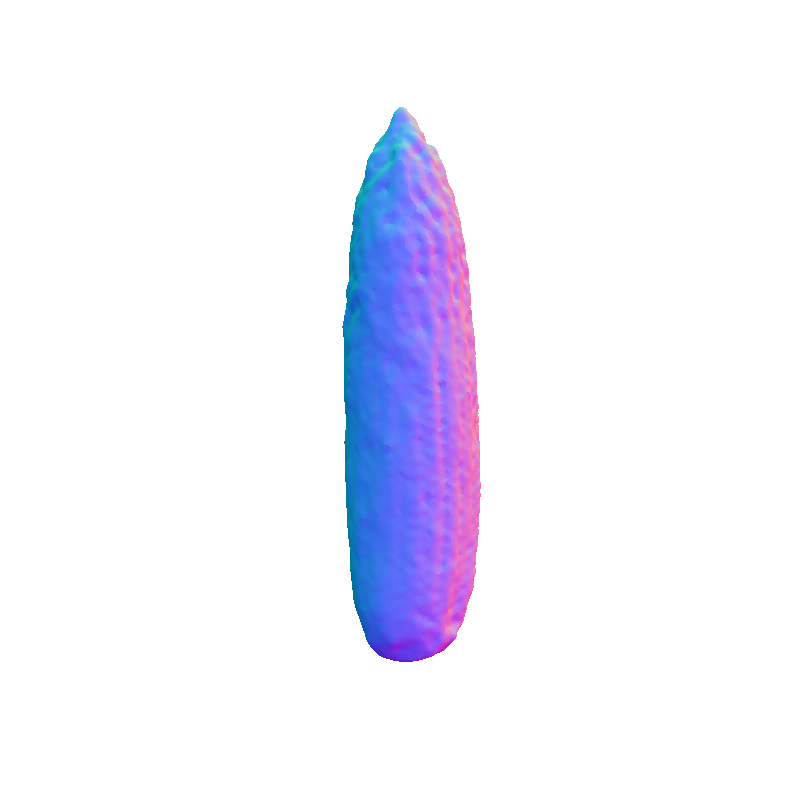} &
        \includegraphics[clip,width=0.07\linewidth]{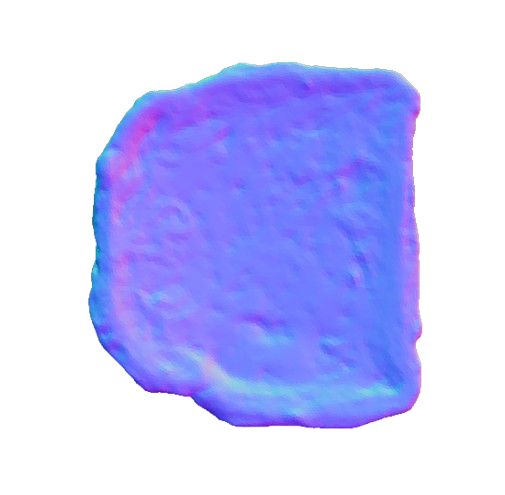} &
        \includegraphics[clip,width=0.07\linewidth]{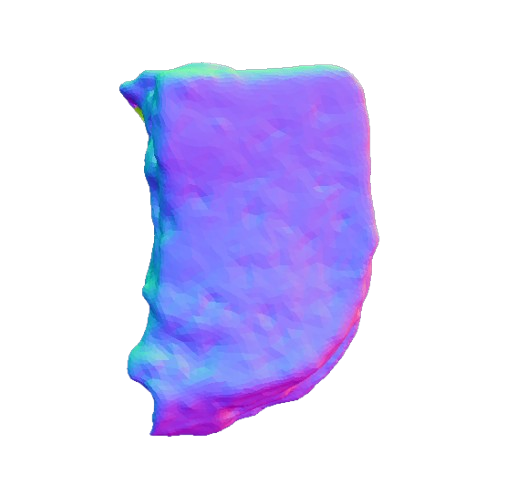} &
        \includegraphics[clip,width=0.07\linewidth]{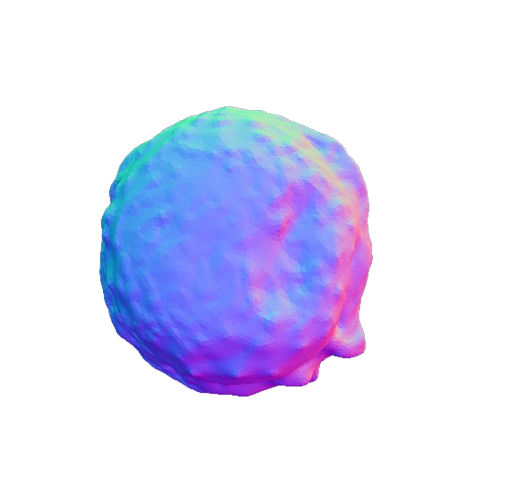} &
        \includegraphics[clip,width=0.07\linewidth]{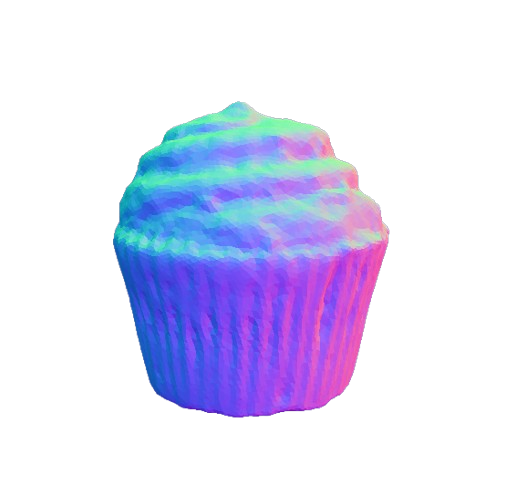} &
        \includegraphics[clip,width=0.07\linewidth]{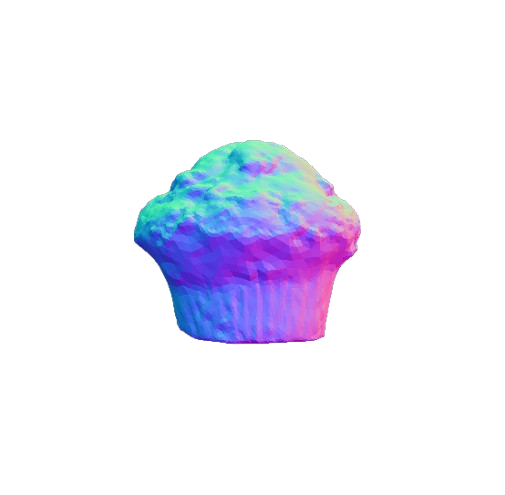} &
        \includegraphics[clip,width=0.07\linewidth]{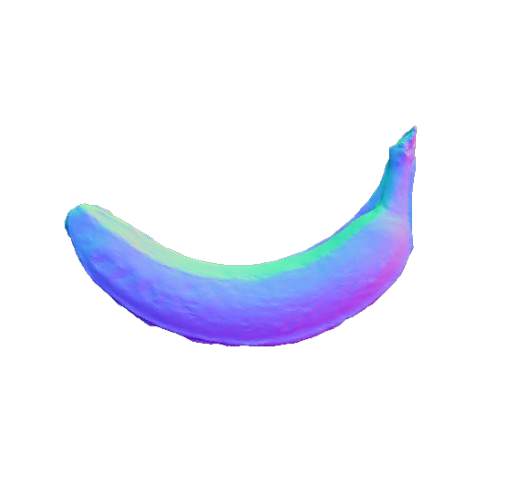} &
        \includegraphics[clip,width=0.07\linewidth]{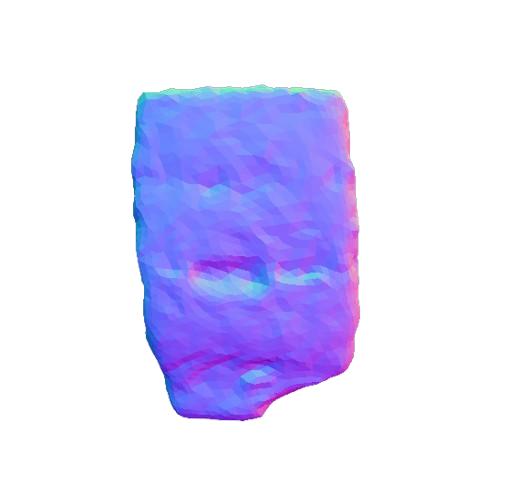} &
        \includegraphics[clip,width=0.07\linewidth]{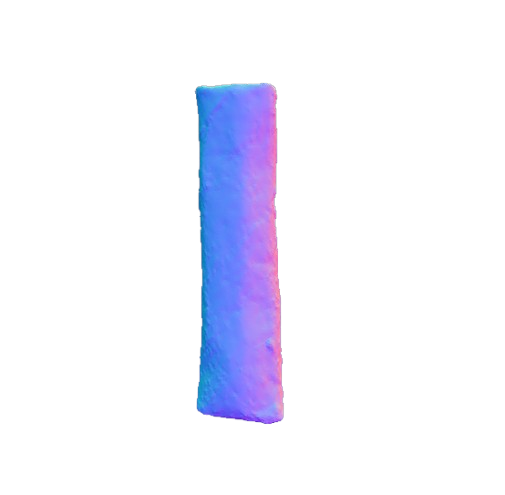} &
        \includegraphics[clip,width=0.07\linewidth]{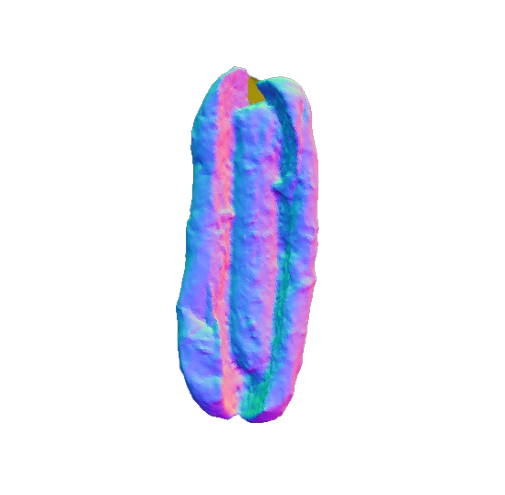}
        \\
        \hline
    \end{tabular}
    \caption{Our framework 3D reconstruction Visual Results on MTF dataset in comparison with VolETA~\cite{almughrabi2024voleta} and ground truth (GT).}
    \label{fig:Results_MTF_dataset}
\end{figure*}

\begin{figure*}[htb]
    \centering
    \setlength{\tabcolsep}{1pt}
    \renewcommand{\arraystretch}{1}
    \arrayrulecolor{black!30}
    \setlength\arrayrulewidth{0.3pt}
    \begin{tabular}{l|c|c|c|c|c|c|c}

        \raisebox{+1.5cm}{\rotatebox[origin=l]{90}{NeuS2}}
        &
        \begin{subfigure}[t]{0.14\linewidth}
            \centering
            \includegraphics[width=\textwidth]{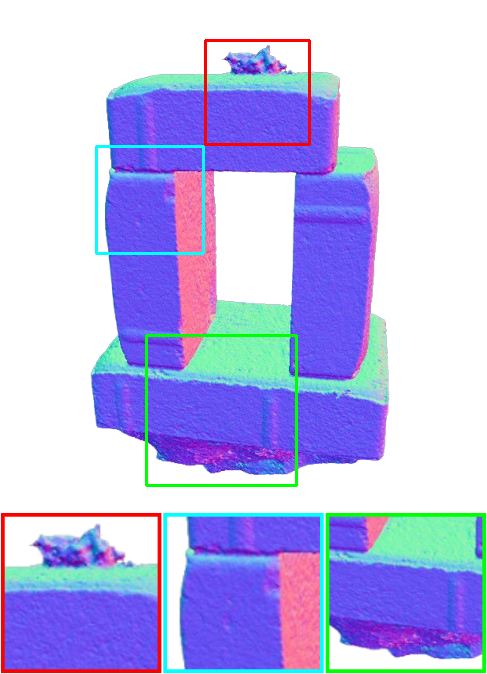}
        \end{subfigure} &
        \begin{subfigure}[t]{0.14\linewidth}
            \centering
            \includegraphics[width=\textwidth]{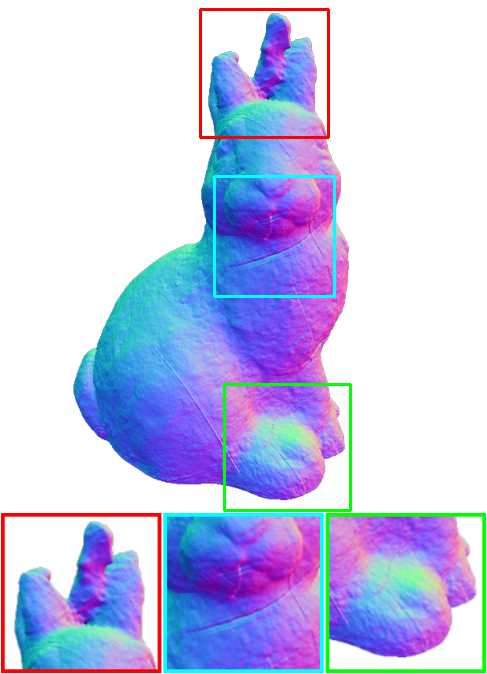}
        \end{subfigure} &
        \begin{subfigure}[t]{0.14\linewidth}
            \centering
            \includegraphics[width=\textwidth]{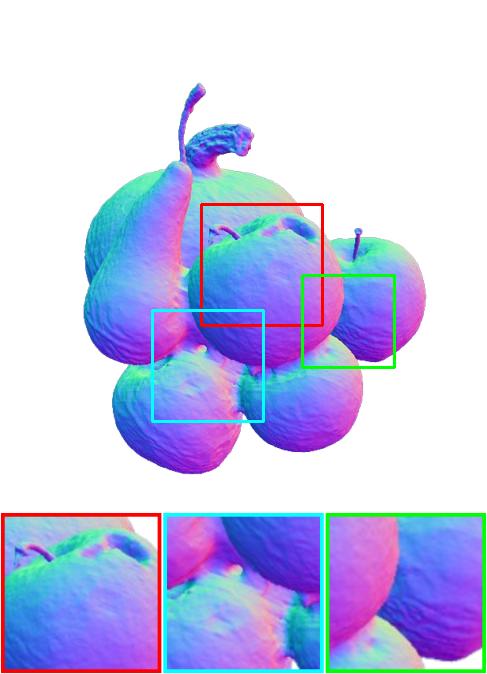}
        \end{subfigure} &
        \begin{subfigure}[t]{0.14\linewidth}
            \centering
            \includegraphics[width=\textwidth]{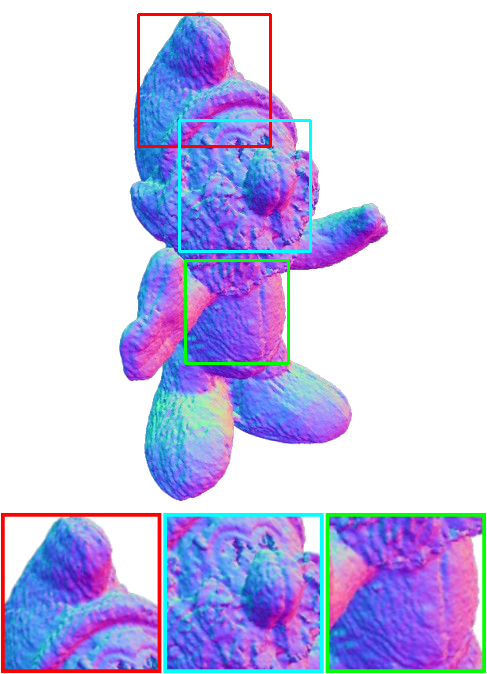}
        \end{subfigure} &
        \begin{subfigure}[t]{0.14\linewidth}
            \centering
            \includegraphics[width=\textwidth]{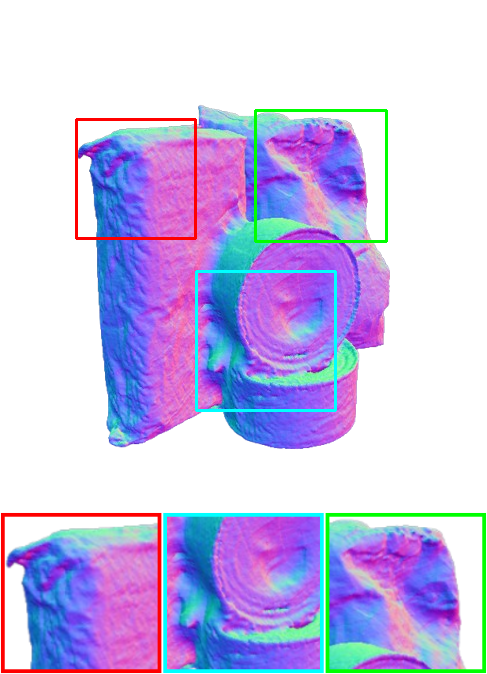}
        \end{subfigure} &
        \begin{subfigure}[t]{0.14\linewidth}
            \centering
            \includegraphics[width=\textwidth]{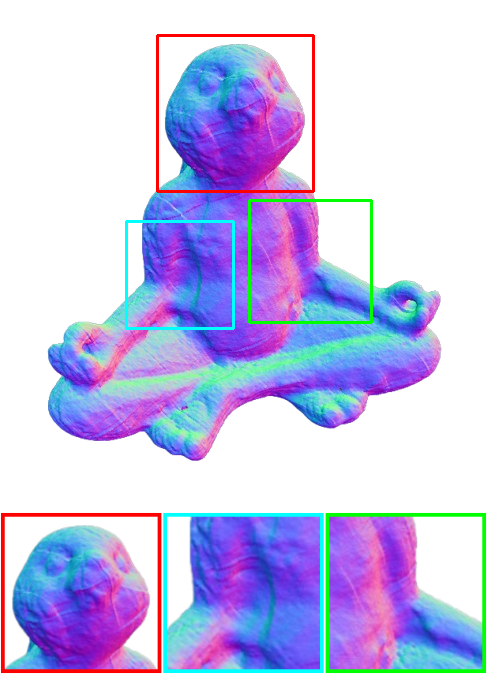}
        \end{subfigure} &
        \begin{subfigure}[t]{0.14\linewidth}
            \centering
            \includegraphics[width=\textwidth]{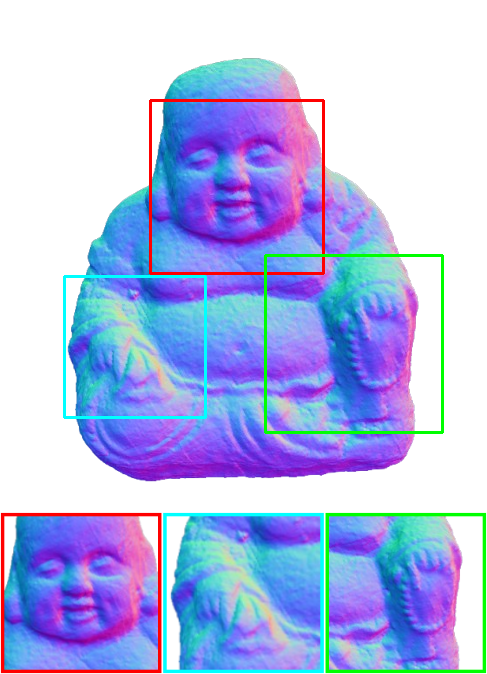}
        \end{subfigure} \\

        \raisebox{+1.5cm}{\rotatebox[origin=c]{90}{OUR}}
        &
        \begin{subfigure}[t]{0.14\linewidth}
            \centering
            \includegraphics[width=\textwidth]{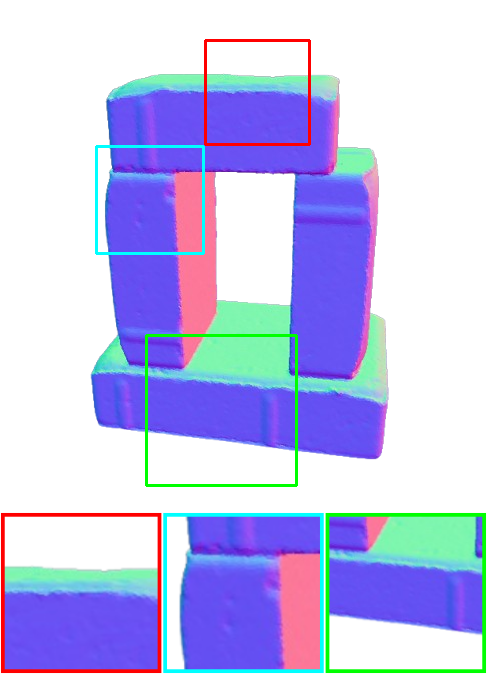}
            \caption{Scan 40}
        \end{subfigure} &
        \begin{subfigure}[t]{0.14\linewidth}
            \centering
            \includegraphics[width=\textwidth]{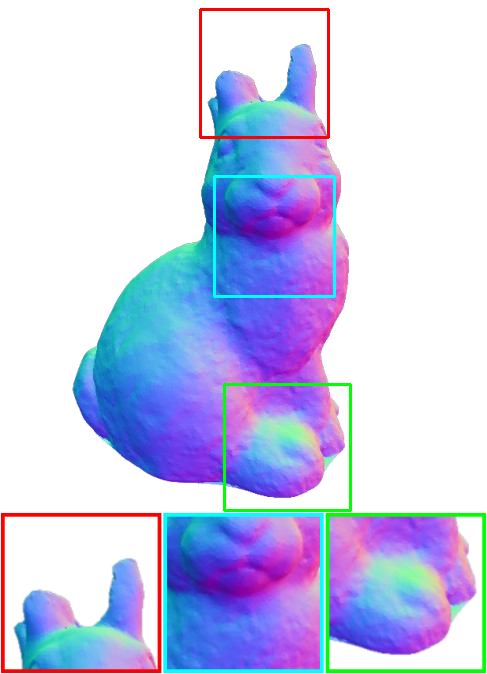}
            \caption{Scan 55}
        \end{subfigure} &
        \begin{subfigure}[t]{0.14\linewidth}
            \centering
            \includegraphics[width=\textwidth]{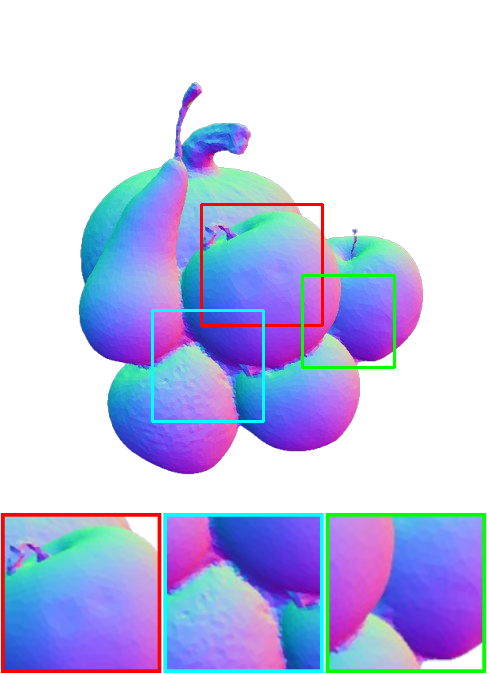}
            \caption{Scan 63}
        \end{subfigure} &
        \begin{subfigure}[t]{0.14\linewidth}
            \centering
            \includegraphics[width=\textwidth]{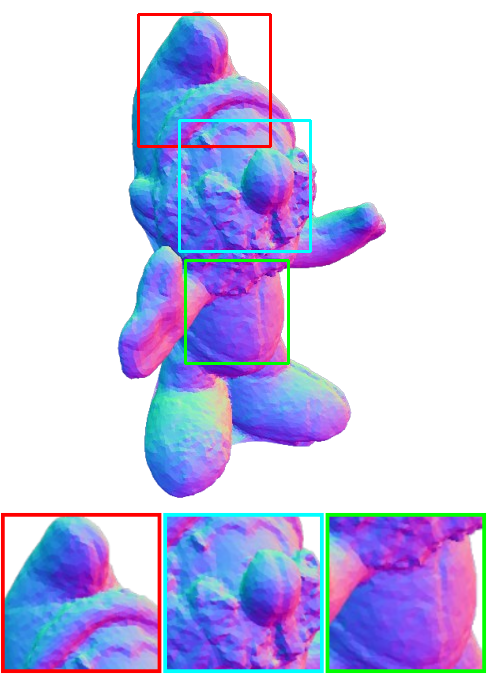}
            \caption{Scan 83}
        \end{subfigure} &
        \begin{subfigure}[t]{0.14\linewidth}
            \centering
            \includegraphics[width=\textwidth]{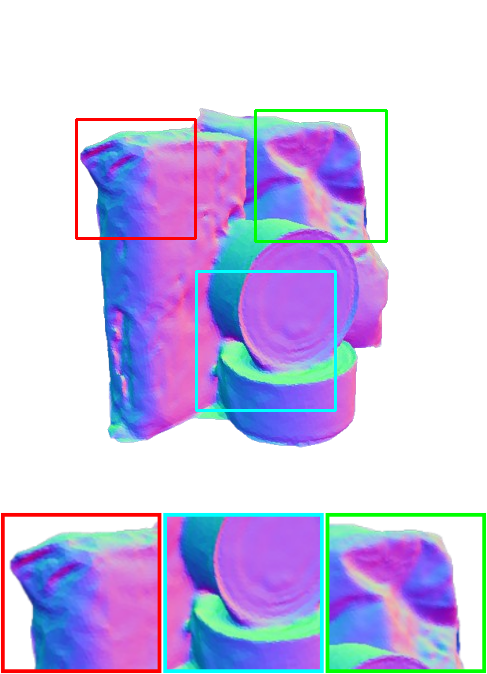}
            \caption{Scan 97}
        \end{subfigure} &
        \begin{subfigure}[t]{0.14\linewidth}
            \centering
            \includegraphics[width=\textwidth]{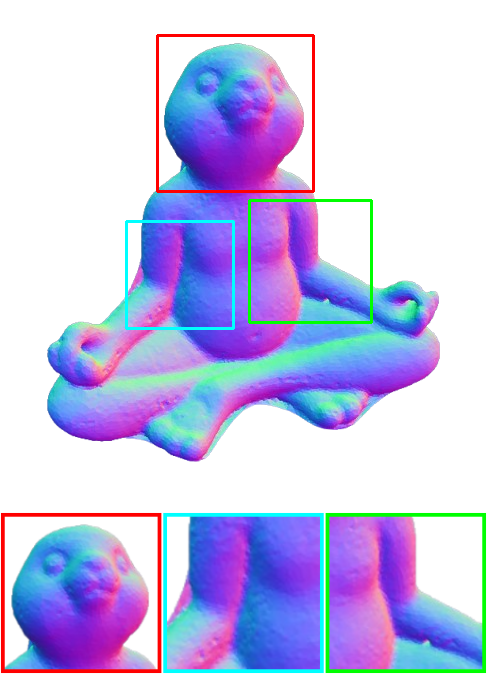}
            \caption{Scan 110}
        \end{subfigure} &
        \begin{subfigure}[t]{0.14\linewidth}
            \centering
            \includegraphics[width=\textwidth]{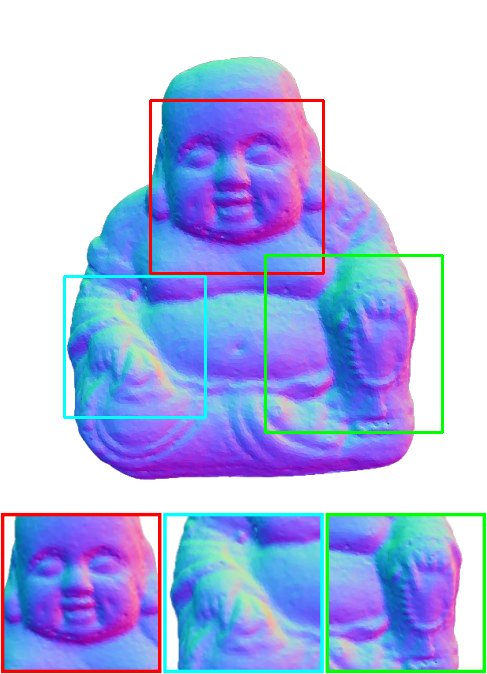}
            \caption{Scan 114}
        \end{subfigure}
    \end{tabular}
    \caption{Qualitative comparison of our framework in comparison with NeuS2~\cite{neus2} framework on DTU dataset.}
    \label{fig:Results_DTU_dataset}
\end{figure*}

Visual comparisons of 3D reconstructions and volume estimations highlight the superior performance of VolE in capturing intricate object details and maintaining geometric accuracy. Fig.~\ref{fig:Results_VolE_dataset} illustrates the qualitative results of the our framework as applied to the Foodkit dataset. It accurately reconstructs various food items, including smooth apples, pears, bananas, and textured items like donuts, durum wheat, and chocolate croissants. Notice how our framework effectively manages shape, size, and surface texture variations. Further validation is provided in Fig. \ref{fig:Results_MTF_dataset}, which presents a comparative analysis of ours against VolETA~\cite{almughrabi2024voleta} on the MTF dataset. The results clearly illustrate VolE's enhanced ability to capture the delicate geometries of various food items, resulting in more accurate 3D representations than VolETA. Finally, Fig. \ref{fig:Results_DTU_dataset} presents a qualitative comparison of ours with NeuS2~\cite{neus2} on the DTU dataset. Our framework exhibits superior performance with finer details and a well-reconstructed shape compared to the Neus2 method.

\subsubsection{Quantitative Results}
We report Mean Absolute Percentage Error (MAPE) values for volume estimation and Chamfer distances for 3D reconstruction accuracy. Our results show that our framework achieves an MAPE of 2.22\% on the MTF dataset, significantly outperforming existing methods.

Table \ref{tab:results_vole} presents a detailed quantitative comparison of volume estimations across various food items using the Foodkit dataset. The results demonstrate our framework's effectiveness in accurately estimating volumes across diverse food shapes and sizes, with a total accuracy of 98.8\% and a low mean error of 1.22\%.

\begin{table*}[htb]
    \small
    \centering
    \setlength{\tabcolsep}{0.1em}
    \caption{Quantitative comparison of volume estimations across various food items using the Foodkit dataset. The table reports mean estimated volumes, standard deviations, relative absolute errors, and accuracies for each item compared to ground truth values. These results validate the effectiveness of our framework in accurately estimating volumes across diverse food shapes and sizes.}

    \begin{tabular}{|c|l|c|c|cc|cc|c|}
    \hline
    \multirow{2}{*}{\textbf{Number}} & \multirow{2}{*}{\textbf{Items}} & \multirow{2}{*}{\textbf{Images}} & \multirow{2}{*}{\textbf{GT (±5)}} & \multicolumn{2}{c|}{\textbf{Estimated Volume (5x)}} & \multicolumn{2}{c|}{\textbf{Absolute Error (5x)}} & \multirow{2}{*}{\textbf{Accuracy}} \\
    \cline{5-8}
     &  &  &  & \textbf{Mean} & \textbf{Std. Dev} & \textbf{Mean} & \textbf{Std. Dev} & \\
    \hline
        1 & Apple & 1005 & 175 & 176.68 & 0.93 & 0.96 & 0.53 & 99.04 \\
        2 & Orange & 1001 & 200 & 201.06 & 3.45 & 1.35 & 1.02 & 98.65 \\
        3 & Aguacate & 1078 & 85 & 83.11 & 1.11 & 2.23 & 1.30 & 97.77 \\
        4 & Lemon & 887 & 140 & 134.74 & 3.19 & 1.76 & 1.33 & 98.24 \\
        5 & Donut & 780 & 245 & 242.24 & 2.57 & 1.13 & 1.04 & 98.87 \\
        6 & Durum & 1006 & 200 & 200.79 & 0.47 & 0.40 & 0.24 & 99.60 \\
        7 & Pear & 849 & 170 & 168.13 & 0.71 & 1.10 & 0.42 & 98.90 \\
        8 & Chocolate Cake & 781 & 195 & 195.38 & 1.21 & 0.38 & 0.50 & 99.62 \\
        9 & Chocolate Croissant & 1122 & 275 & 274.99 & 3.69 & 0.95 & 0.82 & 99.05 \\
        10 & Samosa & 848 & 145 & 144.10 & 2.76 & 1.53 & 1.10 & 98.47 \\
        11 & Apple Pie & 1201 & 135 & 135.52 & 1.02 & 0.51 & 0.66 & 99.49 \\
        12 & Chocolate Bomb & 1111 & 200 & 197.65 & 4.39 & 2.08 & 1.06 & 97.92 \\
        13 & Empanadilla & 926 & 95 & 94.86 & 1.12 & 0.89 & 0.65 & 99.11 \\
        14 & Falafel & 929 & 48 & 47.58 & 2.49 & 3.96 & 2.87 & 96.04 \\
        15 & French Bread & 1139 & 163 & 162.49 & 1.54 & 0.78 & 0.50 & 99.22 \\
        16 & Paxoco Mini & 911 & 150 & 148.08 & 1.62 & 1.40 & 0.89 & 98.60 \\
        17 & Napolitanas & 1071 & 233 & 232.79 & 1.28 & 0.41 & 0.32 & 99.59 \\
        18 & Capsicum & 881 & 320 & 318.64 & 2.54 & 0.76 & 0.35 & 99.24 \\
        19 & Mini Chocolate Panettone & 1209 & 293 & 290.79 & 1.74 & 0.75 & 0.59 & 99.25 \\
        20 & Banana & 1156 & 150 & 153.03 & 1.74 & 2.02 & 1.16 & 97.98 \\
        21 & Yellow Cane & 715 & 350 & 350.07 & 1.45 & 0.30 & 0.24 & 99.70 \\
    \hline
        \hline

        \multicolumn{4}{|c|}{\textbf{Total Error}} & \multicolumn{5}{c|}{\textbf{1.22}}  \\
        \cline{1-9}
        \multicolumn{4}{|c|}{\textbf{Total Accuracy}} & \multicolumn{5}{c|}{\textbf{98.78}} \\
        \cline{1-9}
        \hline
        
    \end{tabular}
    \label{tab:results_vole}
\end{table*}


Table \ref{tab:results_vole_MTF_volume} presents a comprehensive comparison of volume estimation methods on the MTF dataset, featuring VolETA, ININ, FoodR., and our proposed framework. The table showcases predicted volumes and percentage errors for 13 diverse food items, benchmarked against ground truth values. Our framework demonstrates superior performance, achieving the lowest MAPE of 3.08 \% and the smallest standard deviation of 2.63 \%. Our framework consistently outperforms other methods across various food items, securing the best or second-best results in 12 out of 13 cases. This performance is visually highlighted using color-coding, with red, yellow, and orange indicating the best, second-best, and third-best results, respectively. The table effectively illustrates our framework's enhanced accuracy and robustness in volume estimation tasks across a wide range of food types and shapes.

\begin{table*}[htb]
    \centering
    \caption{Comparative analysis of volume estimation and 3D reconstruction methods on the MTF dataset. The table presents predicted volumes, percentage errors, and Chamfer Distance (CD) metrics for VolETA~\cite{almughrabi2024voleta}, ININ~\cite{he2024metafood}, FoodR.~\cite{he2024metafood}, and our proposed VolE framework across various food items. Lower percentage errors and CD values indicate better performance. Best results are highlighted in red, second-best in yellow, and third-best in orange to showcase our method's superior accuracy in volume estimation and 3D reconstruction tasks.}
    
    \label{tab:results_vole_MTF_volume}
    \resizebox{\textwidth}{!}{%
        \begin{tabular}{|c|l|cccc|c|cccc|cccc|}
            \hline
                \multirow{2}{*}{\textbf{ID}} & \multirow{2}{*}{\textbf{Scene Name}} & \multicolumn{4}{c|}{\textbf{Predicted Volume}} & \multirow{2}{*}{\textbf{GT}} & \multicolumn{4}{c|}{\textbf{Error Percentage \(\downarrow\)}} & \multicolumn{4}{c|}{\textbf{Chamfer Distance \(\downarrow\)}} \\
                \cline{3-6}\cline{8-11}
                & & \textbf{VolETA~\cite{almughrabi2024voleta}} & \textbf{ININ~\cite{he2024metafood}} & \textbf{FoodR.~\cite{he2024metafood}} & \textbf{Our} & & \textbf{VolETA~\cite{almughrabi2024voleta}} & \textbf{ININ~\cite{he2024metafood}} & \textbf{FoodR.~\cite{he2024metafood}} & \textbf{Our} & \textbf{VolETA~\cite{almughrabi2024voleta}} & \textbf{ININ~\cite{he2024metafood}} & \textbf{FoodR.~\cite{he2024metafood}} & \textbf{Our} \\
                
                \hline
                
                1 & Strawberry & 40.06 & 37.65 & 44.51 & 37.47 & 38.53 & \cellcolor{orange!25}3.97 & \cellcolor{red!25}2.28 & 15.52 & \cellcolor{yellow!25}2.74 & \cellcolor{yellow!25}0.0016 & \cellcolor{orange!25}0.0020 & \cellcolor{red!25}0.0011 & 0.0028 \\
                2 & Cinnamon bun & 216.90 & 325.44 & 321.26 & 275.38 & 280.36 & 22.64 & \cellcolor{orange!25}16.08 & \cellcolor{yellow!25}14.59 & \cellcolor{red!25}1.78 & 0.0071 & \cellcolor{orange!25}0.0036 & \cellcolor{yellow!25}0.0031 & \cellcolor{red!25}0.0022 \\
                3 & Pork rib & 278.86 & 473.40 & 336.11 & 268.93 & 249.65 & \cellcolor{yellow!25}11.70 & 89.63 & \cellcolor{orange!25}34.63 & \cellcolor{red!25}7.72 & 0.0137 & \cellcolor{red!25}0.0049 & \cellcolor{yellow!25}0.0053 & \cellcolor{orange!25}0.0068 \\
                4 & Corn & 279.02 & 294.32 & 347.54 & 277.56 & 295.13 & \cellcolor{yellow!25}5.46 & \cellcolor{red!25}0.27 & 17.76 & \cellcolor{orange!25}5.95 & \cellcolor{yellow!25}0.0020 & \cellcolor{orange!25}0.0038 & \cellcolor{red!25}0.0015 & 0.0046 \\
                5 & French toast & 395.76 & 353.66 & 389.28 & 394.04 & 392.58 & \cellcolor{yellow!25}0.81 & 9.91 & \cellcolor{orange!25}0.84 & \cellcolor{red!25}0.37 & 0.0137 & \cellcolor{red!25}0.0020 & \cellcolor{orange!25}0.0040 & \cellcolor{yellow!25}0.0021 \\
                6 & Sandwich & 205.17 & 237.88 & 197.82 & 215.21 & 218.31 & \cellcolor{yellow!25}6.02 & \cellcolor{orange!25}8.96 & 9.39 & \cellcolor{red!25}1.42 & 0.0067 & \cellcolor{yellow!25}0.0038 & \cellcolor{red!25}0.0025 & \cellcolor{orange!25}0.0039 \\
                7 & Burger & 372.93 & 361.49 & 412.52 & 370.69 & 368.77 & \cellcolor{yellow!25}1.13 & \cellcolor{orange!25}1.97 & 11.86 & \cellcolor{red!25}0.52 & \cellcolor{orange!25}0.0047 & 0.0048 & \cellcolor{red!25}0.0025 & \cellcolor{yellow!25}0.0036 \\
                8 & Cake & 186.62 & 172.32 & 181.21 & 176.43 & 173.13 & 7.79 & \cellcolor{red!25}0.47 & \cellcolor{orange!25}4.67 & \cellcolor{yellow!25}1.91 & 0.0030 & \cellcolor{orange!25}0.0019 & \cellcolor{red!25}0.0010 & \cellcolor{yellow!25}0.0012 \\
                9 & Blueberry muffin & 224.08 & 253.01 & 233.79 & 233.95 & 232.74 & \cellcolor{orange!25}3.72 & 8.71 & \cellcolor{yellow!25}0.45 & \cellcolor{red!25}0.52 & 0.0039 & \cellcolor{yellow!25}0.0029 & \cellcolor{orange!25}0.0033 & \cellcolor{red!25}0.0029 \\
                10 & Banana & 153.76 & 157.58 & 160.06 & 159.20 & 163.23 & 5.80 & \cellcolor{orange!25}3.46 & \cellcolor{red!25}1.94 & \cellcolor{yellow!25}2.47 & \cellcolor{yellow!25}0.0027 & \cellcolor{orange!25}0.0034 & \cellcolor{red!25}0.0019 & 0.0118 \\
                11 & Salmon & 80.40 & 76.46 & 86.00 & 82.75 & 85.18 & \cellcolor{orange!25}5.61 & 10.24 & \cellcolor{red!25}0.96 & \cellcolor{yellow!25}2.85 & 0.0034 & \cellcolor{red!25}0.0015 & \cellcolor{yellow!25}0.0015 & \cellcolor{orange!25}0.0021 \\
                13 & Burrito & 363.99 & 246.60 & 334.70 & 297.09 & 308.28 & \cellcolor{yellow!25}18.07 & 20.01 & \cellcolor{yellow!25}8.57 & \cellcolor{red!25}3.63 & \cellcolor{orange!25}0.0052 & \cellcolor{red!25}0.0026 & \cellcolor{yellow!25}0.0041 & 0.0055 \\
                14 & Hotdog & 535.44 & 495.10 & 517.75 & 541.58 & 589.82 & \cellcolor{yellow!25}9.22 & 16.06 & \cellcolor{orange!25}12.22 & \cellcolor{red!25}8.18  & \cellcolor{red!25}0.0043 & \cellcolor{yellow!25}0.0044 & \cellcolor{orange!25}0.0046 & 0.0082 \\
                
                \hline
                
                \hline
                
                \multicolumn{2}{|c|}{\textbf{MAPE \(\downarrow\)}} & \cellcolor{yellow!25}7.84 & 14.47 & \cellcolor{orange!25}10.26 & \cellcolor{red!25}3.08 & \textbf{S.D. \(\downarrow\)} &  \cellcolor{yellow!25}6.36 &  23.47 &  \cellcolor{orange!25}9.48 &  \cellcolor{red!25}2.63 & &  & &  \\

                \hline
                
                \multicolumn{2}{|c|}{\textbf{CD (Sum) \(\downarrow\)}} & & & & & & & & & & 0.0720 &  \cellcolor{yellow!25}0.0416 &  \cellcolor{red!25}0.0364 &  \cellcolor{orange!25}0.0576  \\

                \hline
                
                \multicolumn{2}{|c|}{\textbf{CD (Mean) \(\downarrow\)}} & & & & & & & & & & 0.0055 & \cellcolor{yellow!25}0.0032 & \cellcolor{red!25}0.0028 & \cellcolor{orange!25}0.0044  \\

            \hline
        \end{tabular}%
    }
\end{table*}


Table \ref{tab:results_vole_MTF_volume} presents a comparative analysis of 3D reconstruction accuracy on the MTF dataset using Chamfer Distance (CD) metrics. The table shows CD values for VolETA~\cite{almughrabi2024voleta}, ININ~\cite{he2024metafood}, FoodR.~\cite{he2024metafood}, and our proposed method across various food scenes. Lower CD values indicate better performance, with the best, second-best, and third-best results highlighted in red, yellow, and orange, respectively.

Although VolETA~\cite{almughrabi2024voleta} is the winner of the MTF dataset challenge and the state-of-the-art method for estimating food volume for the MTF dataset, our approach demonstrates a competitive performance across different metrics. VolETA excels in volume estimation but falls short in CD measurements. Conversely, ININ and FoodR. Show high MAPE and standard deviation for volume estimation but perform well in CD metrics. Our method strikes a balance, outperforming VolETA in both volume estimation, MAPE, and CD for 3D reconstruction. We achieve the third-best result in overall CD (Mean: 0.0044) while still surpassing VolETA CD (Mean: 0.0055). This multifaceted performance underscores our method's robustness in both volume estimation accuracy and geometric reconstruction quality, positioning it as a strong contender in food volume estimation and 3D reconstruction tasks.

Table \ref{tab:comparison_DTU} (DTU dataset) compares VolE against several state-of-the-art methods, including COLMAP, NeuS, InstantNGP, InstantNSR, and NeuS2. VolE demonstrates competitive performance, achieving the lowest mean Chamfer Distance (0.68) while maintaining efficient processing times comparable to the fastest methods.

\begin{table*}[htb]
\centering
\caption{Performance comparison on the DTU dataset among various methods including COLMAP~\cite{schoenberger2016sfm}, NeuS~\cite{wang2021neus}, InstantNGP~\cite{mueller2022instant}, InstantNSR~\cite{zhao2022human}, NeuS2~\cite{neus2}, and our proposed VolE framework. The table shows Chamfer Distance (CD) metrics indicating geometric reconstruction accuracy, with our method achieving competitive results while maintaining efficient processing times. The best results are highlighted in red, second-best in yellow and third-best in orange to facilitate a quick assessment of performance across different methods.}
\label{tab:comparison_DTU}
\resizebox{0.8\textwidth}{!}{%
\begin{tabular}{|c|c|c|c|c|c|c|}
\hline
\textbf{Scan-ID} & \textbf{COLMAP~\cite{schoenberger2016sfm}} & \textbf{NeuS~\cite{wang2021neus}} & \textbf{InstantNGP~\cite{mueller2022instant}} & \textbf{InstantNSR~\cite{zhao2022human}} & \textbf{NeuS2~\cite{neus2}} & \textbf{Our} \\
\hline
Scan24 & \cellcolor{orange!25}0.81 & 0.83 & 1.68 & 2.86 & \cellcolor{red!25}0.56 & \cellcolor{yellow!25}0.59 \\
Scan37 & 2.05 & \cellcolor{orange!25}0.98 & 1.93 & 2.81 & \cellcolor{yellow!25}0.76 & \cellcolor{red!25}0.74 \\
Scan40 & 0.73 & \cellcolor{orange!25}0.56 & 1.57 & 2.09 & \cellcolor{yellow!25}0.49 & \cellcolor{red!25}0.40 \\
Scan55 & 1.22 & \cellcolor{orange!25}0.37 & 1.16 & 0.81 & \cellcolor{yellow!25}0.37 & \cellcolor{red!25}0.33 \\
Scan63 & 1.79 & \cellcolor{orange!25}1.13 & 2.00 & 1.65 & \cellcolor{red!25}0.92 & \cellcolor{yellow!25}1.08 \\
Scan65 & 1.58 & \cellcolor{red!25}0.59 & 1.56 & 1.39 & \cellcolor{yellow!25}0.71 & \cellcolor{orange!25}0.86 \\
Scan69 & 1.02 & \cellcolor{yellow!25}0.60 & 1.81 & 1.47 & \cellcolor{orange!25}0.76 & \cellcolor{red!25}0.55 \\
Scan83 & 3.05 & \cellcolor{orange!25}1.45 & 2.33 & 1.67 & \cellcolor{yellow!25}1.22 & \cellcolor{red!25}1.10 \\
Scan97 & 1.40 & \cellcolor{red!25}0.95 & 2.16 & 2.47 & \cellcolor{yellow!25}1.08 & \cellcolor{orange!25}1.11 \\
Scan105 & 2.05 & \cellcolor{orange!25}0.78 & 1.88 & 1.12 & \cellcolor{red!25}0.63 & \cellcolor{yellow!25}0.70 \\
Scan106 & 1.00 & \cellcolor{red!25}0.52 & 1.76 & 1.22 & \cellcolor{orange!25}0.59 & \cellcolor{yellow!25}0.59 \\
Scan110 & \cellcolor{orange!25}1.32 & 1.43 & 2.32 & 2.30 & \cellcolor{yellow!25}0.89 & \cellcolor{red!25}0.71 \\
Scan114 & 0.49 & \cellcolor{red!25}0.36 & 1.86 & 0.98 & \cellcolor{yellow!25}0.40 & \cellcolor{orange!25}0.41 \\
Scan118 & 0.78 & \cellcolor{red!25}0.45 & 1.80 & 1.41 & \cellcolor{yellow!25}0.48 & \cellcolor{orange!25}0.58 \\
Scan122 & 1.17 & \cellcolor{yellow!25}0.45 & 1.72 & 0.95 & \cellcolor{orange!25}0.55 & \cellcolor{red!25}0.41 \\
\hline
\hline
CD (Mean) \(\downarrow\) & 1.36 & \cellcolor{orange!25}0.77 & 1.84 & 1.68 & \cellcolor{yellow!25}0.70 & \cellcolor{red!25}0.68 \\
\hline
Runtime & 1 h & 8 h & 5 min & 8.5 min & 5 min & 3.5 min \\
\hline
\end{tabular}%
}
\end{table*}


\subsection{Ablation Study}

To further assess the performance of our framework, we conducted an ablation study to identify the minimum number of images required for accurate volume estimation while reducing computational time. The primary goal of this analysis is to establish an optimal trade-off between estimation accuracy and computational efficiency. To achieve this, we utilized two techniques: (i) Frame-Skipping to reduce the total number of frames processed, and (ii) Hamming Distance-based image selection to maintain diversity among the selected frames. By systematically reducing the number of images used in volume estimation, we assessed the impact of these techniques on performance across various configurations.

\subsubsection{Frame Skipping Experiments}
We first investigated the impact of frame skipping on volume estimation accuracy. Frame skipping intervals of 3, 5, 10, 11, 12, 13, 15, and 20 were evaluated. Each experiment was repeated five times using identical image sets to ensure consistency in the results. Table \ref{tab:apple_five} illustrates the consistency and reliability of our method using 1005 images for an apple scene (ground truth volume: 175 ml). The low standard deviation in estimated volume (0.93 ml) and relative absolute error (0.53\%) across repetitions demonstrates the robustness of our approach. Table \ref{tab:apple_frameskip} presents the impact of frame skipping on estimation accuracy and processing time. As the number of images decreases, we observe a clear trend of increasing mean error and standard deviation. For instance, skipping 5 frames (201 images) yields a mean absolute error of 0.23 ml with 99.77\% accuracy, while skipping 20 frames (50 images) results in a significantly higher mean absolute error of 19.81 ml with 80.19\% accuracy. This analysis reveals a critical trade-off between computational efficiency and estimation accuracy in our volume reconstruction process.

\begin{table}[htbp]
\centering
\caption{Volume estimation results for an apple scene (ground truth: 175 ml) using our framework. The table shows five repetitions of the experiment using 1005 images, demonstrating the consistency and reliability of the method.}
\label{tab:apple_five}
\resizebox{\columnwidth}{!}{%
\begin{tabular}{|c|c|c|c|c|c|}
\hline
\multicolumn{6}{|c|}{\textbf{Scene:} Apple, \textbf{Ground Truth:} 175} \\
\hline
\hline
\textbf{Repetition} & \textbf{Images} & \textbf{Volume} & \textbf{Time} & \textbf{RAE (\%)} & \textbf{Accuracy} \\
\hline
1 & 1005 & 177.49 & 8m 52.207s & 1.42 & 98.58 \\
2 & 1005 & 175.29 & 9m 08.282s & 0.17 & 99.83 \\
3 & 1005 & 177.58 & 8m 38.458s & 1.47 & 98.53 \\
4 & 1005 & 176.63 & 8m 52.717s & 0.93 & 99.07 \\
5 & 1005 & 176.40 & 8m 47.020s & 0.80 & 99.20 \\
\hline
\hline
\multicolumn{2}{|c|}{Mean} & 176.68 & 8m 51.737s & 0.96 & 99.04 \\
\hline
\multicolumn{2}{|c|}{Std Dev} & 0.93 & 9.731 s & 0.53 & 0.53 \\
\hline
\end{tabular}%
}
\end{table}


\begin{table}[htbp]
\centering
\setlength{\tabcolsep}{0.1em}
    \caption{Impact of frame skipping on volume estimation accuracy for an apple scene using VolE framework. The table illustrates the trade-off between computational efficiency and estimation accuracy as the number of images decreases.}
\label{tab:apple_frameskip}
\resizebox{\columnwidth}{!}{%
\begin{tabular}{|c|c|cc|c|cc|c|}
\hline
\multicolumn{8}{|c|}{\textbf{Scene:} Apple, \textbf{Ground Truth:} 175} \\
\hline
\hline
\multirow{2}{*}{\textbf{Frame Skip}} & \multirow{2}{*}{\textbf{Images}} & \multicolumn{2}{c|}{\textbf{Estimated Volume (5x)}} & \multirow{2}{*}{\textbf{Time}} & \multicolumn{2}{c|}{\textbf{Absolute Error (5x)}} & \multirow{2}{*}{\textbf{Accuracy}} \\
\cline{3-4}\cline{6-7}
& & \textbf{Mean} & \textbf{Std Dev} & & \textbf{Mean} & \textbf{Std Dev} & \\
\hline
all & 1005 & 176.68 & 0.93 & 8m 52.717s & 0.96 & 0.53 & 99.04 \\
3 & 335 & 176.95 & 0.35 & 9m 07.736s & 1.11 & 0.20 & 98.89 \\
5 & 201 & 175.37 & 0.57 & 2m 28.260s & 0.23 & 0.31 & 99.77 \\
10 & 100 & 172.61 & 2.09 & 1m 01.020s & 1.59 & 0.78 & 98.41 \\
11 & 92 & 169.65 & 3.59 & 0m 57.717s & 3.22 & 1.71 & 96.78 \\
12 & 83 & 171.38 & 1.29 & 0m 42.459s & 2.07 & 0.74 & 97.93 \\
13 & 78 & 165.67 & 1.39 & 0m 37.072s & 5.33 & 0.80 & 94.67 \\
15 & 68 & 168.21 & 2.14 & 0m 27.924s & 3.88 & 1.22 & 96.12 \\
20 & 50 & 140.34 & 18.01 & 0m 17.636s & 19.81 & 10.29 & 80.19 \\
\hline
\end{tabular}%
}
\end{table}

\subsubsection{Hamming Distance Experiments}
To overcome the limitations of frame skipping and optimize our feature-matching process, we conducted experiments using Hamming distance thresholds. This approach allows for the selective removal of similar images while retaining frames with significant visual differences. We systematically tested Hamming distance thresholds from 12 to 1, decreasing by one at each step. Table \ref{tab:apple_hamming} demonstrates the effects of different Hamming distance thresholds on the number of images used, processing time, and estimation accuracy. The results show that a Hamming distance threshold of 2 achieves the best balance between accuracy and efficiency, using 486 images and yielding a mean absolute error of 0.07 ml with 99.93\% accuracy. This represents a significant improvement over the frame-skipping method, which required 1005 images to achieve comparable accuracy.


\begin{table}[htbp]
\centering
\setlength{\tabcolsep}{0.1em}
    \caption{Effect of Hamming distance threshold on volume estimation accuracy for an apple scene using VolE framework. The table demonstrates how different Hamming distance thresholds affect the number of images used, processing time, and estimation accuracy.}
    \label{tab:apple_hamming}
\resizebox{\columnwidth}{!}{%
\begin{tabular}{|c|c|cc|c|cc|c|}
\hline
\multicolumn{8}{|c|}{\textbf{Scene:} Apple, \textbf{Ground Truth:} 175} \\
\hline
\hline
\multirow{2}{*}{\textbf{HD}} & \multirow{2}{*}{\textbf{Images}} & \multicolumn{2}{c|}{\textbf{Estimated Volume (5x)}} & \multirow{2}{*}{\textbf{Time}} & \multicolumn{2}{c|}{\textbf{Absolute Error (5x)}} & \multirow{2}{*}{\textbf{Accuracy}} \\
\cline{3-4}\cline{6-7}
& & \textbf{Mean} & \textbf{Std Dev} & & \textbf{Mean} & \textbf{Std Dev} & \\
\hline
1 & 776 & 175.92 & 0.91 & 11m 3.549s & 0.53 & 0.52 & 99.47 \\
2 & 486 & 175.12 & 0.61 & 5m 6.942s & 0.07 & 0.35 & 99.93 \\
3 & 486 & 175.54 & 0.55 & 5m 21.006s & 0.35 & 0.26 & 99.65 \\
4 & 329 & 173.7 & 0.63 & 3m 27.466s & 0.74 & 0.36 & 99.26 \\
5 & 329 & 174.9 & 0.73 & 3m 22.393s & 0.33 & 0.21 & 99.67 \\
6 & 235 & 172.56 & 1.94 & 2m 6.564s & 1.39 & 1.11 & 98.61 \\
7 & 235 & 171.71 & 1.54 & 2m 7.616s & 1.88 & 0.88 & 98.12 \\
8 & 160 & 169.63 & 0.74 & 1m 15.485s & 3.07 & 0.42 & 96.93 \\
9 & 160 & 170.02 & 0.48 & 1m 16.59s & 2.85 & 0.28 & 97.15 \\
10 & 118 & 170.73 & 1.07 & 0m 49.184s & 2.44 & 0.61 & 97.56 \\
11 & 118 & 170.24 & 2.15 & 0m 51.35s & 2.72 & 1.23 & 97.28 \\
12 & 82 & 166.76 & 1.74 & 0m 31.562s & 4.71 & 1.00 & 95.29 \\
\hline
\end{tabular}%
}
\end{table}

\subsubsection{Observation}
Our ablation study reveals that both frame skipping and Hamming distance-based image selection can significantly reduce computational time while maintaining high accuracy. However, the Hamming distance approach demonstrates superior performance, allowing for more intelligent frame selection based on visual content rather than fixed intervals. The optimal configuration depends on the specific requirements of the application. For scenarios demanding the highest accuracy, using all available frames (1005 images) provides the most reliable results. In contrast, applications prioritizing speed might opt for a Hamming distance threshold of 10 or 11, which reduces processing time by over 90\% while maintaining accuracy above 97\%. These findings underscore the flexibility of our framework, enabling users to tailor the image selection process to their specific needs, balancing accuracy and computational efficiency.

\subsection{Discussion}

The experimental results provide strong evidence for the effectiveness of VolE in accurately estimating object volumes across various scenarios. VolE's ability to handle low-textured objects and unbounded scenes contributes to its superior performance across diverse food items. Leveraging the AR-capable devices to gather the images and camera locations and FoodMem for food video segmentation is a crucial factor in achieving high accuracy, as shown in Fig. \ref{fig:methodology}(a). Frame sampling experiments reveal that our framework maintains robust performance even with fewer input frames, indicating its efficiency in real-world applications. The Hamming distance analysis provides insights into the optimal threshold for feature matching, balancing accuracy and computational cost. Moreover, our framework consistently outperforms state-of-the-art methods in both volume estimation accuracy (lower MAPE) and 3D reconstruction quality (lower CD), as shown in Table \ref{tab:results_vole_MTF_volume} and  \ref{tab:comparison_DTU}. The framework's strong performance across multiple datasets (Foodkit, MTF, DTU) demonstrates its versatility and potential for broad applicability in various domains requiring accurate volume estimation and 3D reconstruction. These findings underscore our framework's potential as a robust, efficient, and accurate solution for volume estimation and 3D reconstruction tasks, particularly in the context of food volume estimation for dietary assessment applications.

\subsection{Limitations} 

The VolE framework, despite its promising results, faces limitations primarily regarding real-time speed, which is crucial for real-time volume estimation applications. Additionally, adapting the framework for edge devices, such as mobile phones, will enable broader accessibility and usability, allowing users to leverage real-time volume estimation capabilities directly on their personal devices. 

%% file: 5_conclusion.tex
\section{Conclusion}
\label{sec:conclusion}

This paper introduces \textbf{VolE}, a reference- and depth-free framework for precise volume estimation of free-motion food scenes using images and camera locations from mobile devices. \textbf{VolE} leverages ARCore/ARKit and SfM to reconstruct accurately scaled 3D point clouds without reference objects or single-purpose hardware. We also present the Foodkit dataset, a benchmark of 21 food objects with ground truth volumes and masses, addressing the lack of high-quality benchmarks in food volume estimation. The experimental results emphasize that our framework shows superior performance, positioning VolE as a state-of-the-art approach for food volume estimation, offering practical solutions for real-world applications. Future work includes extending VolE to complex food scenarios, integrating with nutritional databases, optimizing the speed of camera pose estimation and 3D reconstruction for real-time applications, and developing algorithms to improve the detection and segmentation of transparent and reflective food items. 

%% file: 4_ack.tex
\section*{Acknowledgment}

This work was partially funded by the EU project MUSAE (No. 01070421), 2021-SGR-01094 (AGAUR), Icrea Academia’2022 (Generalitat de Catalunya), Robo STEAM (2022-1-BG01-KA220-VET000089434, Erasmus+ EU), DeepSense (ACE053/22/000029, ACCIÓ), CERCA Programme/Generalitat de Catalunya, and Grants PID2022141566NB-I00 (IDEATE), PDC2022-133642-I00 (DeepFoodVol), and CNS2022-135480 (A-BMC) funded by MICIU/AEI/10.13039/501100 011033, by FEDER (UE), and by European Union NextGenerationEU/ PRTR. A. AlMughrabi acknowledges the support of FPI Becas, MICINN, Spain. U. Haroon acknowledges the support of FI-SDUR Becas, MICINN, Spain.